\documentclass{article}
\usepackage[utf8]{inputenc}
\usepackage{amsmath, amsfonts, amssymb, amsthm}
\usepackage{algorithm, algpseudocode}
\usepackage{appendix}
\usepackage{bm, bbm}
\usepackage{booktabs}
\usepackage{float}
\usepackage[a4paper, total={6.2in, 8.6in}]{geometry}
\usepackage{graphicx}
\usepackage[colorlinks,allcolors=blue]{hyperref}
\usepackage{mathrsfs, mathtools}
\usepackage{enumitem}
\usepackage{natbib}
\usepackage[parfill]{parskip}
\usepackage{scalerel}
\usepackage{titlesec}
\usepackage{array,booktabs,tabularx}
\usepackage{xcolor}
\usepackage{verbatim}
\numberwithin{equation}{section}
\usepackage{multirow}
\usepackage{nicematrix}

\newcommand\independent{\protect\mathpalette{\protect\independenT}{\perp}}
\def\independenT#1#2{\mathrel{\rlap{$#1#2$}\mkern2mu{#1#2}}}

\def\argmin{\mathop{\rm arg\, min}}

\newcommand{\bI}{\mathbf{I}}

\newtheorem{assumption}{Assumption}
\newtheorem{theorem}{Theorem}
\newtheorem{lemma}{Lemma}
\newtheorem{proposition}{Proposition}
\newtheorem{corollary}{Corollary}
\newtheorem{remark}{Remark}

\title{Guidance-Induced Activation of Score Estimation Errors in Diffusion Models}
\date{}
\author{
  Pengyu Luo \footnotemark[2]
  \and Yisha Yao \thanks{Department of Statistics, Columbia University. yy3381@columbia.edu} 
  \and Bolun Li \thanks{Department of Statistics, Columbia University. }
}

\begin{document}
\maketitle

\begin{abstract}
Classifier-free guidance (CFG) combines conditional and unconditional score estimates to steer diffusion sampling. In unconditional- or conditional-only diffusion, a score estimate can contain an error component that makes no instantaneous contribution to the evolution of marginals, referred to as \emph{weighted-solenoidal} component.  Since the guided density in CFG generally differs from the conditional and unconditional densities, the weighted-solenoidal error component can affect marginal evolution under guidance. 
We call this mechanism \emph{guidance-induced error activation}.
Using a weighted Helmholtz–Hodge decomposition, we identify the activated error field, derive an exact identity that characterizes when activation vanishes, and quantify the field’s magnitude. At each fixed noise time, the activated error field varies continuously with guidance strength. Its first-order variation in the weak sense has two contributions, one from reweighting the conditional and unconditional weighted-solenoidal errors and the other from the change in the guided density.
For finite score errors, ODE characteristic and SDE Feynman--Kac representations connect error activation to generated densities. In the special case of Gaussian mixture, we show that the activated error can account for most of the KL divergence between learned-score and exact-score CFG ODE sampling at large guidance strengths. 
Controlled Gaussian- and Gamma-mixture experiments support our theoretical results. 
These findings point to a possible explanation for sample degradation at large guidance strengths. They also motivate filtering score estimates to remove its weighted-solenoidal component. Ideal filtering eliminates activation, and qualitative CIFAR-10 comparisons suggest that such filtering can help preserve object structure under strong guidance.
These results identify a mechanism through which guidance changes how score errors affect CFG sampling.
\end{abstract}

\section{Introduction}

Classifier-free guidance (CFG) combines conditional and unconditional score estimates to steer diffusion sampling toward a specified condition \citep{ho2022classifier,rombach2022high,saharia2022photorealistic}.
Increasing the guidance strength can improve alignment with the condition \citep{wu2024theoretical}, but it can also reduce diversity and produce oversaturated or distorted images \citep{sadat2025eliminating, jiao2025unifiedguidance}. Even with exact scores, guidance can shift class means and shrink within-class variances in Gaussian mixtures \citep{pavasovic2026overshoot}.

Several theoretical perspectives have studied the effects of guidance. It is shown that CFG generally does not sample from the usual powered-product distribution \citep{NEURIPS2024_9a3942c2, bradley2025classifier}. 
For Gaussian mixtures, \citet{wu2024theoretical} show that guidance can increase classification confidence and reduce entropy.
\citet{jiang2026analytic} characterize the exact-score guided ODE density by a path integral. 
\citet{skreta2025feynman} use Feynman--Kac corrections to sample prescribed geometric combinations of diffusion marginals.
Pavasovic et al.\ \citep{pavasovic2026overshoot} characterize mean overshoot and variance shrinkage through an analysis of dimensional effects in CFG. 
These results explain important features of guided sampling.

Diffusion samplers can also draw guidance from an auxiliary classifier, as in classifier guidance \citep{dhariwal2021diffusion}, or from a weaker version of the generative model, as in autoguidance \citep{karras2024guiding}.
Several methods address artifacts that can arise when the guidance strength of CFG is strong.
Dynamic thresholding clips and rescales the predicted clean sample to limit extreme values \citep{saharia2022photorealistic}. CFG++ modifies the
sampling update to reduce deviations from the data manifold
\citep{chung2025cfg}, while Adaptive Projected Guidance attenuates a component of the guidance update associated with oversaturation \citep{sadat2025eliminating}.


\paragraph{Problem of interest}
We focus on a complementary question: how does guidance
change the effect of score estimation errors on the
evolving marginal distributions under CFG?
Study suggests that errors in the learned scores introduce another source of distortion. \citet[Theorem~6]{NEURIPS2024_9a3942c2} construct a one-dimensional example in which the conditional and unconditional score estimates each have arbitrarily small $L^2$ error under their respective noised densities, yet sufficiently strong guidance sends samples outside the data support with high probability. We ask whether a score error that has no instantaneous effect on its own reference density can become a leading source of error in CFG sampling.

Our analysis is inspired by \citet{khelifa2026diffusion},
who use a weighted Helmholtz--Hodge decomposition to
characterize the marginal effect of score estimation
errors. Let $p_t^c$ and $p_t^u$ denote the
\emph{conditional and unconditional reference densities} obtained by noising the corresponding data distributions to time $t$. For each $b\in\{c,u\}$, write the weighted Helmholtz--Hodge decomposition of the score estimation error as
\[
e_t^b=\hat s_t^b-s_t^b=v_t^b+w_t^b,
\qquad \nabla\cdot(p_t^b w_t^b)=0,
\]
where $v_t^b$ is the gradient component and $w_t^b$ is the
weighted-solenoidal component.
The marginal evolution equation at $p_t^b$ depends on the score error through $\nabla\cdot(p_t^b e_t^b)$. Only $v_t^b$ contributes to this instantaneous source. In this sense, $w_t^b$ is \emph{unobservable} under its reference density.

At guidance strength $\gamma$, CFG combines the two weighted-solenoidal errors as $w_t^\gamma=\gamma w_t^c+(1-\gamma)w_t^u$. 
But the marginal density $\rho_t^\gamma$ on CFG trajectory
generally differs from both reference densities $p_t^c, \ p_t^u$.
Consequently, $\nabla\cdot(\rho_t^\gamma w_t^\gamma)$ can be nonzero even though each error $w_t^b$ has zero weighted divergence under its own reference density.
We call this mechanism \emph{guidance-induced error activation}. The gradient projection of $w_t^\gamma$ under $\rho_t^\gamma$ identifies the part that can affect the guided marginals. We compare learned-score and exact-score CFG at the same guidance strength to quantify both the activated contribution and the contribution inherited from $v_t^c$ and
$v_t^u$.

\paragraph{Contributions}
\begin{itemize}[leftmargin=*, itemsep=4pt]
\item \textbf{Exact characterization of activation.}
We characterize the error component activated by CFG,
give an exact criterion for zero activation, and quantify
its magnitude through an intrinsic activation energy.
We also show that gradient errors (defined under
$p_t^c$ and $p_t^u$, respectively) retain their gradient
structure under CFG. These results apply to both ODE- and SDE-based reverse sampling.

\item \textbf{Variation with guidance strength.}
We show that the activated error varies continuously
with guidance strength and characterize its first-order
variation in the weak sense. This variation separates
into contributions from reweighting the conditional
and unconditional error components and from altered marginal density under CFG.

\item \textbf{Characterization of generated densities.}
For finite score errors, we derive exact representations
of the densities generated by learned-score CFG in
both ODE and SDE sampling. These results connect
guidance-induced error activation to changes in the
generated distribution. We also provide Kullback--Leibler (KL) identities that quantify the resulting distributional discrepancy.
For ODE sampling, we also prove that the activated KL
contribution eventually dominates the inherited contribution
for a class of finite Gaussian mixtures as $\gamma\to\infty$.

\item \textbf{Score filtering and empirical evidence.}
We propose filtering each score estimate to remove its
weighted-solenoidal component.
Ideal filtering eliminates activation under CFG while preserving the gradient error. Qualitative CIFAR-10
comparisons suggest that filtering can help preserve object structure under strong guidance.
\end{itemize}

\section{Preliminaries on diffusion models}
\label{sec:diffusion_preliminaries}

We study ODE-based diffusion sampling, including the continuous-time dynamics underlying deterministic denoising diffusion implicit models (DDIMs). 
Let $p_t^u$ and $p_t^c$ denote the unconditional and conditional distributions at time $t$, respectively. 
A single noise-time index $t\in[0,T]$ is used throughout the paper: forward noising runs from the clean data at $t=0$ to the noisy distribution at $t=T$, and backward sampling runs from $T$ to $0$. Thus, $p_t^b$ and $s_t^b$, $b\in\{u, c\}$,  always refer to the same noise level, regardless of the direction in which the process is traversed. Derivatives such as $\partial_t p_t^b$ are taken in the increasing noise-time coordinate.

\subsection{Unconditional diffusion}
\label{sec:vanilla_diffusion}

An unconditional diffusion model learns a data distribution on $\mathbb R^d$ without taking a conditioning label as input \citep{ho2020denoising,song2021scorebased}. The forward noising process defines the reference densities used throughout the paper:
\begin{equation*}
\mathrm dX_t=f(t)X_t\,\mathrm dt+\sigma(t)\,\mathrm dB_t,
\qquad X_0\sim p_{\mathrm{data}}^u,
\qquad t:0\longrightarrow T,
\end{equation*}
where $B_t$ is standard $d$-dimensional Brownian motion and $f(t),\sigma(t)$ are scalar functions. At each fixed noise time,
\begin{equation*}
X_t\overset{\mathrm d}=\alpha_tX_0+\bar\sigma_t\varepsilon,
\qquad \varepsilon\sim\mathcal N(0,\bI),
\qquad \varepsilon\independent X_0,
\end{equation*}
with
\begin{equation*}
\alpha_t=\exp\!\left(\int_0^t f(u)\,\mathrm du\right),
\quad
\bar\sigma_t^2=\alpha_t^2\int_0^t
\frac{\sigma^2(u)}{\alpha_u^2}\,\mathrm du, \quad \alpha_0=1, \ \bar\sigma_0=0, \ \bar\sigma_t>0 \text{ for } t>0. 
\end{equation*}
In variance-preserving schedules, $\alpha_t^2+\bar\sigma_t^2=1$, and $\alpha_t \searrow 0, \ \bar\sigma_t \nearrow 1$ so the noised distribution approaches $\mathcal N(0, \bI)$. For variance-exploding schedules \citep{song2021scorebased}, $f(t)=0, \ \alpha_t\equiv 1$, and the rescaled variable $X_t/\bar\sigma_t$ approaches $\mathcal N(0,\bI)$
in distribution.
Other common choices include cosine \citep{Nichol2021improved}
and sigmoid profiles \citep{pmlr-v202-jabri23a, chen2023importance}, as well as Laplace- and
Cauchy-based schedules \citep{hang2025improved} for the log-SNR $\log(\alpha_t^2/\bar\sigma_t^2)$.
These choices determine the progression of noise levels
with time.

Let $\phi_v$ denote the density of $\mathcal N(0,v\bI)$ for $v>0$. The noised unconditional density and its exact score are
\begin{equation}
\begin{aligned}
p_t^u(x)&=\int_{\mathbb R^d}
\phi_{\bar\sigma_t^2}(x-\alpha_tz)
p_{\mathrm{data}}^u(z)\,\mathrm dz,\\
s_t^u(x)&=\nabla_x\log p_t^u(x).
\end{aligned}
\label{eq:prelim_unconditional_density}
\end{equation}
Under the usual time-reversal regularity and uniqueness conditions, backward process takes the form
\begin{equation*} 
\mathrm dY_t =\left[f(t)Y_t-\sigma_t^2 s_t^u(Y_t)\right]\mathrm dt +\sigma(t)\,\mathrm d\widetilde B_t,
\qquad t:T\longrightarrow0, \quad 
\qquad Y_t\overset{\mathrm d}=X_t.
\end{equation*}

The corresponding probability-flow ODE is
\begin{equation}
\mathrm dY_t=\zeta_t^u(Y_t)\,\mathrm dt, \quad
\zeta_t^u(x):=f(t)x-\kappa_t s_t^u(x), \quad 
\kappa_t:=\sigma_t^2/2,
\quad t:T\longrightarrow0.
\label{eq:prelim_probability_flow}
\end{equation}
With exact scores and the same initialization, the corresponding reverse-time ODE and SDE have same marginals at all $t$. 
Additionally, the reverse ODE and SDE share the same continuity equation
\begin{equation*}
\partial_t p_t^u+\nabla\cdot(p_t^u\zeta_t^u)=0.
\end{equation*}

The reverse-time SDE describes the continuous-time limit of stochastic Denoising Diffusion Probabilistic Model (DDPM) sampling, while the probability-flow ODE describes that of deterministic Denoising Diffusion Implicit Model (DDIM) sampling. We refer to these formulations as SDE-based and ODE-based sampling, respectively.

To implement DDPM or DDIM, both forward noising and backward sampling run in discrete steps. For a forward grid $0=t_0<t_1<\cdots<t_N=T$, write $X_\ell=X_{t_\ell}$, $p_\ell^u=p_{t_\ell}^u$, and $s_\ell^u=s_{t_\ell}^u$. The exact one-step forward transitions are
\begin{align*}
X_\ell&=a_\ell X_{\ell-1}+b_\ell\varepsilon_\ell,
\quad \varepsilon_\ell\overset{\mathrm{iid}}{\sim}\mathcal N(0,\bI), \quad \varepsilon_\ell\independent X_k \ \forall\ k<\ell\\
a_\ell&=\frac{\alpha_{t_\ell}}{\alpha_{t_{\ell-1}}},
\qquad b_\ell^2=\bar\sigma_{t_\ell}^2-a_\ell^2\bar\sigma_{t_{\ell-1}}^2>0.
\end{align*}  
Let $Y_\ell$ denote the state at noise time $t_\ell$ on the reverse chain.
Deterministic DDIM updates $Y_\ell$ to $Y_{\ell-1}$ for $\ell=N,\ldots,1$ according to
\begin{equation*}
\begin{aligned}
Y_{\ell-1} =\frac{\alpha_{t_{\ell-1}}}{\alpha_{t_\ell}}Y_\ell + \left(
  \frac{\alpha_{t_{\ell-1}}}{\alpha_{t_\ell}}
  \bar\sigma_{t_\ell}^2
  -\bar\sigma_{t_{\ell-1}}\bar\sigma_{t_\ell}
  \right)\hat s_{t_\ell}^u(Y_\ell).
\end{aligned}
\end{equation*}
The corresponding reverse DDPM takes the form
\begin{equation*}
Y_{\ell-1} =\frac{Y_\ell+b_\ell^2s_\ell^u(Y_\ell)}{a_\ell} + \frac{b_\ell}{a_\ell}\varepsilon_\ell', \quad \varepsilon_\ell' \overset{\mathrm{iid}}{\sim}\mathcal N(0,\bI),
\quad Y_N\overset{\mathrm d}=X_N, \quad \varepsilon_\ell'\independent Y_k \ \forall\ k\geq\ell,
\quad \ell=N,\ldots,1.
\end{equation*}

In practice, the score is estimated by denoising score matching:
\begin{equation*}
\begin{aligned}
\mathcal J_u(\theta)
&=\mathbb E_{t,X_0,\varepsilon}\!\left[
\lambda(t)\left\|s_\theta^u(\alpha_tX_0+\bar\sigma_t\varepsilon,t)
+\frac{\varepsilon}{\bar\sigma_t}\right\|^2\right],\\
\hat\theta_u&\in\argmin_\theta\mathcal J_u(\theta),
\qquad \hat s_t^u(x):=s_{\hat\theta_u}^u(x,t),
\end{aligned}
\end{equation*}
where $\lambda(t)>0$ is a positive time weight. The unrestricted population minimizer is $s_t^u$. 
Under $s_\theta^u(x,t)=-\varepsilon_\theta^u(x,t)/\bar\sigma_t$, the equivalent squared noise-prediction loss has weight $\lambda(t)/\bar\sigma_t^2$. Replacing $s_t^u$ by its estimate $\hat s_t^u$, the discrete SDE and ODE sampler are fully implementable. 

The error of DDPM or DDIM originates from three sources: discretization error due to finite-step Gaussian chain, initialization bias as the reverse chain starts at $\mathcal N(0, \bI)$ instead of $p_T^u$, and score estimation error as implementations only have access to learned scores.

\subsection{Conditional diffusion}
\label{sec:conditional_diffusion}

Let $C$ be a class label with probabilities $\pi_k=\mathbb P(C=k)$, and fix a target label $c_\star$. Define
\begin{equation*}
p_{\mathrm{data}}^u(x)
=\sum_k\pi_k p_{\mathrm{data}}(x\mid C=k),
\qquad
p_{\mathrm{data}}^c(x):=p_{\mathrm{data}}(x\mid C=c_\star).
\end{equation*}
Using the same noising schedule, the conditional reference density and score are
\begin{equation}
\begin{aligned}
p_t^c(x)&=\int_{\mathbb R^d}
\phi_{\bar\sigma_t^2}(x-\alpha_tz)
p_{\mathrm{data}}^c(z)\,\mathrm dz, \quad X_0^c \sim p_0^c=p_{\mathrm{data}}^c, \\
s_t^c(x)&=\nabla_x\log p_t^c(x).
\end{aligned}
\label{eq:prelim_conditional_density}
\end{equation}
Conditional sampling is obtained by replacing $s^u$ with $s^c$ in the reverse SDE or ODE, and backward Gaussian chain approximation, without changing the noise schedule $\alpha_t, \bar\sigma_t$ or step coefficients $a_\ell, b_\ell$. 
Initialization from $p_T^c$ gives marginals $p_t^c$, under the same regularity and uniqueness conditions. The conditional DDIM or DDPM update likewise replaces $\hat s^u$ by $\hat s^c$. 

A conditional score network takes the class label as an additional input and is trained by
\begin{equation*}
\begin{aligned}
\mathcal J_c(\theta)
&=\mathbb E_{t,(X_0,C),\varepsilon}\!\left[
\lambda(t)\left\|s_\theta(\alpha_tX_0+\bar\sigma_t\varepsilon,t,C)
+\frac{\varepsilon}{\bar\sigma_t}\right\|^2\right],\\
\hat\theta_c&\in\argmin_\theta\mathcal J_c(\theta),
\qquad \hat s_t^c(x):=s_{\hat\theta_c}(x,t,c_\star).
\end{aligned}
\end{equation*}
At the target label, the unrestricted population minimizer is $s_t^c$.

\subsection{Classifier-free guidance}
\label{sec:cfg_preliminaries}

CFG combines conditional and unconditional score estimates during sampling \citep{ho2022classifier}. The estimates may be trained separately or jointly in a shared network with a null label $\varnothing$. With independent label-drop probability $\delta\in(0,1)$, the shared model minimizes
\begin{equation*}
\mathcal J_{\mathrm{CFG}}(\theta)
=(1-\delta)\mathcal J_c(\theta)+\delta\mathcal J_u(\theta),
\qquad s_\theta^u(x,t)=s_\theta(x,t,\varnothing).
\end{equation*}
In this case, $\hat s_t^c$ and $\hat s_t^u$ are evaluations of the same fitted network at $c_\star$ and $\varnothing$.

For a finite guidance strength $\gamma\geq0$, define
\begin{equation*}
\begin{aligned}
s_t^\gamma&:=\gamma s_t^c+(1-\gamma)s_t^u,\\
\hat s_t^\gamma&:=\gamma\hat s_t^c+(1-\gamma)\hat s_t^u.
\end{aligned}
\end{equation*}
The choices $\gamma=0$ and $\gamma=1$ recover ordinary unconditional and conditional sampling, while $\gamma>1$ extrapolates beyond the conditional score. 
Write the individual and combined score errors as
\begin{equation*}
e_t^b:=\hat s_t^b-s_t^b,\qquad b\in\{c,u\},
\qquad e_t^\gamma:=\gamma e_t^c+(1-\gamma)e_t^u.
\end{equation*}

The exact-score and learned-score guided ODEs use the velocities
\begin{equation*}
\begin{aligned}
\zeta_t^\gamma(x)&:=f(t)x-\kappa_t s_t^\gamma(x),\\
\hat\zeta_t^\gamma(x)&:=f(t)x-\kappa_t\hat s_t^\gamma(x)
=\zeta_t^\gamma(x)-\kappa_t e_t^\gamma(x).
\end{aligned}
\end{equation*}
Let $\rho_t^\gamma$ and $\hat\rho_t^\gamma$ denote the marginal densities of the exact-score and learned-score guided ODEs under CFG. Both start from the same density $\pi$ at $t=T$ and evolve according to
\begin{equation*}
\begin{aligned}
\partial_t\rho_t^\gamma
+\nabla\cdot(\rho_t^\gamma\zeta_t^\gamma)&=0,
&\rho_T^\gamma&=\pi,\\
\partial_t\hat\rho_t^\gamma
+\nabla\cdot(\hat\rho_t^\gamma\hat\zeta_t^\gamma)&=0,
&\hat\rho_T^\gamma&=\pi.
\end{aligned}
\end{equation*}

For the reverse SDEs, let \(\rho_{t,\mathrm S}^\gamma\) and \(\hat\rho_{t,\mathrm S}^\gamma\) denote the exact-score and learned-score guided marginal densities under CFG. 
Their drift fields take the forms
\begin{equation*}
    \begin{aligned}
\xi_t^\gamma(x) &:=f(t)x-\sigma_t^2s_t^\gamma(x)
=f(t)x-2\kappa_t s_t^\gamma(x),\\
\hat \xi_t^\gamma(x) &:=f(t)x-\sigma_t^2\hat s_t^\gamma(x) = \xi_t^\gamma(x)-2\kappa_t e_t^\gamma(x).
\end{aligned}
\end{equation*}
The SDEs satisfy the reverse-time Fokker–Planck equations
\begin{equation*}
    \begin{aligned}
\partial_t\rho_{t,\mathrm S}^\gamma
+\nabla\cdot\!\left(\rho_{t,\mathrm S}^\gamma \xi_t^\gamma\right)
&=-\kappa_t\Delta\rho_{t,\mathrm S}^\gamma,\\
\partial_t\hat\rho_{t,\mathrm S}^\gamma
+\nabla\cdot\!\left(\hat\rho_{t,\mathrm S}^\gamma\hat\xi_t^\gamma\right)
&=-\kappa_t\Delta\hat\rho_{t,\mathrm S}^\gamma,
\end{aligned}
\end{equation*}

The reference densities $p_t^u,p_t^c$ are fixed by forward noising, whereas the guided ODE and SDE densities depend on their respective sampling dynamics and starting density $\pi$. At a finite noising horizon, reproducing the reference path $p_t^b$ with either exact-score sampler requires initialization from $p_T^b$, for $b\in\{u,c\}$.
Under the idealization $p_T^u=p_T^c=\pi$, the stated
uniqueness assumptions give
\begin{equation*}
\rho_t^0=\rho_{t,\mathrm S}^0 =p_t^u, \quad \rho_t^1=\rho_{t,\mathrm S}^1 =p_t^c.
\end{equation*}

The reference densities $p_t^u,p_t^c$ are fixed by forward noising. At a finite noising horizon, identifying either reference path requires initialization from its own density $p_T^b, \ b=u,c$.
Whereas $\rho_t^\gamma,\hat\rho_t^\gamma$ ($\rho_{t,\mathrm S}^\gamma, \hat\rho_{t,\mathrm S}^\gamma$) are determined by the guided ODEs (SDEs).
Under the idealization $p_T^u=p_T^c=\pi$, uniqueness gives $\rho_{t,\mathrm S}^0 = \rho_t^0=p_t^u$ and $\rho_{t,\mathrm S}^1=\rho_t^1=p_t^c$.

Although
\begin{equation}
s_t^\gamma = \nabla_x\log\left[(p_t^c)^\gamma(p_t^u)^{1-\gamma}\right],
\label{eq:prelim_powered_product_score}
\end{equation}
the guided densities $\rho_t^\gamma, \ \rho_{t,\mathrm S}^\gamma$ generally differs from this powered product, even when the latter can be normalized. In particular, $s_t^\gamma$ need not equal $\nabla\log\rho_t^\gamma$ or $\nabla\log \rho_{t,\mathrm S}^\gamma$. Consequently, the exact-score marginal equivalence between the reverse ODE and SDE does not automatically extend to CFG \citep{bradley2025classifier}. Throughout the analysis below, $\rho_t^\gamma$ ($\rho_{t,\mathrm S}^\gamma$) refers to the guided ODE (SDE) density.

\section{Error activation at a fixed guidance strength}
\label{sec:fixed_guidance}
In unconditional or conditional diffusion, the score estimation error can be decomposed into a gradient component and a weighted-solenoidal component with respect to the corresponding reference density, \(p_t^u\) or \(p_t^c\). This weighted Helmholtz–Hodge decomposition depends on the reference density: a component with zero weighted divergence under one density need not have zero weighted divergence under another. CFG changes the sampling dynamics and induces the combined-score guided density \(\rho_t^\gamma\), which generally differs from both \(p_t^u\) and \(p_t^c\). Consequently, error components that are solenoidal under their original reference densities can acquire a nonzero gradient projection under \(\rho_t^\gamma\) and become observable in the guided marginal dynamics. In this section, we characterize this guidance-induced error activation at a fixed guidance strength.

\paragraph{Notations} For a probability density \(\rho\) on \(\mathbb R^d\), let \(L^2(\rho):=L^2(\rho;\mathbb R^d)\) denote the Hilbert space of square-integrable vector fields, equipped with the inner product \(\langle z_1,z_2\rangle_\rho=\int z_1(x)\cdot z_2(x)\,\rho(x)\,dx\).
Let
\begin{equation}
\mathcal G_\rho
=\overline{\{\nabla\psi:\psi\in C_c^\infty(\mathbb R^d)\}}^{\,L^2(\rho)}
\label{eq:main_gradient_space}
\end{equation}
be the closed gradient subspace, and let $P_\rho$ be its orthogonal projection.

\subsection{Weighted error decomposition for CFG}
\label{sec:cfg_hodge}
Under CFG, the combined score estimation error is
\begin{equation}
e_t^\gamma=\gamma e_t^c+(1-\gamma)e_t^u.
\end{equation}
For each $b\in\{c,u\}$, decompose the score error under its reference density:
\begin{equation}
v_t^b=P_{p_t^b}e_t^b,\qquad
w_t^b=(I-P_{p_t^b})e_t^b.
\label{eq:native_decomposition}
\end{equation}
We call \(v_t^b\) the \emph{gradient error} and \(w_t^b\) the \emph{weighted-solenoidal error} relative to the reference density \(p_t^b\). By construction, \(w_t^b\) satisfies \(\nabla\cdot(p_t^b w_t^b)=0\) in the distributional sense and therefore contributes no instantaneous source to the marginal continuity equation evaluated at \(p_t^b\). Throughout this paper, observability refers to whether an error component produces a nonzero instantaneous source under the density being considered. If the zero-source condition holds at every time, the reference and perturbed dynamics share the same distribution at \(T\), and the perturbed continuity equation has a unique solution, then adding the weighted-solenoidal component alone preserves the entire reference marginal path.

Under CFG, we combine the gradient and weighted-solenoidal errors defined with respect to their respective reference densities:
\begin{equation}\label{eq:cfg_error_combination}
\begin{aligned}
v_t^\gamma&=\gamma v_t^c+(1-\gamma)v_t^u,\\
w_t^\gamma&=\gamma w_t^c+(1-\gamma)w_t^u.
\end{aligned}
\end{equation}
The total guided score error therefore satisfies
$e_t^\gamma=v_t^\gamma+w_t^\gamma$.
Although $w_t^c$ and $w_t^u$ are weighted-solenoidal under
$p_t^c$ and $p_t^u$, respectively, their combination
$w_t^\gamma$ need not be weighted-solenoidal under the guided
density $\rho_t^\gamma$. To characterize their contributions
to the guided marginal dynamics, we project these fields onto
the gradient subspace $\mathcal G_{\rho_t^\gamma}$:
\begin{equation}
\begin{aligned}
h_t^\gamma&:=P_{\rho_t^\gamma}v_t^\gamma,\\
g_t^\gamma&:=P_{\rho_t^\gamma}w_t^\gamma,\\
o_t^\gamma&:=P_{\rho_t^\gamma}e_t^\gamma
=h_t^\gamma+g_t^\gamma,
\end{aligned}
\label{eq:leakage_decomposition}
\end{equation}
where the final equality follows from the linearity of the projection.

Here $h_t^\gamma$ is the \emph{inherited gradient contribution}, and $g_t^\gamma$ is the \emph{activated error component}.
We use \emph{guidance-induced error activation} to describe
the emergence of a nonzero $g_t^\gamma$ when errors that are
weighted-solenoidal under their respective reference densities
become observable in the guided marginal dynamics.
Both $h_t^\gamma$ and $g_t^\gamma$ belong to
$\mathcal G_{\rho_t^\gamma}$, and their $L^2(\rho_t^\gamma)$ inner product need not vanish.
Consequently, the decomposition
$o_t^\gamma=h_t^\gamma+g_t^\gamma$ is generally not orthogonal,
and the inherited and activated contributions can reinforce
or cancel one another.

\begin{figure}
    \centering
    \includegraphics[width=1\linewidth]{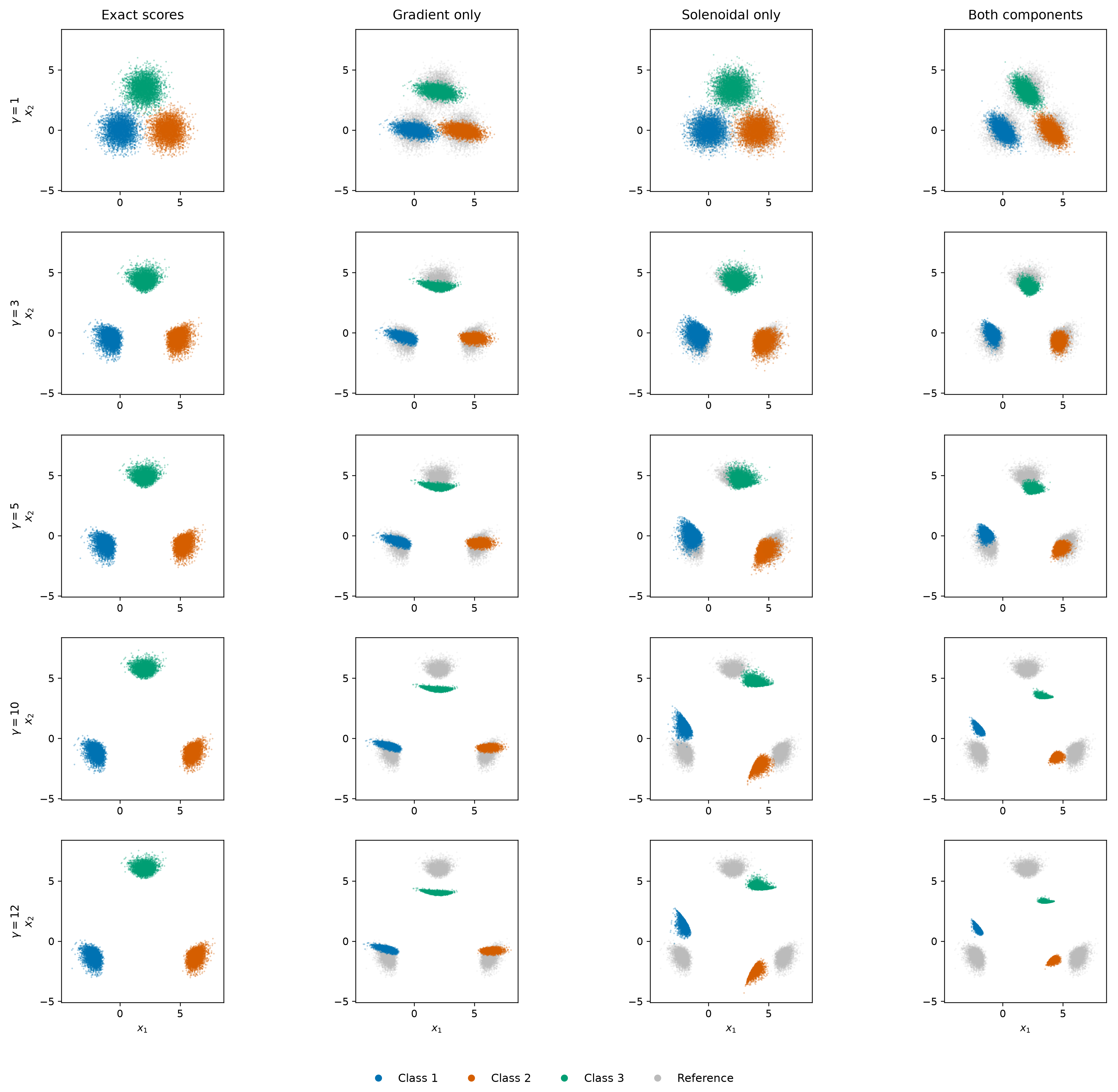}
    \caption{How the two error components, gradient and weighted-solenoidal components affect CFG sampling. The first column depicts the distributions generated by exact-score CFG across different guidance strength. In the second and third columns, the scores used in CFG are contaminated with gradient-only and solenoidal-only errors, respectively. The last column depicts distributions generated by CFG with scores contaminated with both error. In the second, third, fourth columns, the gray shadows correspond to the data clouds generated by exact-score CFG. }
    \label{fig:gmm_error_components}
\end{figure}

To provide intuition for how the two error components affect CFG sampling, we consider a simple Gaussian mixture in which each component represents a class. We add explicitly constructed gradient and weighted-solenoidal error fields to the exact scores, allowing us to examine their individual and joint effects in a controlled setting.
Figure~\ref{fig:gmm_error_components} compares samples
generated using exact scores and scores perturbed by
gradient errors alone, weighted-solenoidal errors alone,
or both components, across different guidance strengths.

To quantify error activation, let $\dot H^1(\rho)$ denote the
completion of $C_c^\infty(\mathbb R^d)$, the space of smooth, compactly supported scalar-valued functions, under the gradient seminorm
\[
\|\psi\|_{\dot H^1(\rho)}
:=\left(\int_{\mathbb R^d}\|\nabla\psi(x)\|^2
\,\rho(x)\,dx\right)^{1/2} .
\]
Denote its continuous dual by $\dot H^{-1}(\rho)$.
For a vector field $z\in L^2(\rho)$, define the
\emph{negative weighted-divergence operator}
$Q_\rho z\in\dot H^{-1}(\rho)$ through
\begin{equation}
\langle Q_\rho z,\psi\rangle
=\int_{\mathbb R^d}z\cdot\nabla\psi\,\rho\,dx,
\qquad \psi\in\dot H^1(\rho),
\label{eq:weighted_divergence_operator}
\end{equation}
where $\langle\cdot,\cdot\rangle$ denotes the dual pairing.
The functional is continuous by the Cauchy--Schwarz inequality,
so $Q_\rho$ is well defined. When $\rho$ is strictly positive and both $\rho$ and $z$ are sufficiently smooth, integration by parts gives the pointwise representation $Q_\rho z=-\rho^{-1}\nabla\cdot(\rho z)$.
Intuitively, \(Q_\rho z\) measures how a vector field \(z\) redistributes probability mass relative to the density \(\rho\).
Imagine particles distributed according to \(\rho\), and move each particle slightly by the vector field $z$,
\[ x\longmapsto x + \varepsilon \cdot z(x). \]
If \(\rho_\varepsilon\) is the resulting density, then, under suitable regularity,
\[
\left.\frac{\partial\rho_\varepsilon}{\partial\varepsilon}\right|_{\varepsilon=0} = -\nabla\cdot(\rho z) =\rho\,Q_\rho z.
\]
Thus, \(Q_\rho z\) is the instantaneous relative change in density at each location:
\[
Q_\rho z =\left.\frac{\partial}{\partial\varepsilon} \log\rho_\varepsilon\right|_{\varepsilon=0}.
\]
Positive values indicate accumulation of probability mass, while negative values indicate depletion.

We define the \emph{activation residual} and
\emph{activation energy}, respectively, as
\begin{equation}
q_t^\gamma:=Q_{\rho_t^\gamma}w_t^\gamma,
\qquad
\mathcal L_t(\gamma)
:=\|g_t^\gamma\|_{L^2(\rho_t^\gamma)}^2.
\label{eq:leakage_source_energy}
\end{equation}
The residual $q_t^\gamma$ detects whether the combined field
$w_t^\gamma$ has a nonzero weighted divergence under the
guided density $\rho_t^\gamma$. $q_t^\gamma$ vanishes precisely when $w_t^\gamma$ is weighted-solenoidal with respect to that density
\[
q_t^\gamma=0 \quad\Longleftrightarrow\quad \nabla\cdot(\rho_t^\gamma w_t^\gamma)=0.
\]
The energy $\mathcal L_t(\gamma)$ quantifies the strength of
activation through the squared $L^2(\rho_t^\gamma)$ norm of
the activated gradient component $g_t^\gamma$.

\subsection{Exact characterization at a fixed guidance strength}
\label{sec:fixed_guidance_theorem}

\begin{assumption}[Regularity at a fixed guidance strength]
\label{ass:fixed_guidance}
For each noise time $t$ and finite guidance strength $\gamma$
under consideration, the densities $p_t^c$, $p_t^u$, and
$\rho_t^\gamma$ are strictly positive, and the corresponding
exact-score dynamics are well posed. For each $b\in\{c,u\}$,
assume that $e_t^b\in L^2(p_t^b)$ and
$v_t^b,w_t^b\in L^2(\rho_t^\gamma)$. In addition, the
log-density ratios $\log(\rho_t^\gamma/p_t^b)$ are locally
$C^1$, and
$w_t^b\cdot\nabla\log(\rho_t^\gamma/p_t^b)$ is locally
integrable with respect to $\rho_t^\gamma(x)\,dx$.
\end{assumption}
\bigskip
\begin{theorem}[Error activation at a fixed guidance strength]
\label{thm:fixed_guidance}
Under Assumption~\ref{ass:fixed_guidance}, the activation
residual $q_t^\gamma$ and activated error component
$g_t^\gamma$ satisfy the following identities.

\textit{(i) Change-of-density identity.}
The activation residual is given by
\begin{equation}
q_t^\gamma
=-\gamma w_t^c\cdot\nabla\log\frac{\rho_t^\gamma}{p_t^c}
-(1-\gamma)w_t^u\cdot\nabla\log\frac{\rho_t^\gamma}{p_t^u}.
\label{eq:exact_leakage_source}
\end{equation}
Moreover, $Q_{\rho_t^\gamma}g_t^\gamma=q_t^\gamma$, that is,
the combined field and its gradient projection have the same
negative weighted divergence under the guided density.

\textit{(ii) Activation energy.}
The activation energy equals the squared dual norm of the
activation residual:
\begin{equation}
\mathcal L_t(\gamma)
=\|g_t^\gamma\|_{L^2(\rho_t^\gamma)}^2
=\|q_t^\gamma\|_{\dot H^{-1}(\rho_t^\gamma)}^2.
\label{eq:intrinsic_leakage_energy}
\end{equation}

\textit{(iii) Criterion for zero activation.}
Activation is absent precisely when the following equivalent
conditions hold:
\begin{equation}
g_t^\gamma=0
\iff q_t^\gamma=0
\iff \nabla\cdot(\rho_t^\gamma w_t^\gamma)=0,
\label{eq:zero_leakage_criteria}
\end{equation}
where the last equality is understood in the distributional
sense. 
\end{theorem}

Equation~\eqref{eq:exact_leakage_source} shows how a change
of reference density activates previously weighted-solenoidal
errors (invisible under the conditional or unconditional reference density). Each contribution depends on the alignment of $w_t^b$ with the gradient of the log-density ratio
$\log(\rho_t^\gamma/p_t^b)$. Activation occurs when these
interactions produce a nonzero combined residual. It vanishes
if the two contributions cancel. A sufficient condition for
zero activation is $w_t^b\cdot\nabla\log(\rho_t^\gamma/p_t^b)=0$ for
$b\in\{c,u\}$. Interpreted geometrically: each field $w_t^b$ is tangent to the level sets of its density ratio. These identities do not imply that activation grows monotonically with $\gamma$.

The activation energy also admits the dual variational
representation
\begin{equation}
\mathcal L_t(\gamma)
=\sup_{\psi\in\dot H^1(\rho_t^\gamma)}
\left\{
2\langle q_t^\gamma,\psi\rangle
-\|\nabla\psi\|_{L^2(\rho_t^\gamma)}^2
\right\}.
\label{eq:leakage_dual}
\end{equation}
This representation provides a variational objective for
estimating activation energy without explicitly solving the weighted Poisson equation. Proposition 1 in Appendix~\ref{app:main_theorem_one} gives equivalent Poisson and minimum-energy representations.

We now examine how score errors change the density generated
by CFG. Consider a sampler using the error-contaminated score
$s_t^\gamma+\varepsilon e_t^\gamma$, with velocity
$\zeta_t^\gamma-\varepsilon\kappa_t e_t^\gamma$.
Here $\varepsilon=0$ gives exact-score CFG, while
$\varepsilon=1$ gives learned-score CFG. Let
$\rho_t^{\gamma,\varepsilon}$ denote the resulting marginal
density. Every sampler starts from the same distribution
at $t=T$ and runs backward to $t=0$, with
$\rho_T^{\gamma,\varepsilon}=\rho_T^\gamma=\pi$.
Assume these flows exist uniquely, their marginal densities
are differentiable in $\varepsilon$ near zero, and the
derivative can be passed through the weak continuity equation.
Define
\[
\eta_t^\gamma
:=\left.\partial_\varepsilon
\rho_t^{\gamma,\varepsilon}\right|_{\varepsilon=0}.
\]
For small $\varepsilon$, the density change
$\rho_t^{\gamma,\varepsilon}-\rho_t^\gamma$ is therefore
approximately $\varepsilon\eta_t^\gamma$ in the weak sense.
Differentiating the continuity equation gives
\begin{equation}
\partial_t\eta_t^\gamma
+\nabla\cdot(\zeta_t^\gamma\eta_t^\gamma)
=\kappa_t\nabla\cdot(\rho_t^\gamma e_t^\gamma),
\qquad \eta_T^\gamma=0.
\label{eq:linearized_density_response}
\end{equation}
For small $\varepsilon$, the score error produces a density difference
\[
\rho_t^{\gamma,\varepsilon}-\rho_t^\gamma
\approx\varepsilon\eta_t^\gamma,
\]
which the reference sampling flow moves and reshapes.
The term $\nabla\cdot(\zeta_t^\gamma\eta_t^\gamma)$ accounts
for how the exact-score guided velocity moves and reshapes
the density difference between the perturbed and reference
samplers.
The right-hand side describes the additional redistribution of probability mass caused by the score error. The condition
$\eta_T^\gamma=0$ follows from the common entrance law.

Since $e_t^\gamma=v_t^\gamma+w_t^\gamma$, the right-hand side term separates into contributions from these two error components.
The contribution from $w_t^\gamma$ is
\begin{equation}\label{eq:leakage_density_source}
\kappa_t\nabla\cdot(\rho_t^\gamma w_t^\gamma) = -\kappa_t\rho_t^\gamma q_t^\gamma.
\end{equation}
Thus, the activation residual determines how errors that are
weighted-solenoidal under their own reference densities
contribute to the guided marginal dynamics. This contribution
can reinforce or cancel that of $v_t^\gamma$, so nonzero
activation alone does not imply an increase in a given
sampling discrepancy.

To complete the decomposition, we also examine how changing
the reference density affects the gradient errors $v_t^\gamma$.
The following corollary shows that their gradient structure
is preserved under suitable smoothness and integrability
conditions. This identifies the inherited contribution
directly with $v_t^\gamma$ and simplifies the expression
for the total observable error under $\rho_t^\gamma$.
 
\bigskip
\begin{corollary}[Preservation of gradient structure under a change of density]
\label{cor:one_way_transfer}
Under Assumption~\ref{ass:fixed_guidance}, suppose additionally
that $p_t^c$, $p_t^u$, and $\rho_t^\gamma$ are $C^1$.
Then the combined gradient error remains in the guided
gradient subspace:
\begin{equation}\label{eq:one_way_transfer}
P_{\rho_t^\gamma}v_t^\gamma=v_t^\gamma,
\qquad (I-P_{\rho_t^\gamma})v_t^\gamma=0.
\end{equation}
Consequently, $h_t^\gamma=v_t^\gamma$, and the total
observable error is
\begin{equation*}
o_t^\gamma=v_t^\gamma+g_t^\gamma.
\end{equation*}
\end{corollary}

\subsection{Extension to reverse-time SDE sampling}
\label{sec:reverse_sde_activation}

The activation identities depend on the density used in the
weighted projection and therefore also apply to the
continuous-time reverse SDE underlying DDPM sampling.
Consider exact-score CFG with
\begin{equation}
\begin{aligned}
\mathrm dY_t &=\xi_t^\gamma(Y_t)\,\mathrm dt + \sigma(t)\,\mathrm d\widetilde B_t,
\quad Y_T\sim\pi, \quad t:T\longrightarrow 0,\\
\xi_t^\gamma(x)&=f(t)x-\sigma_t^2s_t^\gamma(x),
\end{aligned}
\label{eq:reverse_sde_guidance}
\end{equation}
where $\widetilde B_t$ is reverse-time Brownian motion with
increment covariance $|\mathrm dt|\bI$. Let $\rho_{t,\mathrm S}^\gamma$ denote its marginal density. Even with the same starting
density $\pi$, this density generally differs from the ODE
marginal $\rho_t^\gamma$ under guidance
\citep{bradley2025classifier}.

Define the SDE activation quantities using its own marginal density:
\begin{equation*} 
g_{t,\mathrm S}^\gamma :=P_{\rho_{t,\mathrm S}^\gamma}w_t^\gamma, \qquad
q_{t,\mathrm S}^\gamma :=Q_{\rho_{t,\mathrm S}^\gamma}w_t^\gamma.
\end{equation*}
The corresponding activation energy is
$\mathcal L_{t,\mathrm S}(\gamma)
:=\|g_{t,\mathrm S}^\gamma\|_{L^2(\rho_{t,\mathrm S}^\gamma)}^2$.

\begin{proposition}[Activation under reverse-time SDE sampling] \label{prop:reverse_sde_activation}
Suppose Eq.~\eqref{eq:reverse_sde_guidance} admits a unique non-explosive weak solution on $[0,T]$.
Fix a noise time $t\in(0,T)$ and a finite guidance strength $\gamma$. Assume that $p_t^c,p_t^u$, and
$\rho_{t,\mathrm S}^\gamma$ are strictly positive $C^1$
probability densities and, for each $b\in\{c,u\}$,
\begin{equation*}
e_t^b\in L^2(p_t^b), \qquad
v_t^b,w_t^b\in L^2(\rho_{t,\mathrm S}^\gamma).
\end{equation*}
Then the following statements hold.

\textit{(i) Activation identity.}
In the weighted weak sense,
\begin{equation}\label{eq:sde_change_density_identity}
q_{t,\mathrm S}^\gamma =-\gamma w_t^c\cdot\nabla\log
  \frac{\rho_{t,\mathrm S}^\gamma}{p_t^c} - (1-\gamma)w_t^u\cdot\nabla\log
  \frac{\rho_{t,\mathrm S}^\gamma}{p_t^u}.
\end{equation}
Moreover,
$Q_{\rho_{t,\mathrm S}^\gamma}g_{t,\mathrm S}^\gamma
=q_{t,\mathrm S}^\gamma$.

\textit{(ii) Activation energy and zero activation.}
The intrinsic energy satisfies
\begin{equation*} 
\mathcal L_{t,\mathrm S}(\gamma) = \|g_{t,\mathrm S}^\gamma\|_{L^2(\rho_{t,\mathrm S}^\gamma)}^2
= \|q_{t,\mathrm S}^\gamma\|_{\dot H^{-1}(\rho_{t,\mathrm S}^\gamma)}^2.
\end{equation*}
Moreover,
\begin{equation}
g_{t,\mathrm S}^\gamma=0
\iff q_{t,\mathrm S}^\gamma=0
\iff\nabla\cdot(\rho_{t,\mathrm S}^\gamma w_t^\gamma)=0,
\label{eq:sde_zero_activation}
\end{equation}
with the final equality understood distributionally.

\textit{(iii) Preservation of gradient structure.}
The combined gradient error remains in the SDE gradient
subspace:
\begin{equation}
P_{\rho_{t,\mathrm S}^\gamma}v_t^\gamma=v_t^\gamma,
\qquad
(I-P_{\rho_{t,\mathrm S}^\gamma})v_t^\gamma=0.
\label{eq:sde_gradient_preservation}
\end{equation}
Consequently, the total projected error is
\begin{equation}
o_{t,\mathrm S}^\gamma
:=P_{\rho_{t,\mathrm S}^\gamma}e_t^\gamma
=v_t^\gamma+g_{t,\mathrm S}^\gamma.
\label{eq:sde_observable_error}
\end{equation}
\end{proposition}

For $t>0$, positivity and smoothness of the reference
densities $p_t^b$ follow from Gaussian noising whenever
$\bar\sigma_t>0$. Regularity of $\rho_{t,\mathrm S}^\gamma$
and square-integrability of the errors under that density
must also be verified. For example, the latter follows if
$|v_t^b(x)|+|w_t^b(x)|\leq C(1+|x|^m)$ for some $m\geq0$
and $\rho_{t,\mathrm S}^\gamma$ has a finite $2m$-th moment.

\paragraph{Linearized marginal dynamics.}
We now connect the activation residual to the first-order
change in the density generated by reverse-SDE sampling.
Fix $\gamma$ and replace the exact score  $s_t^\gamma$ by its perturbation $s_t^\gamma + \varepsilon e_t^\gamma$. The resulting sampler is
\begin{equation*} 
\begin{aligned}
\mathrm dY_t^{\gamma,\varepsilon} &= \left[\xi_t^\gamma(Y_t^{\gamma,\varepsilon}) -\varepsilon\sigma_t^2e_t^\gamma(Y_t^{\gamma,\varepsilon}) \right]\mathrm dt + \sigma(t)\,\mathrm d\widetilde B_t,\\
Y_T^{\gamma,\varepsilon}&\sim\pi, \qquad t:T\longrightarrow0.
\end{aligned}
\end{equation*}
The choices $\varepsilon=0$ and $\varepsilon=1$ recover
the exact-score and learned-score drift fields, respectively.
Let $\rho_{t,\mathrm S}^{\gamma,\varepsilon}$ denote the
perturbed marginal density, with
\begin{equation*}
\rho_{t,\mathrm S}^{\gamma,0}=\rho_{t,\mathrm S}^\gamma, \qquad
\rho_{T,\mathrm S}^{\gamma,\varepsilon}=\pi.
\end{equation*}

Assume additionally that the perturbed SDEs are well posed for $\varepsilon$ near zero, their densities are differentiable in $\varepsilon$ at zero, and differentiation is valid in the weak Fokker--Planck equation. 
Fokker–Planck equation for the reverse SDE driven by the perturbed score is
\begin{equation}\label{eq:sde_perturbed_fokker_planck}
\partial_t\rho_{t,\mathrm S}^{\gamma,\varepsilon}
 + \nabla\cdot\!\left( \xi_t^\gamma\rho_{t,\mathrm S}^{\gamma,\varepsilon} \right) =-\kappa_t\Delta\rho_{t,\mathrm S}^{\gamma,\varepsilon} + 2\varepsilon\kappa_t\nabla\cdot\!\left(
\rho_{t,\mathrm S}^{\gamma,\varepsilon} e_t^\gamma\right),
\end{equation}
where the negative diffusion sign reflects the noise-time convention $t:T\to0$.
Define the first-order density perturbation by
\begin{equation*}
\eta_{t,\mathrm S}^\gamma :=\left.\partial_\varepsilon
\rho_{t,\mathrm S}^{\gamma,\varepsilon}
\right|_{\varepsilon=0}.
\end{equation*}
Differentiating Eq.~\eqref{eq:sde_perturbed_fokker_planck}
at $\varepsilon=0$ gives the linearized equation (in the weak sense)
\begin{equation}\label{eq:sde_linearized_density_response}
\partial_t\eta_{t,\mathrm S}^\gamma
+\nabla\cdot\!\left(\xi_t^\gamma\eta_{t,\mathrm S}^\gamma\right) =-\kappa_t\Delta\eta_{t,\mathrm S}^\gamma +2\kappa_t\nabla\cdot\!\left(
\rho_{t,\mathrm S}^\gamma e_t^\gamma\right), \quad \eta_{T,\mathrm S}^\gamma =0.
\end{equation}
The terminal condition follows because the starting density $\pi$ does not depend
on $\varepsilon$. For small $\varepsilon$,
$\rho_{t,\mathrm S}^{\gamma,\varepsilon}
-\rho_{t,\mathrm S}^\gamma
\approx\varepsilon\eta_{t,\mathrm S}^\gamma$ in the
corresponding weak sense. The left-hand side and diffusion term describe how the SDE propagates this density perturbation. The remaining term is the force produced by the score error.

Using $e_t^\gamma=v_t^\gamma+w_t^\gamma$, the force is decomposed into inherited and activated contributions:
\begin{equation*} 
2\kappa_t\nabla\cdot
(\rho_{t,\mathrm S}^\gamma e_t^\gamma) = 2\kappa_t\nabla\cdot
(\rho_{t,\mathrm S}^\gamma v_t^\gamma) - 2\kappa_t\rho_{t,\mathrm S}^\gamma
q_{t,\mathrm S}^\gamma.
\end{equation*}
In particular, the contribution from the combined
originally weighted-solenoidal field is
\begin{equation}
\sigma_t^2\nabla\cdot
(\rho_{t,\mathrm S}^\gamma w_t^\gamma)
=-\sigma_t^2\rho_{t,\mathrm S}^\gamma q_{t,\mathrm S}^\gamma.
\label{eq:sde_activation_source}
\end{equation}
Thus the same activation residual governs the
weighted-solenoidal contribution to the linearized SDE
marginal dynamics. Compared with the ODE linearization in \eqref{eq:leakage_density_source},
Eq.~\eqref{eq:sde_linearized_density_response} includes
the diffusion term and uses the score-error coefficient
$\sigma_t^2=2\kappa_t$.

\section{How guidance strength changes error activation}
\label{sec:leakage_evolution}

Fix a noise time $t$ and let $I\subset\mathbb R$ be an open interval of guidance strengths. We study how the activated error $g_t^\gamma$ changes as $\gamma$ varies in $I$. Recall that
\begin{equation*}
g_t^\gamma = P_{\rho_t^\gamma} w_t^\gamma, \qquad
w_t^\gamma=\gamma w_t^c+(1-\gamma)w_t^u.
\end{equation*}
Changing $\gamma$ changes both the combination of conditional and unconditional errors $w_t^\gamma$ and the guided density used to define their projection $P_{\rho_t^\gamma}$. We separate these effects by comparing the fields at the same spatial locations and taking difference quotients of the projection identity. We first state the results for the ODE density $\rho_t^\gamma$ and then extend them to the reverse-SDE density $\rho_{t,\mathrm S}^\gamma$.

\subsection{Results for the reverse-time ODE sampling}

\paragraph{The projection identity.}
Because $g_t^\gamma$ is the gradient projection of
$w_t^\gamma$, their difference is orthogonal to all gradient
test fields:
\begin{equation}
\int_{\mathbb R^d}(g_t^\gamma-w_t^\gamma)
\cdot\nabla\psi\,\rho_t^\gamma\,dx=0,
\qquad \psi\in C_c^\infty(\mathbb R^d).
\label{eq:guidance_projection_orthogonality}
\end{equation}
The projection identity uses the same compactly supported test functions at every guidance strength, allowing direct comparisons across $\gamma$. Under the regularity assumptions below, we use this identity to establish continuity of the activated error and decompose its weak rate of change into
contributions from changes in the combined error $w_t^\gamma$ and the guided density.

\paragraph{Regularity assumptions along the guidance path.}
We first give conditions on the density and reference
error fields that imply continuity of the activated
field. These conditions will be used when we differentiate the projection identity to characterize how $g_t^\gamma$ varies with respect to $\gamma$.

\vspace{3mm}
\begin{assumption}[Regularity along the guidance path]
\label{ass:guidance_path}
For a fixed noise time $t$ and interval $I$, assume:\\
(i) \textit{Local regularity of the densities.}
For every $\gamma\in I$, the probability density
$\rho_t^\gamma$ is strictly positive and $C^1$.
It is continuous with respect to \(\gamma\), uniformly on compact spatial regions,
\[
\lim_{h\to0}\sup_{x\in K}
|\rho_t^{\gamma+h}(x)-\rho_t^\gamma(x)|=0 \quad\text{for every compact } K\subset\mathbb R^d.
\]
(ii) \textit{Differentiability of the densities.}
For every $\gamma\in I$, there is a scalar function
$\partial_\gamma\rho_t^\gamma$, square-integrable on every
compact spatial region, such that
\begin{equation}\label{eq:guidance_density_derivative}
\lim_{h\to0}\int_K \left| \frac{\rho_t^{\gamma+h}(x)-\rho_t^\gamma(x)}{h} -\partial_\gamma\rho_t^\gamma(x)
\right|^2dx=0 \quad\text{for every compact } K\subset\mathbb R^d.
\end{equation}
(iii) \textit{Integrability of the error components $w_t^b$.}
The fields $w_t^c$ and $w_t^u$ are measurable. For every compact interval $J\subset I$, there is a nonnegative function $H_{t,J}\in L^1(\mathbb R^d)$ such that
\begin{equation}\label{eq:guidance_error_envelope}
\rho_t^\eta(x)
\bigl(|w_t^c(x)|^2+|w_t^u(x)|^2\bigr)
\leq H_{t,J}(x)
\quad\text{for all }\eta\in J \text{ and almost every } x\in\mathbb R^d .
\end{equation}
(iv) \textit{Boundedness of the density contribution.}
For each $\gamma\in I$, define
\begin{equation}
\mathcal D_{t,\gamma}(\psi)
:=\int_{\mathbb R^d}(w_t^\gamma-g_t^\gamma)
\cdot\nabla\psi\,\partial_\gamma\rho_t^\gamma\,dx,
\qquad \psi\in C_c^\infty(\mathbb R^d).
\label{eq:measure_functional}
\end{equation}
There is a finite constant $C_{t,\gamma}$ such that
\begin{equation}\label{eq:bounded-D}
|\mathcal D_{t,\gamma}(\psi)|
\leq C_{t,\gamma}\|\nabla\psi\|_{L^2(\rho_t^\gamma)}
\quad\text{for every }\psi\in C_c^\infty(\mathbb R^d).
\end{equation}
Thus $\mathcal D_{t,\gamma}$ extends uniquely to an element of $\dot H^{-1}(\rho_t^\gamma)$.
\end{assumption}

Condition~(i) holds, for example, when the positive densities are jointly continuous in $(\gamma,x)$ and $C^1$ in $x$.
Condition~(iii) controls the weighted errors throughout space, including their tails. It implies
$w_t^c,w_t^u\in L^2(\rho_t^\eta)$ for every $\eta\in I$. 
Together, these two conditions imply continuity of the
activated field, as shown below.
Conditions~(ii) and (iv) are used to characterize the first-order variation of the activated error $g_t^\gamma$ with respect to $\gamma$.

The field $w_t^\gamma-g_t^\gamma$ is orthogonal to gradients under $\rho_t^\gamma$. The functional $\mathcal D_{t,\gamma}$ measures how its orthogonality to gradients changes when the density $\rho_t^\gamma$ varies. A sufficient condition for its boundedness in \eqref{eq:bounded-D} is given at the end of this section.
A weak derivative of $g_t^\gamma$ will be obtained from the projection identity, and the functional \(\mathcal D_{t,\gamma}\) contributes to that characterization:
\[
\langle G_t^\gamma,\nabla\psi\rangle_{\rho_t^\gamma} =
\langle w_t^c-w_t^u,\nabla\psi\rangle_{\rho_t^\gamma} +\mathcal D_{t,\gamma}(\psi).
\]

\paragraph{Continuity of the activated error.}
The next lemma derives continuity of $g_t^\gamma$ from the assumptions on the density and reference errors. The conclusion also ensures continuity of the activation energy. Recall the activation energy defined earlier:
\(
\mathcal L_t(\gamma) =\|g_t^\gamma\|_{L^2(\rho_t^\gamma)}^2 .
\)

\vspace{3mm}
\begin{lemma}[Continuity of the activated error]
\label{lem:guidance_projection_continuity}
Under conditions~(i) and (iii) of
Assumption~\ref{ass:guidance_path},
\begin{equation}
\left\|
\sqrt{\rho_t^{\gamma+h}}\,g_t^{\gamma+h}
-\sqrt{\rho_t^\gamma}\,g_t^\gamma
\right\|_{L^2(dx)}\longrightarrow0
\qquad\text{as }h\to0.
\label{eq:weighted_activation_continuity}
\end{equation}
Here $L^2(dx)$ denotes the unweighted space of vector
fields. Consequently, for every compact
$K\subset\mathbb R^d$,
\begin{equation}
\lim_{h\to0}\int_K
|g_t^{\gamma+h}(x)-g_t^\gamma(x)|^2\,dx=0.
\label{eq:local_activation_continuity}
\end{equation}
Moreover, $\mathcal L_t(\gamma+h)\to\mathcal L_t(\gamma)$.
\end{lemma}

The envelope in condition~(iii) is a sufficient condition for the weighted input continuity
\[
\sqrt{\rho_t^{\gamma+h}}\,w_t^{\gamma+h} \overset{h \rightarrow 0}{\xrightarrow{\hspace{1cm}}}
\sqrt{\rho_t^\gamma}\,w_t^\gamma
\quad\text{in }L^2(dx).
\]
The lemma remains valid if condition~(iii) is replaced
by this continuity together with
$w_t^c,w_t^u\in L^2(\rho_t^\eta)$ for every $\eta\in I$.
The envelope provides a concrete way to verify the
condition using the reference errors.

\paragraph{The first-order variation at fixed spatial locations.}
Lemma~\ref{lem:guidance_projection_continuity} supplies the local continuity needed to pass to the limit in difference quotients of the projection identity.
We now use conditions~(ii) and (iv) to characterize the first-order change in $g_t^\gamma$.

We describe the change in $g_t^\gamma$ through its pairings with gradient test fields. Specifically, we seek a gradient field $G_t^\gamma\in\mathcal G_{\rho_t^\gamma}$ satisfying
\begin{equation}
\langle G_t^\gamma,\nabla\psi\rangle_{\rho_t^\gamma}
=\lim_{h\to0}\int_{\mathbb R^d}
\frac{g_t^{\gamma+h}-g_t^\gamma}{h}
\cdot\nabla\psi\,\rho_t^\gamma\,dx,
\qquad \psi\in C_c^\infty(\mathbb R^d).
\label{eq:weak_eulerian_response}
\end{equation}
Here $g_t^{\gamma+h}(x)$ and $g_t^\gamma(x)$ are evaluated at the same location $x$, and the density is held fixed at $\rho_t^\gamma$. We call $G_t^\gamma$ the \emph{weak Eulerian rate of change}, or weak first-order variation at fixed spatial coordinates. The next theorem establishes the existence of the limit and its continuous extension to $\dot H^1(\rho_t^\gamma)$.
Differentiability here refers to these gradient-test
pairings, and differentiability of the full field in
$L^2(\rho_t^\gamma)$ is a stronger property.

\vspace{3mm}
\begin{theorem}[Decomposition of the activation rate of change]
\label{thm:two_channel}
Under Assumption~\ref{ass:guidance_path}, the limit in
Eq.~\eqref{eq:weak_eulerian_response} exists for every
$\gamma\in I$ and every compactly supported smooth test function. The resulting functional extends continuously to $\dot H^1(\rho_t^\gamma)$ and is represented by a unique gradient field $G_t^\gamma\in\mathcal G_{\rho_t^\gamma}$.
For every $\psi\in\dot H^1(\rho_t^\gamma)$, this continuous extension satisfies
\begin{equation}
\langle G_t^\gamma,\nabla\psi\rangle_{\rho_t^\gamma}
=\langle\partial_\gamma w_t^\gamma,\nabla\psi
\rangle_{\rho_t^\gamma}
+\mathcal D_{t,\gamma}(\psi).
\label{eq:two_channel_weak}
\end{equation}
Let $D_t^\gamma\in\mathcal G_{\rho_t^\gamma}$ be the
unique gradient field representing the density contribution:
\begin{equation}
\langle D_t^\gamma,\nabla\psi\rangle_{\rho_t^\gamma}
=\mathcal D_{t,\gamma}(\psi),
\qquad \psi\in\dot H^1(\rho_t^\gamma),
\label{eq:measure_channel}
\end{equation}
and define
\begin{equation}
M_t^\gamma:=P_{\rho_t^\gamma}
(\partial_\gamma w_t^\gamma).
\label{eq:branch_channel_general}
\end{equation}
Then
\begin{equation}
G_t^\gamma=M_t^\gamma+D_t^\gamma.
\label{eq:two_channel_response}
\end{equation}
For linear CFG, $\partial_\gamma w_t^\gamma=w_t^c-w_t^u$,
so
\begin{equation}
M_t^\gamma=P_{\rho_t^\gamma}(w_t^c-w_t^u).
\label{eq:branch_channel}
\end{equation}
Moreover,
\begin{equation}
\|G_t^\gamma\|_{L^2(\rho_t^\gamma)}
\leq\|\partial_\gamma w_t^\gamma\|_{L^2(\rho_t^\gamma)}
+\|\mathcal D_{t,\gamma}\|_{\dot H^{-1}(\rho_t^\gamma)}.
\label{eq:two_channel_bound}
\end{equation}
\end{theorem}

The term $M_t^\gamma$ is the contribution from changing the error combination: it varies the weights of $w_t^c$ and $w_t^u$ while keeping the guided density fixed.
The term $D_t^\gamma$ is the contribution from changing the reference density: it varies the projection while holding the current error field fixed. Although the testing density in Eq.~\eqref{eq:weak_eulerian_response} is fixed, $g_t^{\gamma+h}$ is computed using $\rho_t^{\gamma+h}$, so the rate of change includes both effects.

\vspace{3mm}
\begin{remark}[A sufficient condition for the density contribution]
The boundedness requirement on $\mathcal D_{t,\gamma}$
is satisfied if
\begin{equation*}
(w_t^\gamma-g_t^\gamma)
\frac{\partial_\gamma\rho_t^\gamma}{\rho_t^\gamma}
\in L^2(\rho_t^\gamma;\mathbb R^d).
\end{equation*}
Indeed, writing Eq.~\eqref{eq:measure_functional} as a
$\rho_t^\gamma$-weighted pairing and applying the
Cauchy--Schwarz inequality gives
\[
\|\mathcal D_{t,\gamma}\|_{\dot H^{-1}(\rho_t^\gamma)}
\leq\left\|
(w_t^\gamma-g_t^\gamma)
\frac{\partial_\gamma\rho_t^\gamma}{\rho_t^\gamma}
\right\|_{L^2(\rho_t^\gamma)}.
\]
In this case, the density contribution has the explicit projection representation
\begin{equation}
D_t^\gamma=P_{\rho_t^\gamma}\left[
(w_t^\gamma-g_t^\gamma)
\frac{\partial_\gamma\rho_t^\gamma}{\rho_t^\gamma}
\right].
\label{eq:density_response_projection}
\end{equation}
This condition controls the product of the currently
weighted-solenoidal component and the relative density
variation. It allows that variation to grow in space,
provided the product remains square-integrable under
the guided density.
\end{remark}

\subsection{Extending to reverse-time SDE sampling}
The preceding arguments rely on the weighted projection identity and the stated regularity assumptions. They therefore also apply to the reverse SDE in Section~\ref{sec:reverse_sde_activation}, provided these assumptions hold with $\rho_{t,\mathrm S}^\gamma$ as the weighting density.
Recall that
\[
g_{t,\mathrm S}^\gamma
=P_{\rho_{t,\mathrm S}^\gamma}w_t^\gamma,
\qquad
\mathcal L_{t,\mathrm S}(\gamma)
=\|g_{t,\mathrm S}^\gamma\|_{L^2(\rho_{t,\mathrm S}^\gamma)}^2.
\]
For the SDE, their combination $w_t^\gamma$ is projected under $\rho_{t,\mathrm S}^\gamma$. To describe how changes in this density affect the projection, define
\begin{equation} \label{eq:sde_guidance_density_functional}
\mathcal D_{t,\gamma,\mathrm S}(\psi)
:=\int_{\mathbb R^d}
(w_t^\gamma-g_{t,\mathrm S}^\gamma)\cdot\nabla\psi\,
\partial_\gamma\rho_{t,\mathrm S}^\gamma\,dx,
\qquad \psi\in C_c^\infty(\mathbb R^d).
\end{equation}
The residual $w_t^\gamma-g_{t,\mathrm S}^\gamma$ is
orthogonal to gradients under $\rho_{t,\mathrm S}^\gamma$.
The functional $\mathcal D_{t,\gamma,\mathrm S}$ measures how these pairings change as the weighting density varies with $\gamma$, while holding the residual field fixed.
For the following result, impose Assumption~\ref{ass:guidance_path} on the SDE density family: replace $\rho_t^\eta$, $g_t^\eta$, and
$\mathcal D_{t,\gamma}$ by $\rho_{t,\mathrm S}^\eta$,
$g_{t,\mathrm S}^\eta$, and $\mathcal D_{t,\gamma,\mathrm S}$, respectively, for $\eta\in I$.

\vspace{3mm}
\begin{corollary}[Variation along different guidance strengths in reverse-SDE sampling]
\label{cor:sde_guidance_variation}
Fix $t\in(0,T)$ and an open guidance interval $I$ on which
the exact-score guided SDEs are well posed. Under the SDE
versions of conditions~(i) and (iii) of
Assumption~\ref{ass:guidance_path},
\begin{equation}
\left\|
\sqrt{\rho_{t,\mathrm S}^{\gamma+h}}\,
g_{t,\mathrm S}^{\gamma+h}
-\sqrt{\rho_{t,\mathrm S}^\gamma}\,
g_{t,\mathrm S}^\gamma
\right\|_{L^2(dx)}\longrightarrow0
\quad\text{as }h\to0.
\label{eq:sde_weighted_activation_continuity}
\end{equation}
Consequently, $g_{t,\mathrm S}^{\gamma+h}\to
g_{t,\mathrm S}^\gamma$ in local $L^2$, and
$\mathcal L_{t,\mathrm S}(\gamma+h)\to
\mathcal L_{t,\mathrm S}(\gamma)$.

If all four conditions hold, there is a unique gradient
field $G_{t,\mathrm S}^\gamma\in
\mathcal G_{\rho_{t,\mathrm S}^\gamma}$ such that, for
every $\psi\in C_c^\infty(\mathbb R^d)$,
\begin{equation}
\begin{aligned}
\langle G_{t,\mathrm S}^\gamma,\nabla\psi
\rangle_{\rho_{t,\mathrm S}^\gamma}
&=\lim_{h\to0}\int_{\mathbb R^d}
\frac{g_{t,\mathrm S}^{\gamma+h}-g_{t,\mathrm S}^\gamma}{h}
\cdot\nabla\psi\,\rho_{t,\mathrm S}^\gamma\,dx\\
&=\langle w_t^c-w_t^u,\nabla\psi
\rangle_{\rho_{t,\mathrm S}^\gamma}
+ \mathcal D_{t,\gamma,\mathrm S}(\psi),
\end{aligned}
\label{eq:sde_guidance_rate_weak}
\end{equation}
where $\mathcal D_{t,\gamma,\mathrm S}(\psi)$ is defined in \eqref{eq:sde_guidance_density_functional}.
The final expression extends continuously to
$\dot H^1(\rho_{t,\mathrm S}^\gamma)$. In particular,
\begin{equation}
G_{t,\mathrm S}^\gamma=M_{t,\mathrm S}^\gamma
+D_{t,\mathrm S}^\gamma,
\qquad
M_{t,\mathrm S}^\gamma
:=P_{\rho_{t,\mathrm S}^\gamma}(w_t^c-w_t^u),
\label{eq:sde_guidance_rate_decomposition}
\end{equation}
and $D_{t,\mathrm S}^\gamma\in
\mathcal G_{\rho_{t,\mathrm S}^\gamma}$ is characterized by
\begin{equation}
\langle D_{t,\mathrm S}^\gamma,\nabla\psi
\rangle_{\rho_{t,\mathrm S}^\gamma}
=\mathcal D_{t,\gamma,\mathrm S}(\psi),
\qquad\psi\in\dot H^1(\rho_{t,\mathrm S}^\gamma).
\label{eq:sde_guidance_density_component}
\end{equation}
Moreover,
\begin{equation}
\|G_{t,\mathrm S}^\gamma\|_{L^2(\rho_{t,\mathrm S}^\gamma)}
\leq\|w_t^c-w_t^u\|_{L^2(\rho_{t,\mathrm S}^\gamma)}
+\|\mathcal D_{t,\gamma,\mathrm S}\|_{
\dot H^{-1}(\rho_{t,\mathrm S}^\gamma)}.
\label{eq:sde_guidance_rate_bound}
\end{equation}
\end{corollary}

The proof is given in Appendix~\ref{app:variation_with_strength}.
As in the ODE case, $M_{t,\mathrm S}^\gamma$ describes the
changing error combination, while $D_{t,\mathrm S}^\gamma$
describes the changing guided density. A sufficient
condition for the latter functional to be bounded is
\begin{equation}
(w_t^\gamma-g_{t,\mathrm S}^\gamma)
\frac{\partial_\gamma\rho_{t,\mathrm S}^\gamma}
{\rho_{t,\mathrm S}^\gamma}
\in L^2(\rho_{t,\mathrm S}^\gamma;\mathbb R^d).
\label{eq:sde_density_variation_integrability}
\end{equation}
Under this condition, $D_{t,\mathrm S}^\gamma$ has the explicit projection representation
\begin{equation*}
D_{t,\mathrm S}^\gamma=P_{\rho_{t,\mathrm S}^\gamma}\left[ (w_t^\gamma-g_{t,\mathrm S}^\gamma)
\frac{\partial_\gamma\rho_{t,\mathrm S}^\gamma}{\rho_{t,\mathrm S}^\gamma}
\right].
\end{equation*}
The guided density
$\rho_{t,\mathrm S}^\gamma$ defines the projection
$P_{\rho_{t,\mathrm S}^\gamma}$ and hence $M_{t,\mathrm S}^\gamma$, while its derivative
$\partial_\gamma\rho_{t,\mathrm S}^\gamma$ appears in
the density contribution $\mathcal D_{t,\gamma,\mathrm S}$.

For both samplers, these results describe how the activated
field varies with $\gamma$ at a fixed noise time.
Section~\ref{sec:activation_dominance} asks whether, over the
full ODE sampling interval, activated error can contribute
more than inherited error to the KL divergence between the
learned-score and exact-score endpoint densities.

\section{Density representations for learned CFG}
\label{sec:analytic_learned_cfg}
We derive the density generated by learned-score CFG
through two comparisons. First, we express $\hat\rho_0^\gamma$ relative to the powered-product function $R_0^\gamma$ (defined in \eqref{eq:powered-product-function}), making score estimation errors explicit in a characteristic formula. We then use the exact-score guided density $\rho_0^\gamma$ as the baseline to identify how inherited and activated errors affect the difference between $\hat\rho_0^\gamma$ and $\rho_0^\gamma$.
For each fixed $\gamma$ and finite score errors, we derive
these comparisons using ODE characteristics
(Section~\ref{sec:ode_density_representations}) and expectations over auxiliary diffusion paths for the reverse SDE (Section~\ref{sec:sde_density_representations}). Section~\ref{sec:activation_dominance} applies the ODE decomposition to a finite Gaussian mixture in which the activated contribution to endpoint KL eventually dominates the inherited contribution as the guidance strength increases.

\subsection{Density representations for ODE sampling}
\label{sec:ode_density_representations}

\paragraph{Sampling trajectories and regularity.}
The learned guided ODE has velocity
\begin{equation*}
\hat\zeta_t^\gamma(x)
=f(t)x-\kappa_t\bigl(s_t^\gamma(x)+e_t^\gamma(x)\bigr)
=\zeta_t^\gamma(x)-\kappa_t e_t^\gamma(x).
\end{equation*}
For $0\leq t\leq u\leq T$, let
$\hat\Phi_{t,u}^\gamma(x)$ denote the position at
time $u$ on the learned trajectory that passes through
$x$ at time $t$. Thus,
\[
\partial_u \hat\Phi_{t,u}^\gamma(x) = \hat\zeta_u^\gamma
\bigl(\hat\Phi_{t,u}^\gamma(x)\bigr), \quad \hat\Phi_{t,t}^\gamma(x)=x.
\]
Since backward sampling runs from \(t=T\) to \(t=0\), the curve
\[
u\longmapsto\hat\Phi_{0,u}^\gamma(x), \qquad 0\leq u\leq T,
\]
retraces the sampling trajectory from the generated sample \(x\) at time \(0\) to its initial noisy state \(\hat\Phi_{0,T}^\gamma(x)\) at time \(T\).

We impose the following classical regularity conditions for probability flow.
\begin{enumerate}
\item \emph{Reference dynamics.} The coefficients $f$ and
$\kappa$ are continuous. The reference densities $p_t^c$
and $p_t^u$ are strictly positive, $C^1$ in time, and
$C^2$ in space, and satisfy the reference continuity
equations from Section~2.
\item \emph{Score errors.} For each $b\in\{c,u\}$,
$e_t^b(x)$ and its first spatial derivatives are
continuous in $(t,x)$.
\item \emph{Guided trajectories.} The ODEs with velocities
$\zeta_t^\gamma$ and $\hat\zeta_t^\gamma$ admit unique
trajectories throughout $[0,T]$ through each space-time
point under consideration.
\item \emph{Guided marginal densities.} The densities
$\rho_t^\gamma(x)$ and $\hat\rho_t^\gamma(x)$ are
strictly positive and $C^1$ in $(t,x)$.
\item \emph{Common starting distribution.} The two guided
flows share the starting density
$\rho_T^\gamma=\hat\rho_T^\gamma=\pi$ at time $T$.
\end{enumerate}
These conditions justify the divergence calculations and
the chain rule along trajectories. Proofs are given in
Appendix~\ref{app:learned-CFG-generated-density}.

\paragraph{Comparison with the powered-product function.}
Define
\begin{equation}\label{eq:powered-product-function}
    H_t:=\log\frac{p_t^c}{p_t^u}, \qquad
R_t^\gamma:=(p_t^c)^\gamma(p_t^u)^{1-\gamma}.
\end{equation}
Then $\nabla H_t=s_t^c-s_t^u$ and $\nabla\log R_t^\gamma=s_t^\gamma$. This score identity motivates the heuristic that CFG samples from the normalized powered product. However, in general, exact-score CFG generate a different distribution other than normalized $R_t^\gamma$ \citep{bradley2025classifier}.

\citet{jiang2026analytic} derive a path-integral representation of the density generated by the exact-score guided ODE. We use a similar characteristic argument with the score estimation errors included. The next theorem expresses $\hat\rho_0^\gamma$ as $R_0^\gamma$ multiplied by a starting-density ratio and an exponential correction accumulated along the learned trajectory.

\vspace{3mm}
\begin{theorem}[Characteristic density formula for learned CFG]
\label{thm:analytic_learned_cfg}
Under the conditions above, the density generated by CFG with learned scores $\hat s_t^c$ and $\hat s_t^u$ satisfies
\begin{equation}\label{eq:sec5_analytic_learned_distribution}
\begin{aligned}
\hat\rho_0^\gamma(x)
&=R_0^\gamma(x)
\frac{\pi\bigl(\hat\Phi_{0,T}^\gamma(x)\bigr)}
{R_T^\gamma\bigl(\hat\Phi_{0,T}^\gamma(x)\bigr)}\\
&\quad\times\exp\!\Bigg[
-\int_0^T\kappa_t
\Big\{\gamma(\gamma-1)|\nabla H_t|^2
+\nabla\cdot e_t^\gamma+e_t^\gamma\cdot s_t^\gamma\Big\}
\bigl(\hat\Phi_{0,t}^\gamma(x)\bigr)\,dt
\Bigg].
\end{aligned}
\end{equation}
\end{theorem}

The starting-density ratio accounts for the difference between the sampler's starting density $\pi$ and $R_T^\gamma$ at a finite noising horizon. It equals one when $p_T^c=p_T^u=\pi$. The full expression gives the normalized generated density without a separate normalization of $R_0^\gamma$.

At each time $t$, the scalar expression in braces is evaluated at the point $\hat\Phi_{0,t}^\gamma(x)$ on the learned trajectory. When $e_t^\gamma=0$ throughout sampling, this trajectory coincides with the exact-score trajectory $\Phi_{0,t}^\gamma(x)$, and $\gamma(\gamma-1)|\nabla H_t|^2$ is the only term in the integral.
This recovers the exact-score formula~\citep{jiang2026analytic} with the same starting density. Learned scores introduce the terms $\nabla\cdot e_t^\gamma+e_t^\gamma\cdot s_t^\gamma$ and can also
change the path along which all terms are evaluated.

\paragraph{Comparison with the exact-score guided density.}
The comparison with $R_0^\gamma$ includes both the effect of guidance with exact scores and the effect of score estimation error. To isolate the change due to score estimation, we now use the exact-score guided density $\rho_0^\gamma$ as the baseline. The error decomposition from Section~3 then identifies how $w_t^\gamma$ enters the ratio $\hat\rho_0^\gamma/\rho_0^\gamma$.

Recall that $e_t^\gamma=v_t^\gamma+w_t^\gamma$ and that
the activation residual in
Theorem~\ref{thm:fixed_guidance} is
\begin{equation*}
q_t^\gamma = -\gamma w_t^c\cdot\nabla\log\frac{\rho_t^\gamma}{p_t^c}
-(1-\gamma)w_t^u\cdot\nabla\log\frac{\rho_t^\gamma}{p_t^u}.
\end{equation*}
It satisfies
$q_t^\gamma=Q_{\rho_t^\gamma}w_t^\gamma = Q_{\rho_t^\gamma}g_t^\gamma$.
Thus the combined weighted-solenoidal error and its
activated gradient component have the same weighted
divergence under $\rho_t^\gamma$. The next theorem combines
this identity with the two guided continuity equations to separate
the inherited and activated contributions to the endpoint
log-density ratio.

\vspace{3mm}
\begin{theorem}[Inherited and activated contributions to the density ratio]
\label{thm:sec5_native_leakage_distribution}
Under the conditions of
Theorem~\ref{thm:analytic_learned_cfg}, suppose additionally that Assumption~\ref{ass:fixed_guidance} holds along the noise schedule, that $v_t^\gamma$ is $C^1$ in space, and that the displayed path integral is finite. Then
\begin{equation}\label{eq:sec5_native_leakage_density}
\hat\rho_0^\gamma(x) =\rho_0^\gamma(x) \times\exp\!\left[
\int_0^T\kappa_t
\Big\{-\nabla\cdot v_t^\gamma
-v_t^\gamma\cdot\nabla\log\rho_t^\gamma
+q_t^\gamma \Big\}
\bigl(\hat\Phi_{0,t}^\gamma(x)\bigr)\,dt
\right].
\end{equation}
\end{theorem}

In Eq.~\eqref{eq:sec5_native_leakage_density}, the inherited gradient error enters through
$Q_{\rho_t^\gamma}v_t^\gamma
=-\nabla\cdot v_t^\gamma-v_t^\gamma\cdot\nabla\log\rho_t^\gamma$.
Errors that are weighted-solenoidal under their respective
reference densities enter through $q_t^\gamma$. The two scalar
terms are evaluated along the same learned trajectory, whose
velocity depends on the full error $e_t^\gamma=v_t^\gamma+w_t^\gamma$.

\paragraph{Two terms in the activation residual.}
To relate the activation residual to the powered-product
function, we rewrite $q_t^\gamma$ using $R_t^\gamma$.
Substituting
$\nabla\log R_t^\gamma=s_t^\gamma$ and
$\nabla H_t=s_t^c-s_t^u$ into
Eq.~\eqref{eq:exact_leakage_source} gives
\begin{equation}
q_t^\gamma
=\gamma(1-\gamma)(w_t^c-w_t^u)\cdot\nabla H_t
-w_t^\gamma\cdot\nabla\log\frac{\rho_t^\gamma}{R_t^\gamma}.
\label{eq:sec5_leakage_two_sources}
\end{equation}
The first term depends on the alignment between the
error difference $w_t^c-w_t^u$ and the score difference
$s_t^c-s_t^u$.
The second depends on the alignment of $w_t^\gamma$
with $\nabla\log\rho_t^\gamma-s_t^\gamma$, the difference
between the score of the guided marginal and the combined
exact score.

\paragraph{Cumulative contributions to KL divergence.}
The same characteristic argument applies at any noise time
$t\in[0,T]$. For a trajectory passing through $x$ at time $t$,
define the accumulated inherited and activated contributions by
\begin{equation}
\begin{aligned}
C_I^{\gamma,t}(x)
&=-\int_t^T\kappa_u
\left(\nabla\cdot v_u^\gamma
+v_u^\gamma\cdot\nabla\log\rho_u^\gamma\right)
\bigl(\hat\Phi_{t,u}^\gamma(x)\bigr)\,\mathrm du,\\
C_A^{\gamma,t}(x)
&=\int_t^T\kappa_u q_u^\gamma
\bigl(\hat\Phi_{t,u}^\gamma(x)\bigr)\,\mathrm du.
\end{aligned}
\label{eq:reported_path_contributions}
\end{equation}
Since the two guided flows have the same starting density at $T$,
integration along this trajectory gives
\[
\log\frac{\hat\rho_t^\gamma(x)}{\rho_t^\gamma(x)}
=C_I^{\gamma,t}(x)+C_A^{\gamma,t}(x).
\]
When both path contributions are absolutely integrable under
$\hat\rho_t^\gamma$, denote their expectations by
\begin{equation*}
D_I^{\gamma,t}:=\mathbb E_{\hat\rho_t^\gamma}[C_I^{\gamma,t}],
\qquad
D_A^{\gamma,t}:=\mathbb E_{\hat\rho_t^\gamma}[C_A^{\gamma,t}].
\end{equation*}
Taking the expectation of the log-density ratio gives
\begin{equation}
D_I^{\gamma,t} + D_A^{\gamma,t} = \mathrm{KL}(\hat\rho_t^\gamma\|\rho_t^\gamma).
\label{eq:reported_kl_decomposition}
\end{equation}

At the endpoint, write $D_I^\gamma:=D_I^{\gamma,0}$ and
$D_A^\gamma:=D_A^{\gamma,0}$. For the finite Gaussian mixture and prescribed score errors in Section~\ref{sec:activation_dominance},
we show that
\begin{equation*}
\lim_{\gamma\to\infty}
\frac{D_A^\gamma}{\mathrm{KL}(\hat\rho_0^\gamma\|\rho_0^\gamma)}=1.
\end{equation*}
The simulations in Section~\ref{sec:simulations} support such results. These examples illustrate how activation can
account for much of the discrepancy between learned-score and exact-score CFG under strong guidance.

\subsection{Extension to reverse-time SDE sampling}
\label{sec:sde_density_representations}
Both density comparisons extend to reverse-SDE sampling through Feynman--Kac representations involving expectations over stochastic paths. The exact-score and learned-score guided densities satisfy Fokker-Planck equations, respectively, 
\begin{equation}
\begin{aligned}
\partial_t\rho_{t,\mathrm S}^\gamma
+\nabla\cdot(\xi_t^\gamma\rho_{t,\mathrm S}^\gamma)
&=-\kappa_t\Delta\rho_{t,\mathrm S}^\gamma,\\
\partial_t\hat\rho_{t,\mathrm S}^\gamma
+\nabla\cdot(\hat\xi_t^\gamma\hat\rho_{t,\mathrm S}^\gamma)
&=-\kappa_t\Delta\hat\rho_{t,\mathrm S}^\gamma,
\end{aligned}
\label{eq:sec5_sde_density_equations}
\end{equation}
where
\[
\xi_t^\gamma(x)=f(t)x-2\kappa_t s_t^\gamma(x),
\qquad
\hat\xi_t^\gamma=\xi_t^\gamma-2\kappa_t e_t^\gamma.
\]
Both samplers start from
$\rho_{T,\mathrm S}^\gamma
=\hat\rho_{T,\mathrm S}^\gamma=\pi$.

\paragraph{Auxiliary diffusions.}
To evaluate the generated density at $x$, we introduce two
auxiliary diffusion processes, both starting from $x$ at time $0$ and running toward $T$:
\begin{equation}
\begin{aligned}
\mathrm dZ_t^R
&=\bigl[f(t)Z_t^R-2\kappa_t e_t^\gamma(Z_t^R)\bigr]
\,\mathrm dt+\sigma(t)\,\mathrm dW_t,\\
\mathrm dZ_t^\rho
&=\bigl[\hat\xi_t^\gamma
+2\kappa_t\nabla\log\rho_{t,\mathrm S}^\gamma\bigr]
(Z_t^\rho)\,\mathrm dt+\sigma(t)\,\mathrm dW_t,\\
Z_0^R&=Z_0^\rho=x,\qquad t:0\longrightarrow T.
\end{aligned}
\label{eq:sec5_sde_auxiliary_diffusions}
\end{equation}
Here $W_t$ is standard Brownian motion, and
$\mathbb E_x^R$ and $\mathbb E_x^\rho$ denote expectations
under the respective path laws. These auxiliary drifts
arise from the density-ratio equations derived in
Appendix~\ref{app:sde_density_representations}.

\paragraph{Regularity and integrability conditions.}
We impose the following conditions for any fixed $\gamma$. 
\begin{enumerate}
\item \emph{Reference dynamics.}
\label{cond:sec5_sde_reference}
The coefficients $f$ and $\kappa$ are continuous, with
$\kappa_t=\sigma^2(t)/2$. The reference densities $p_t^c$
and $p_t^u$ are strictly positive and $C^{1,2}$ in $(t,x)$,
and satisfy
\[
\partial_t p_t^b
+\nabla\cdot\bigl(f(t)x\,p_t^b\bigr)
=\kappa_t\Delta p_t^b,
\qquad b\in\{c,u\}.
\]

\item \emph{Score errors.}
\label{cond:sec5_sde_errors}
For each $b\in\{c,u\}$, the error $e_t^b(x)$ and its
first spatial derivatives are continuous in $(t,x)$.

\item \emph{Guided SDEs.}
\label{cond:sec5_sde_wellposed}
The exact-score and learned-score guided reverse SDEs
admit unique non-explosive weak solutions throughout
the sampling interval $t:T\to0$.

\item \emph{Guided marginal densities.}
\label{cond:sec5_sde_marginals}
The marginal densities in
Eq.~\eqref{eq:sec5_sde_density_equations} are strictly
positive and $C^{1,2}$ in $(t,x)$: once continuously
differentiable in time and twice continuously
differentiable in space.

\item \emph{Auxiliary diffusions and Feynman--Kac integrability.}
\label{cond:sec5_sde_fk}
For every starting point $x$ under consideration, both
auxiliary diffusions in
Eq.~\eqref{eq:sec5_sde_auxiliary_diffusions} admit unique
non-explosive weak solutions on $[0,T]$, and the path
integrals in the density representations are finite
almost surely. For each process, the stopped,
exponentially weighted density ratios defined in
Appendix~\ref{app:sde_density_representations} form a
uniformly integrable family.
\end{enumerate}

The next proposition gives both density comparisons
and an additive KL identity that includes the effect
of diffusion.

\vspace{3mm}
\begin{proposition}[Density representations and KL decomposition
for reverse-SDE CFG]
\label{prop:sde_density_representations}
Assume conditions~\ref{cond:sec5_sde_reference}--
\ref{cond:sec5_sde_marginals} and the common starting
density stated above. Parts~(i)--(ii) additionally require
condition~\ref{cond:sec5_sde_fk}.

\textit{(i) Comparison with the powered product.}
The learned-score guided density satisfies
\begin{equation}
\begin{aligned}
\hat\rho_{0,\mathrm S}^\gamma(x)
&=R_0^\gamma(x)\,\mathbb E_x^R\!\Bigg[
\frac{\pi(Z_T^R)}{R_T^\gamma(Z_T^R)}\\
&\qquad\times\exp\!\Bigg(-\int_0^T\kappa_t
\Big\{\gamma(\gamma-1)|\nabla H_t|^2
+2\nabla\cdot e_t^\gamma
+2e_t^\gamma\cdot s_t^\gamma\Big\}(Z_t^R)\,\mathrm dt
\Bigg)\Bigg].
\end{aligned}
\label{eq:sec5_sde_powered_product_density}
\end{equation}

\textit{(ii) Comparison with the exact-score SDE density.}
Assume additionally that the hypotheses of
Proposition~\ref{prop:reverse_sde_activation} hold for
$t\in(0,T)$, and that $v_t^\gamma$ and its first spatial
derivatives are continuous in $(t,x)$. The weighted
divergences below then have scalar function
representations, and
\begin{equation}
\frac{\hat\rho_{0,\mathrm S}^\gamma(x)}
{\rho_{0,\mathrm S}^\gamma(x)} = \mathbb E_x^\rho\!\left[
\exp\!\left(\int_0^T2\kappa_t
\big\{Q_{\rho_{t,\mathrm S}^\gamma}v_t^\gamma
+q_{t,\mathrm S}^\gamma\big\}(Z_t^\rho)\,\mathrm dt
\right)\right],
\label{eq:sec5_sde_baseline_density}
\end{equation}
where
\[
Q_{\rho_{t,\mathrm S}^\gamma}v_t^\gamma =-\nabla\cdot v_t^\gamma -v_t^\gamma\cdot\nabla\log\rho_{t,\mathrm S}^\gamma, \qquad
q_{t,\mathrm S}^\gamma =Q_{\rho_{t,\mathrm S}^\gamma}w_t^\gamma.
\]

\textit{(iii) KL decomposition.}
Assume the additional projection regularity in part~(ii).
Suppose also that relative entropy is finite and
absolutely continuous in time, differentiation under
the integral is valid, integrations by parts produce
no boundary terms, and all displayed time integrals
are finite, with absolute integrability for the signed
terms. Then
\begin{equation}
\begin{aligned}
\mathrm{KL}(\hat\rho_{0,\mathrm S}^\gamma
\|\rho_{0,\mathrm S}^\gamma)
&=2\int_0^T\kappa_t\,
\mathbb E_{\hat\rho_{t,\mathrm S}^\gamma}
\big[Q_{\rho_{t,\mathrm S}^\gamma}v_t^\gamma\big]
\,\mathrm dt + 2\int_0^T\kappa_t\,
\mathbb E_{\hat\rho_{t,\mathrm S}^\gamma}
[q_{t,\mathrm S}^\gamma]\,\mathrm dt\\
&\quad-\int_0^T\kappa_t\,
\mathbb E_{\hat\rho_{t,\mathrm S}^\gamma}\!\left[
\left|\nabla\log
\frac{\hat\rho_{t,\mathrm S}^\gamma}
{\rho_{t,\mathrm S}^\gamma}\right|^2
\right]\,\mathrm dt.
\end{aligned}
\label{eq:sec5_sde_kl_decomposition}
\end{equation}
\end{proposition}

Setting $e_t^\gamma=0$ in
Eq.~\eqref{eq:sec5_sde_powered_product_density} recovers
the exact-score guided SDE density. The auxiliary
process $Z^R$ then follows the forward noising SDE,
and only the guidance term remains in the exponential.
The starting-density ratio equals one when
$p_T^c=p_T^u=\pi$; otherwise, it accounts for the
finite-horizon mismatch. No separate normalization
of $R_0^\gamma$ is required.

Equation~\eqref{eq:sec5_sde_baseline_density} identifies
the explicit contribution of activation through
$2\kappa_tq_{t,\mathrm S}^\gamma$. The errors also affect
the auxiliary path laws. Because the logarithm is taken
after the expectation, separating the exponent does
not generally yield separate additive inherited and
activated contributions to the endpoint log-density
ratio.

The activation residual also retains the algebraic
decomposition
\begin{equation*}
q_{t,\mathrm S}^\gamma =\gamma(1-\gamma)(w_t^c-w_t^u)\cdot\nabla H_t - w_t^\gamma\cdot\nabla\log
\frac{\rho_{t,\mathrm S}^\gamma}{R_t^\gamma},
\end{equation*}
which follows from Eq.~\eqref{eq:sde_change_density_identity} and
$\nabla\log R_t^\gamma=s_t^\gamma$.

In the KL identity~\eqref{eq:sec5_sde_kl_decomposition},
the first two terms are the inherited and activated
contributions. The final term is nonpositive and
represents relative-entropy dissipation by diffusion
in the sampling direction $t:T\to0$. The identity
does not by itself imply that activation increases KL
or that filtering reduces it. These results concern
the continuous-time reverse SDE underlying DDPM, while finite-step sampling introduces additional
discretization error. Proofs are given in
Appendix~\ref{app:sde_density_representations}.

\subsection{Dominance of activated error under strong guidance} \label{sec:activation_dominance}

For the guided ODE, the inherited and activated contributions
to endpoint KL are signed and can reinforce or cancel one
another. We give a finite Gaussian mixture in which the
activated contribution exceeds any fixed positive multiple of
the inherited contribution for all sufficiently large guidance
strengths. As in the Gaussian-mixture experiment in
Section~\ref{sec:gmm_experiment}, the target class is one
component of the unconditional distribution.

Throughout this subsection, the evaluation time is fixed
at $t=0$ and suppressed in the notation. Define
\begin{equation}
\begin{aligned}
D_I^\gamma
&:=\int_0^T\kappa_t\,
\mathbb E_{\hat\rho_t^\gamma}
\bigl[Q_{\rho_t^\gamma}v_t^\gamma\bigr]\,\mathrm dt,\\
D_A^\gamma
&:=\int_0^T\kappa_t\,
\mathbb E_{\hat\rho_t^\gamma}[q_t^\gamma]\,\mathrm dt.
\end{aligned}
\label{eq:dominance_cumulative_contributions}
\end{equation}
When both sources in Eq.~\eqref{eq:dominance_cumulative_contributions}
are absolutely integrable against $\kappa_t\hat\rho_t^\gamma(x)\,\mathrm dx\,\mathrm dt$,
Theorem~\ref{thm:sec5_native_leakage_distribution} gives
$D_I^\gamma+D_A^\gamma
=\mathrm{KL}(\hat\rho_0^\gamma\|\rho_0^\gamma)$.
If integration by parts is justified, writing
$r_t^\gamma:=\nabla\log(\hat\rho_t^\gamma/\rho_t^\gamma)$
also gives
\begin{equation}
\begin{aligned}
D_I^\gamma
&=\int_0^T\kappa_t\,
\mathbb E_{\hat\rho_t^\gamma}
[v_t^\gamma\cdot r_t^\gamma]\,\mathrm dt,\\
D_A^\gamma
&=\int_0^T\kappa_t\,
\mathbb E_{\hat\rho_t^\gamma}
[g_t^\gamma\cdot r_t^\gamma]\,\mathrm dt.
\end{aligned}
\label{eq:dominance_alignment_identities}
\end{equation}
Thus a large activated field is insufficient on its own:
its cumulative alignment with the density discrepancy
also matters. Appendix~\ref{app:general_activation_dominance}
gives a general sufficient condition and a guidance
threshold based on this identity. The factor
$\gamma(\gamma-1)$ in
Eq.~\eqref{eq:sec5_leakage_two_sources} alone does not
establish dominance, since the density-mismatch term can
offset it and the inherited source also depends on the
guided density. We showcase potential mechanism behind this phenomenon in the special case of Gaussian mixture. 

\paragraph{A finite Gaussian mixture.}
Let $d\geq2$ and $K\geq2$. Class $k$ has probability
$\pi_k>0$ and data distribution $\mathcal N(\mu_k,\bI_d)$
for $0\leq k<K$, with $\sum_{k=0}^{K-1}\pi_k=1$.
Take class $0$ as the target and set $\mu_0=0$.
For a finite horizon $T>0$, use the forward noising process
\[
\mathrm dX_t=-X_t\,\mathrm dt+\sqrt{2}\,\mathrm dB_t,
\qquad 0\leq t\leq T.
\]
Here $f(t)=-1$ and $\kappa_t=1$. At time $t$, class $k$
has mean $m_{k,t}=e^{-t}\mu_k$ and covariance $\bI_d$.
Writing $\varphi_d$ for the standard Gaussian density on
$\mathbb R^d$, the two reference densities are
\begin{equation*}
p_t^c(x)=\varphi_d(x),\qquad
p_t^u(x)=\sum_{k=0}^{K-1}\pi_k\varphi_d(x-m_{k,t}).
\end{equation*}
Write $E_0=\operatorname{span}\{\mu_1,\ldots,\mu_{K-1}\}$.
Assume that there are unit vectors $u\in E_0$ and $\vartheta\perp E_0$ with
\[ u\cdot\mu_k<0\quad(1\leq k<K). \]
All classes then have the same marginal along $\vartheta$, while the target mean lies a gap beyond the other means in direction $u$. The means can span up to $d-1$ dimensions. For example, if $\mu_1,\ldots,\mu_{K-1}$ are linearly independent and $K\leq d$,
one can choose $\vartheta$ perpendicular to their span and normalize a vector in their span whose inner product with each $\mu_k$ is $-1$.

Let $\boldsymbol{J}=\vartheta u^\top - u\vartheta^\top$, so $\boldsymbol{J} u=\vartheta$ and $\boldsymbol{J} \vartheta = -u$. Give both score estimates a common gradient bias and use the rotation
of each exact score as its weighted-solenoidal error by setting
\begin{equation*}
v_t^b = -c^\prime\vartheta,\qquad w_t^b(x)=-\omega_t \boldsymbol{J} s_t^b(x),\qquad
\omega_t=\omega e^{-2(T-t)},\qquad b\in\{c,u\},
\end{equation*}
where $c^\prime>0$ and $0<\omega\leq 1$. Indeed,
$p_t^b w_t^b=-\omega_t \boldsymbol{J}\nabla p_t^b$ has zero divergence by skew symmetry. For $\gamma>1$, the rotational score error deflects part of the guidance drift toward $-\vartheta$. If $\omega_t$ decays as sampling proceeds from $T$ to $0$, then guidance at each noise time makes a negative contribution to the endpoint mean along $\vartheta$.

\vspace{3mm}
\begin{proposition}[Eventual dominance for a finite Gaussian mixture]
\label{prop:finite_mixture_activation_dominance}
Under the mixture and error assumptions above, initialize both guided ODEs from $\pi=p_T^c$. Then $D_A^1=0$,
$D_I^\gamma>0$ for $\gamma\geq1$, and
\begin{equation}
\lim_{\gamma\to\infty}\frac{D_A^\gamma}{D_I^\gamma}=\infty,
\qquad
\lim_{\gamma\to\infty}
\frac{D_A^\gamma}{\mathrm{KL}(\hat\rho_0^\gamma\|\rho_0^\gamma)}=1.
\label{eq:finite_mixture_activation_share}
\end{equation}
In particular, for every $R>0$ there is a finite guidance
strength beyond which $D_A^\gamma>R D_I^\gamma$.
Appendix~\ref{app:finite_mixture_activation_dominance}
gives the proof and a sufficient threshold in terms of the
mixture parameters, the score errors, and $T$.
\end{proposition}

Under exact-score CFG, \(\vartheta^\top X_t\sim\mathcal N(0,1)\) at every noise time \(t\). Under learned-score CFG, the rotational error deflects part of the guidance drift along \(u\) toward \(-\vartheta\). Write \(\delta_t^\gamma:=-\mathbb E_{\hat\rho_t^\gamma}[\vartheta^\top X]\) for the displacement of the learned mean in that direction. Then
\[
\begin{aligned}
D_I^\gamma =c^\prime\int_0^T\delta_t^\gamma\,\mathrm dt,\quad
D_I^\gamma+D_A^\gamma =\mathrm{KL} (\hat\rho_0^\gamma\|\rho_0^\gamma)
\geq \tfrac12(\delta_0^\gamma)^2.
\end{aligned}
\]
The appendix shows that \(\delta_0^\gamma\to\infty\) as \(\gamma\to\infty\) and that \(D_I^\gamma\) grows at most linearly in \(\delta_0^\gamma\). Since total KL grows at least quadratically in the same displacement, the activated proportion tends to one. This holds even in dimension two, where the inherited bias and the rotational error act in the same plane. On $\mathbb R$, a square-integrable weighted-solenoidal field vanishes for every strictly positive probability density, so dimension two is the smallest in which activation is possible.
Section~\ref{sec:gmm_experiment} examines more relaxed  settings with component-specific covariances and more general errors.

\section{Numerical Experiments}
\label{sec:numerical-experiment}

We examine two implications of the theory. Synthetic Gaussian and Gamma mixtures show how errors that have zero marginal source under their reference densities can contribute to the guided density ratio. These experiments use prescribed gradient and weighted-solenoidal errors, without training a diffusion network. Pretrained image models then provide a qualitative evaluation of score filtering, which is motivated by the reference-density projections in Section~\ref{sec:cfg_hodge}.

\subsection{Synthetic distributions}\label{sec:simulations}

Following Theorem~\ref{thm:sec5_native_leakage_distribution}, we report the contributions of inherited and activated error to the log-density ratio $\log\big( \hat\rho_0^\gamma(x)
\big/ \rho_0^\gamma(x) \big)$. 
Whichever has a larger contribution is the main reason that $\hat \rho_0^\gamma$ deviates from $\rho_0^\gamma$. 
Such results give a general sense of how substantially the activated error impacts the quality of generated distribution. 
Recall that 
\begin{equation*}
    D_I^{\gamma,t} = \mathbb E_{\hat\rho_t^\gamma}[C_I^{\gamma,t}], \quad D_A^{\gamma,t} = \mathbb E_{\hat\rho_t^\gamma}[C_A^{\gamma,t}], \quad D_I^{\gamma,t} + D_A^{\gamma,t} = \mathrm{KL}(\hat\rho_t^\gamma\|\rho_t^\gamma).
\end{equation*}
We will compare the magnitudes of $D_I^{\gamma,t}$ and $D_A^{\gamma,t}$ across different $\gamma$ in synthetic distributions with controlled setting.

\subsubsection{Gaussian mixture} \label{sec:gmm_experiment}
We consider five equally weighted Gaussian components in dimension $d=20$. Class $0$ is the target distribution, and the full mixture is the unconditional distribution. For each $b\in\{c,u\}$, we manufacture artificial errors via
\begin{equation}\label{eq:gmm_error}
e_t^b(x)=\frac{1}{2} M_{b,t}x + \frac{1}{2} w_{b,t}(x), \quad
M_{b,t} \text{ is a symmetric real matrix,} \quad
\nabla\cdot\!\left(p_t^b w_{b,t}\right)=0, 
\end{equation}
where $c_g M_{b,t}x$ serves as the gradient component $v_t^b$, and $w_{b,t}$ is the weighted-solenoidal component constructed using skew-symmetric matrices. 
For each $b\in\{c,u\}$, we balance the magnitudes of the visible component $v_t^b$ and the invisible component $w_t^b$ through controlling $R_{\mathrm{inv}}^b(t)$, where 
\begin{equation*}
 R_{\mathrm{inv}}^b(t) = E_{\mathrm{inv}}^b(t)/E_{\mathrm{tot}}^b(t), \quad 
 E_{\mathrm{vis}}^b(t) =\mathbb E_{p_t^b}\left[\|v_t^b\|^2\right], \quad E_{\mathrm{inv}}^b(t) =\mathbb E_{p_t^b}\left[\|w_t^b\|^2\right]. 
\end{equation*}
The values of $R_{\mathrm{inv}}^b(t)$ across independent replicas range from 32\% to 47\%, details reported in Table~\ref{tab:gmm_native_energy} in Appendix~\ref{sec:gmm decomposition}.

We randomly generate five different types of error as in \eqref{eq:gmm_error} and sample $512$ reverse ODE chains for each type of error.  Every reverse chain has $100$ steps, starting from $0$ (initialization) and ending at $100$.  The noise schedule, relationship between step and $t$, and numerical evaluation are described in Appendix~\ref{sec:simu_settings}.

Table~\ref{tab:gmm_cfg_kl} reports the inherited contribution $D_I^{\gamma,t}$ and activated contribution $D_A^{\gamma,t}$ along the CFG trajectories across different $\gamma$.
The activated error has no effect at the baseline $\gamma=1$, as expected. But its effect increases substantially at large guidance strengths. When $\gamma \geq 10$, the effect from activated error dominates, accounting for nearly 80\% or 90\%. These results illustrate that the originally invisible error can substantially impact the  CFG probability flow.

\subsubsection{Gamma mixture}
\label{sec:gamma_experiment}

We consider five equally weighted Gamma components $Gamma(4)$ with different centers in dimension $d=20$, with shape parameter $4$. For each $b\in\{c,u\}$, we add artificial errors via
\begin{equation}
e_t^b(x)=\frac{1}{2} M_{b,t}x + \frac{1}{2} w_{b,t}(x),
\qquad
\nabla\cdot\!\left(p_t^b w_{b,t}\right)=0,
\label{eq:gamma_error}
\end{equation}
where $v_t^b(x)=(1/2) M_{b,t}x$ is the gradient component and $w_t^b(x)=(1/2) w_{b,t}(x)$ is the weighted-solenoidal component under $p_t^b$. 
The magnitude ratio $R_{\mathrm{inv}}^b(t)$ is approximately 47--48\% for the conditional errors and 32--38\% for the unconditional errors, comparable to the Gaussian mixture. Details are reported in Table~\ref{tab:gamma_native_energy} in Appendix~\ref{sec:gamma decomposition}. Construction and evaluation details are given in Appendix~\ref{sec:simu_settings}.
Table~\ref{tab:gamma_cfg_kl} reports consistent results as in the Gaussian mixture case. 

\begin{figure}[H]
    \centering
    \includegraphics[width=0.8\linewidth]{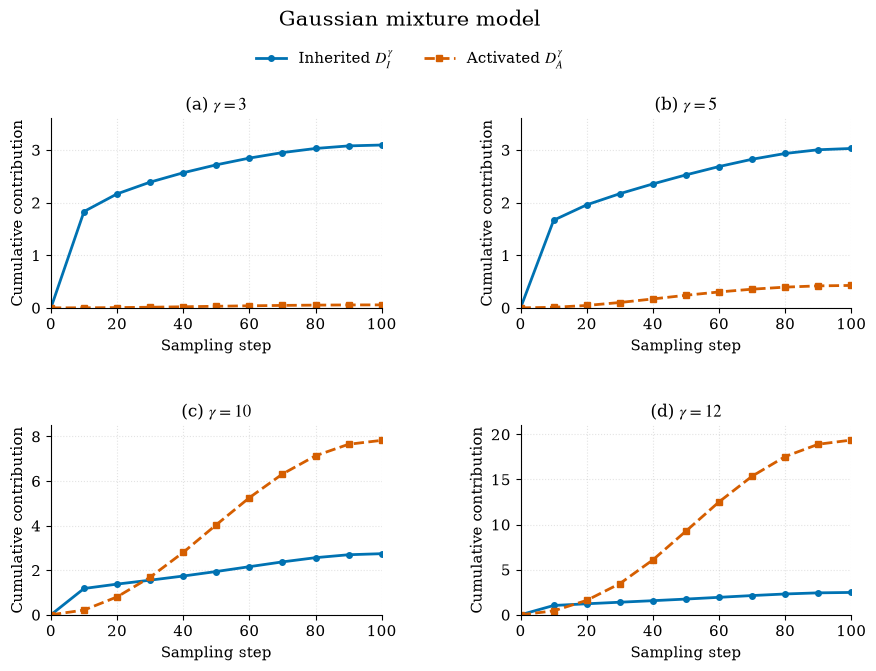}
    \caption{The effects that pulls $\hat\rho_0^\gamma$ away from $\rho_0^\gamma$ come from two sources: the inherited error and activated error. At each $\gamma$, the plot compares the cumulative effects from inherited error $D_I^\gamma$ (blue) and activated error $D_A^\gamma$ (orange) along the CFG trajectories. }
    \label{fig:GMM_DI_DA}
    \vspace{2mm}
    \begin{minipage}{\linewidth}
\begin{table}[H]
\centering
\caption{The size of effect that pulls $\hat\rho_0^\gamma$ away from $\rho_0^\gamma$. 
Five types of error are generated and $512$ reverse chains are sampled for each type. Thus, each entry is the mean over $5\times 512$ evaluations. 
When $\gamma=1$, CFG reduces to the conditional DDIM, serving as the baseline.
}
\vspace{3mm}
\label{tab:gmm_cfg_kl}
\setlength{\tabcolsep}{4pt}
\resizebox{0.9\linewidth}{!}{%
\begin{tabular}{r||*{5}{rr}}
\toprule
\multicolumn{1}{c||}{\multirow{3}{*}[-1ex]{Step}}
& \multicolumn{2}{c}{Baseline}
& \multicolumn{4}{c}{Moderate guidance}
& \multicolumn{4}{c}{Strong guidance} \\
\cmidrule(lr){2-3}
\cmidrule(lr){4-7}
\cmidrule(lr){8-11}
& \multicolumn{2}{c}{$\gamma=1$}
& \multicolumn{2}{c}{$\gamma=3$}
& \multicolumn{2}{c}{$\gamma=5$}
& \multicolumn{2}{c}{$\gamma=10$}
& \multicolumn{2}{c}{$\gamma=12$} \\
\cmidrule(lr){2-3}
\cmidrule(lr){4-5}
\cmidrule(lr){6-7}
\cmidrule(lr){8-9}
\cmidrule(lr){10-11}
& $D_I^\gamma$ & $D_A^\gamma$
& $D_I^\gamma$ & $D_A^\gamma$
& $D_I^\gamma$ & $D_A^\gamma$
& $D_I^\gamma$ & $D_A^\gamma$
& $D_I^\gamma$ & $D_A^\gamma$ \\
\midrule
0
& 0.0000 & 0.0000
& 0.0000 & 0.0000
& 0.0000 & 0.0000
& 0.0000 & 0.0000
& 0.0000 & 0.0000 \\
10
& \textbf{1.9224} & 0.0000
& \textbf{1.8304} & 0.0009
& \textbf{1.6653} & 0.0102
& \textbf{1.1790} & 0.2111
& \textbf{1.0298} & 0.4309 \\
20
& \textbf{2.2756} & 0.0000
& \textbf{2.1655} & 0.0049
& \textbf{1.9575} & 0.0452
& \textbf{1.3761} & 0.8025
& 1.2191 & \textbf{1.6001} \\
30
& \textbf{2.4907} & 0.0000
& \textbf{2.3888} & 0.0116
& \textbf{2.1692} & 0.1012
& 1.5529 & \textbf{1.6860}
& 1.3918 & \textbf{3.4524} \\
40
& \textbf{2.6418} & 0.0000
& \textbf{2.5668} & 0.0204
& \textbf{2.3546} & 0.1692
& 1.7381 & \textbf{2.7998}
& 1.5619 & \textbf{6.0845} \\
50
& \textbf{2.7534} & 0.0000
& \textbf{2.7175} & 0.0299
& \textbf{2.5263} & 0.2387
& 1.9409 & \textbf{4.0394}
& 1.7407 & \textbf{9.2733} \\
60
& \textbf{2.8367} & 0.0000
& \textbf{2.8454} & 0.0388
& \textbf{2.6846} & 0.3019
& 2.1575 & \textbf{5.2536}
& 1.9347 & \textbf{12.5133} \\
70
& \textbf{2.8976} & 0.0000
& \textbf{2.9499} & 0.0462
& \textbf{2.8233} & 0.3544
& 2.3730 & \textbf{6.3147}
& 2.1302 & \textbf{15.3594} \\
80
& \textbf{2.9393} & 0.0000
& \textbf{3.0283} & 0.0518
& \textbf{2.9330} & 0.3935
& 2.5618 & \textbf{7.1342}
& 2.3053 & \textbf{17.5399} \\
90
& \textbf{2.9638} & 0.0000
& \textbf{3.0771} & 0.0552
& \textbf{3.0038} & 0.4175
& 2.6934 & \textbf{7.6497}
& 2.4299 & \textbf{18.9017} \\
100
& \textbf{2.9718} & 0.0000
& \textbf{3.0937} & 0.0563
& \textbf{3.0283} & 0.4256
& 2.7408 & \textbf{7.8253}
& 2.4754 & \textbf{19.3642} \\
\bottomrule
\end{tabular}%
}
\end{table}
    \end{minipage}
\end{figure}

\newpage
\begin{figure}[H]
    \centering
    \includegraphics[width=0.8\linewidth]{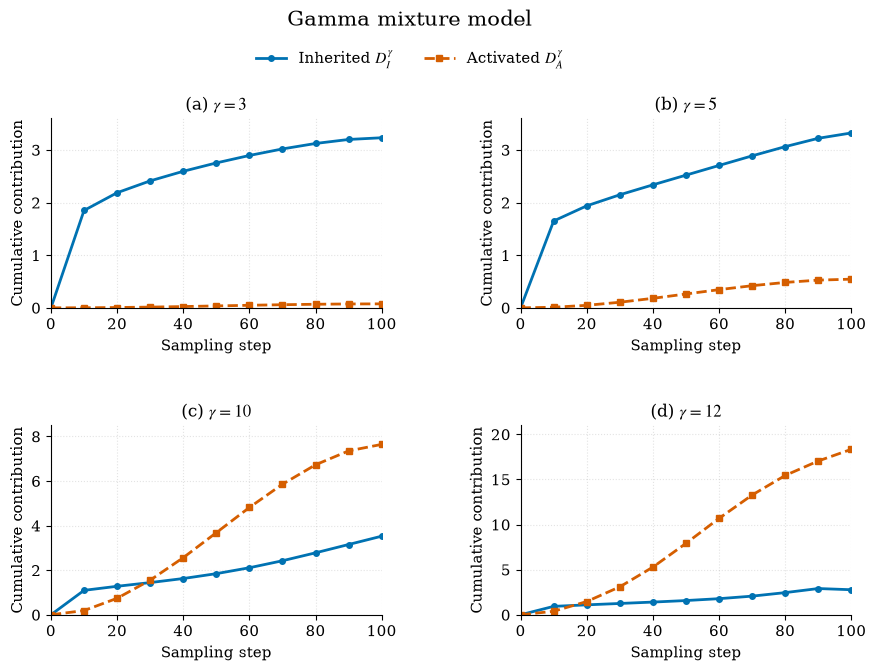}
    \caption{The effects that pulls $\hat\rho_0^\gamma$ away from $\rho_0^\gamma$ come from two sources: the inherited error and activated error. At each $\gamma$, the plot compares the cumulative effects from inherited error $D_I^\gamma$ (blue) and activated error $D_A^\gamma$ (orange) along the CFG trajectories. }
    \label{fig:Gamma_DI_DA}
    \vspace{2mm}
    \begin{minipage}{\linewidth}

\begin{table}[H]
\centering
\caption{The size of effect that pulls $\hat\rho_0^\gamma$ away from $\rho_0^\gamma$. 
Five types of error are generated and $512$ reverse chains are sampled for each type. Thus, each entry is the mean over $5\times 512$ evaluations. 
When $\gamma=1$, CFG reduces to the conditional DDIM, serving as the baseline.
}
\vspace{3mm}
\label{tab:gamma_cfg_kl}
\setlength{\tabcolsep}{4pt}
\resizebox{0.9\linewidth}{!}{%
\begin{tabular}{r||*{5}{rr}}
\toprule
\multicolumn{1}{c||}{\multirow{3}{*}[-1ex]{Step}}
& \multicolumn{2}{c}{Baseline}
& \multicolumn{4}{c}{Moderate guidance}
& \multicolumn{4}{c}{Strong guidance} \\
\cmidrule(lr){2-3}
\cmidrule(lr){4-7}
\cmidrule(lr){8-11}
& \multicolumn{2}{c}{$\gamma=1$}
& \multicolumn{2}{c}{$\gamma=3$}
& \multicolumn{2}{c}{$\gamma=5$}
& \multicolumn{2}{c}{$\gamma=10$}
& \multicolumn{2}{c}{$\gamma=12$} \\
\cmidrule(lr){2-3}
\cmidrule(lr){4-5}
\cmidrule(lr){6-7}
\cmidrule(lr){8-9}
\cmidrule(lr){10-11}
& $D_I^\gamma$ & $D_A^\gamma$
& $D_I^\gamma$ & $D_A^\gamma$
& $D_I^\gamma$ & $D_A^\gamma$
& $D_I^\gamma$ & $D_A^\gamma$
& $D_I^\gamma$ & $D_A^\gamma$ \\
\midrule
0
& 0.0000 & 0.0000
& 0.0000 & 0.0000
& 0.0000 & 0.0000
& 0.0000 & 0.0000
& 0.0000 & 0.0000 \\
10
& \textbf{1.9594} & 0.0000
& \textbf{1.8502} & 0.0015
& \textbf{1.6505} & 0.0113
& \textbf{1.0957} & 0.1888
& \textbf{0.9322} & 0.3887 \\
20
& \textbf{2.3181} & 0.0000
& \textbf{2.1871} & 0.0063
& \textbf{1.9383} & 0.0477
& \textbf{1.2780} & 0.7393
& 1.0999 & \textbf{1.4681} \\
30
& \textbf{2.5363} & 0.0000
& \textbf{2.4124} & 0.0140
& \textbf{2.1482} & 0.1063
& 1.4440 & \textbf{1.5525}
& 1.2532 & \textbf{3.0989} \\
40
& \textbf{2.6893} & 0.0000
& \textbf{2.5946} & 0.0244
& \textbf{2.3371} & 0.1808
& 1.6263 & \textbf{2.5612}
& 1.4073 & \textbf{5.2846} \\
50
& \textbf{2.8030} & 0.0000
& \textbf{2.7536} & 0.0365
& \textbf{2.5214} & 0.2631
& 1.8446 & \textbf{3.6837}
& 1.5772 & \textbf{7.9167} \\
60
& \textbf{2.8890} & 0.0000
& \textbf{2.8952} & 0.0486
& \textbf{2.7053} & 0.3455
& 2.1096 & \textbf{4.8140}
& 1.7844 & \textbf{10.6845} \\
70
& \textbf{2.9536} & 0.0000
& \textbf{3.0195} & 0.0594
& \textbf{2.8878} & 0.4208
& 2.4238 & \textbf{5.8575}
& 2.0660 & \textbf{13.2720} \\
80
& \textbf{3.0000} & 0.0000
& \textbf{3.1231} & 0.0681
& \textbf{3.0629} & 0.4829
& 2.7792 & \textbf{6.7289}
& 2.4558 & \textbf{15.4418} \\
90
& \textbf{3.0291} & 0.0000
& \textbf{3.1992} & 0.0737
& \textbf{3.2216} & 0.5265
& 3.1525 & \textbf{7.3522}
& 2.9080 & \textbf{17.0473} \\
100
& \textbf{3.0396} & 0.0000
& \textbf{3.2317} & 0.0756
& \textbf{3.3230} & 0.5449
& 3.5261 & \textbf{7.6441}
& 2.7753 & \textbf{18.3386} \\
\bottomrule
\end{tabular}%
}
\end{table}
    \end{minipage}
\end{figure}

\subsubsection{Score filtering} \label{sec:synthetic_score_filtering}
The preceding theory shows that retaining only the gradient component of each score error prevents guidance-induced activation. 
We adapt the score filtering procedure used in \citep{pmlr-v258-thornton25a}: for each conditional or unconditional score estimator, fit a scalar potential $\Phi_\theta^b$ by
\begin{equation}\label{eq:score_filtering}
 \min_\theta\; \mathbb E_t\, \lambda_t \mathbb E_{x\sim p_t^b} \bigl\|\nabla_x\Phi_\theta^b(x,t) -\hat s_t^b(x)\bigr\|^2.
\end{equation}
The minimizer $\nabla_x \hat \Phi$ is the projection of $\hat s_t^b$ onto the gradient space $\mathcal G_{p_t^b}$. 
Therefore, the filtering procedure gives
\begin{equation}
s_t^{b,\mathrm{filter}} :=P_{p_t^b}\hat s_t^b =s_t^b+v_t^b.
\label{eq:ideal_score_filtering}
\end{equation}
The ideal filtering removes the weighted-solenoidal errors and therefore eliminates guidance-induced error activation.

\begin{figure}[!htbp]
    \centering
    \includegraphics[width=1\linewidth]{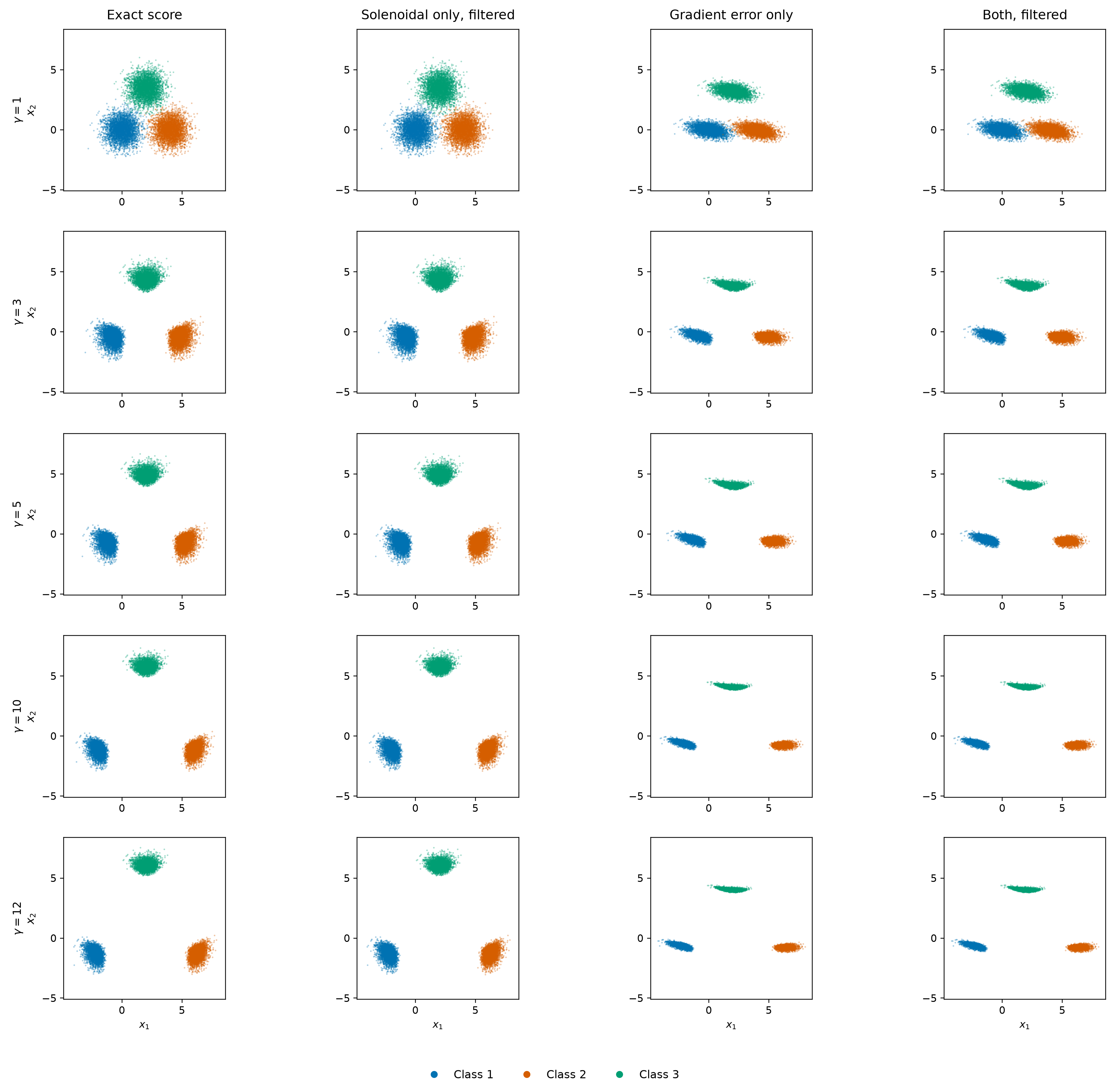}
    \caption{
    Ideal score filtering in the Gaussian-mixture example.
    Columns from left to right show exact-score CFG, filtering of the weighted-solenoidal-only case, gradient-only sampling, and filtering of the combined-error case.
    Rows correspond to $\gamma=1,3,5,10,12$, from top
    to bottom. After filtering, the generated samples in second column closely match exact-score CFG in first column; the generated samples in fourth column closely match those in third column. 
    }
    \label{fig:gmm_score_filtering}
\end{figure}

We then implement CFG with the filtered conditional and unconditional scores in the Gaussian-mixture example of Figure~\ref{fig:gmm_error_components}. 
Figure~\ref{fig:gmm_score_filtering} illustrates the
result across guidance strengths. When the original
error is weighted-solenoidal only, filtering recovers
the exact scores, and the generated samples closely
match exact-score CFG. When both error components are
present, filtering recovers the gradient-only scores,
and the samples closely match the gradient-only CFG. The remaining distortions reflect the effect by gradient error.

\subsection{Pretrained image models}
For pretrained image models, where the exact scores are unknown, we ask whether filtering helps preserve object structure as guidance strength increases. We use publicly available class-conditional and unconditional Elucidated Diffusion Model (EDM) checkpoints, trained independently on CIFAR-10, which contains \(32\times32\) images from ten classes.

For each checkpoint, we fit a scalar potential using the score-filtering objective in \eqref{eq:score_filtering} and use its spatial gradient as the filtered score. The expectation under \(p_t^b\) can be estimated by adding noise at level \(t\) to images from the specified class for \(b=c\) or from the full dataset for \(b=u\). With finite data, limited model capacity, and optimization error, the fitted gradient may differ from the ideal projection and thus change the gradient component of the score error.

We compare the original and filtered scores at \(\gamma\in\{1,3,5,7.5,10,12\}\) using matched network architectures. Within each image pair, we use the same initial noise, class label, and postprocessing. All runs use a 40-step deterministic Heun sampler (Algorithm 1 in \citep{karras2022edm}) with \(\sigma_{\max}=80\), \(\sigma_{\min}=0.002\), and an EDM noise-schedule exponent of \(7\). The appendix gives further experimental details.

\begin{figure*}[!htbp]
  \centering
  \includegraphics[width=\textwidth]{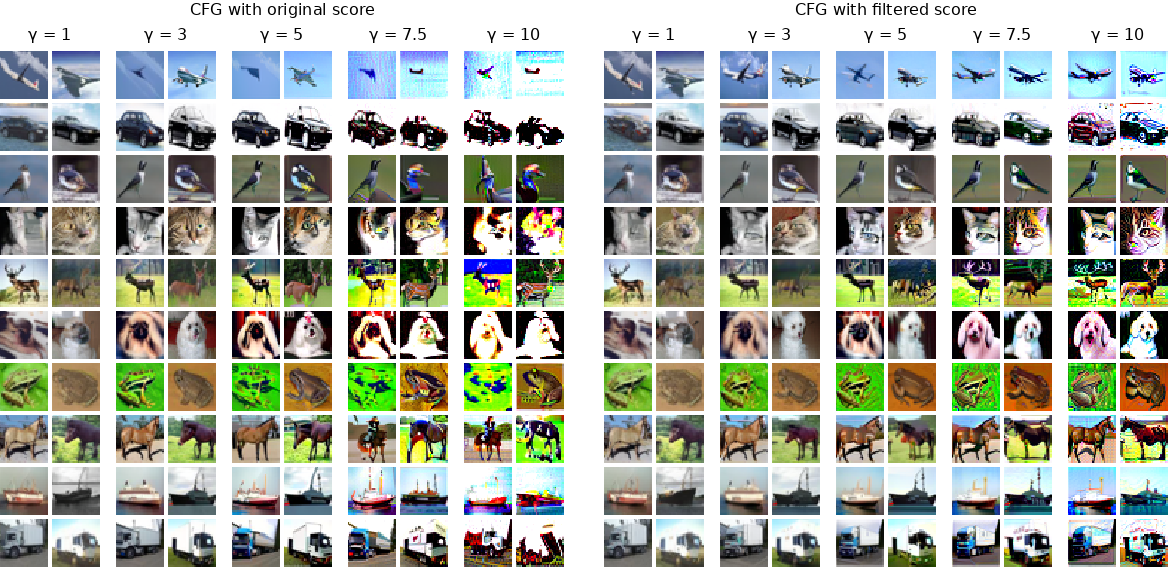}
  \caption{CIFAR-10 samples under classifier-free guidance from the
  original EDM (left) and its filtered counterpart (right),
  generated from the same initial noise and class labels (Heun, 40 steps).
  Each row is one class; two seeds per class.}
  \label{fig:edm}
\end{figure*}

Figure~\ref{fig:edm} shows that the original and filtered models produce nearly identical samples at $\gamma=1$, while their differences become more apparent as guidance increases. At $\gamma=3$ and $5$, both models generate recognizable objects, with several filtered samples retaining more natural colors and clearer details. At $\gamma=7.5$ and $10$, the original samples exhibit severe distortion, saturation, washed-out regions, and loss of object structure. Filtering mitigates these distortions in most examples, preserving clearer airplane silhouettes, car bodies, and animal contours, although they still exhibit saturation and background noise at the largest guidance strength.

As negative controls, we compare the original and filtered scores in unconditional-only and conditional-only ODE sampling.
Figure~\ref{fig:edm_distilled_branches} in Appendix~\ref{sec:edm_distilled_branches} shows almost no visible change in either setting. The greater benefit observed under CFG is consistent with our reasoning: reducing guidance-induced activation of errors that are weighted-solenoidal under their respective reference densities. However, while ideal filtering preserves the gradient error, the fitted filter may also modify it. These visual comparisons therefore support the proposed mechanism without isolating its contribution.

At high guidance strengths, samples from the original EDM show pronounced color saturation and loss of fine detail (Appendix Figures~\ref{fig:edm_strong_guidance} and \ref{fig:edm_ddim_all_classes}), consistent with the degradation reported in prior work \citep{saharia2022photorealistic, sadat2025eliminating}.

\section{Discussion}\label{sec:discussion}
We identify a mechanism that the score estimation error could have amplified impact on the evolving marginals under CFG. 
A score error that is weighted-solenoidal under its conditional or unconditional reference density at every noise time preserves the corresponding marginal path when it acts alone. Under CFG, the combination of two such errors can have nonzero weighted divergence under the guided density and therefore alter the guided marginals.
We call this mechanism \emph{guidance-induced error activation}. The activated error field is the projection of the combined solenoidal errors onto gradient fields under the guided density.

Controlled experiments with Gaussian and Gamma mixtures show that activation can contribute substantially to the discrepancy between learned-score and exact-score CFG at large guidance strengths. On CIFAR-10 images, score filtering helped preserve object contours. Saturation and background noise persisted at the largest guidance strength, suggesting that other mechanisms may also contribute to sampling degradation under strong guidance.

\bibliography{bib}
\bibliographystyle{iclr2027_conference}

\newpage
\appendix

\section{Proofs for Section 3}
\label{app:fixed_guidance}
This section provides the functional-analytic foundations of
Theorem~\ref{thm:fixed_guidance} and
Corollary~\ref{cor:one_way_transfer}.
Section~\ref{app:main_theorem_one} proves the activation
identities at a fixed guidance strength and establishes the
associated variational and dynamical characterizations.
Section~\ref{app:one_way_transfer} shows that the reference
gradient errors retain their gradient structure under the
guided density, subject to the stated smoothness and
integrability assumptions.

\subsection{Proof of Theorem~\ref{thm:fixed_guidance} and related characterizations}
\label{app:main_theorem_one}

\paragraph{Weighted Sobolev and Hodge spaces.}
Fix a strictly positive probability density $\rho$ on
$\mathbb R^d$. As in the main text, write
$L^2(\rho)=L^2(\rho;\mathbb R^d)$ for the space of
square-integrable vector fields. Let $\dot H^1(\rho)$ be the
completion of the smooth, compactly supported scalar-valued
functions $C_c^\infty(\mathbb R^d)$ under the gradient norm
\begin{equation}
\|\psi\|_{\dot H^1(\rho)}
:=\|\nabla\psi\|_{L^2(\rho)}.
\label{eq:app_sobolev_norm}
\end{equation}
The gradient map extends by completion to an isometric
isomorphism from $\dot H^1(\rho)$ onto
\begin{equation}
\mathcal G_\rho
:=\overline{\{\nabla\psi:\psi\in C_c^\infty(\mathbb R^d)\}}
^{\,L^2(\rho;\mathbb R^d)}.
\label{eq:app_gradient_space}
\end{equation}
Consequently, each element of this closed gradient subspace
has a unique potential as an element of $\dot H^1(\rho)$.
Its orthogonal complement $\mathcal S_\rho:=\mathcal G_\rho^\perp$
is the space of weighted-solenoidal fields:
\begin{equation}
\mathcal S_\rho
=\{z\in L^2(\rho;\mathbb R^d):
\nabla\cdot(\rho z)=0
\text{ in }\mathcal D'(\mathbb R^d)\}.
\label{eq:app_solenoidal_space}
\end{equation}
Here $\mathcal D'(\mathbb R^d)$ denotes the space of
distributions: continuous linear functionals on the
test-function space $C_c^\infty(\mathbb R^d)$. In particular,
$\nabla\cdot(\rho z)=0$ in this space means that
\[
\int_{\mathbb R^d}z\cdot\nabla\psi\,\rho\,dx=0
\qquad\text{for every }\psi\in C_c^\infty(\mathbb R^d).
\]
The characterization in Eq.~\eqref{eq:app_solenoidal_space}
therefore follows directly from orthogonality to gradients
and the definition of $\mathcal G_\rho$ as their closure.

Denote by $P_\rho$ the orthogonal projection onto
$\mathcal G_\rho$ and by $\dot H^{-1}(\rho)$ the continuous
dual of $\dot H^1(\rho)$. Because this construction uses the
gradient norm, it requires neither a global Poincar\'e
inequality nor a separate square-integrability bound on the
potential itself.

Define the operators
$Q_\rho:L^2(\rho;\mathbb R^d)\to\dot H^{-1}(\rho)$ and
$L_\rho:\dot H^1(\rho)\to\dot H^{-1}(\rho)$ by
\begin{equation}
\begin{aligned}
\langle Q_\rho z,\psi\rangle
&=\int_{\mathbb R^d}z\cdot\nabla\psi\,\rho\,dx,\\
\langle L_\rho\phi,\psi\rangle
&=\int_{\mathbb R^d}\nabla\phi\cdot\nabla\psi\,\rho\,dx,
\qquad \psi\in\dot H^1(\rho).
\end{aligned}
\label{eq:app_weak_operators}
\end{equation}
The brackets denote dual pairings. Both operators are bounded,
and $L_\rho$ is the Riesz isomorphism associated with the
gradient inner product on $\dot H^1(\rho)$.

\begin{lemma}[Weighted Poisson representation]
\label{lem:app_weighted_poisson}
For every $z\in L^2(\rho;\mathbb R^d)$, there is a unique
$\phi_z\in\dot H^1(\rho)$ such that
\begin{equation}
\int\nabla\phi_z\cdot\nabla\psi\,\rho\,dx
=\int z\cdot\nabla\psi\,\rho\,dx
\qquad\text{for every }\psi\in\dot H^1(\rho).
\label{eq:app_poisson_weak}
\end{equation}
This potential represents the gradient projection of $z$:
\begin{equation}
P_\rho z=\nabla\phi_z,
\qquad L_\rho\phi_z=Q_\rho z.
\label{eq:app_poisson_projection}
\end{equation}
\end{lemma}

\begin{proof}
By construction, $\dot H^1(\rho)$ is a Hilbert space with
inner product
\[
\langle\phi,\psi\rangle_{\dot H^1(\rho)}
:=\int\nabla\phi\cdot\nabla\psi\,\rho\,dx.
\]
For a fixed $z\in L^2(\rho;\mathbb R^d)$, define
\[
\ell_z(\psi):=\int z\cdot\nabla\psi\,\rho\,dx,
\qquad \psi\in\dot H^1(\rho).
\]
This functional is linear, and the Cauchy--Schwarz inequality
gives
\[
|\ell_z(\psi)|
\leq\|z\|_{L^2(\rho)}\|\nabla\psi\|_{L^2(\rho)}
=\|z\|_{L^2(\rho)}\|\psi\|_{\dot H^1(\rho)}.
\]
Thus $\ell_z$ is continuous on $\dot H^1(\rho)$.
The Riesz representation theorem therefore provides a unique
$\phi_z\in\dot H^1(\rho)$ such that
\[
\langle\phi_z,\psi\rangle_{\dot H^1(\rho)}
=\ell_z(\psi)
\qquad\text{for every }\psi\in\dot H^1(\rho).
\]
By the definitions of the inner product and $\ell_z$, this
is precisely Eq.~\eqref{eq:app_poisson_weak}.

We next identify $\nabla\phi_z$ as the gradient projection
of $z$. Rearranging the weak equation gives
\[
\int(z-\nabla\phi_z)\cdot\nabla\psi\,\rho\,dx=0
\qquad\text{for every }\psi\in\dot H^1(\rho).
\]
The gradient map sends $\dot H^1(\rho)$ onto
$\mathcal G_\rho$. Hence this identity says that
$z-\nabla\phi_z$ is orthogonal to every field in
$\mathcal G_\rho$, so it belongs to $\mathcal S_\rho$.
Since $\nabla\phi_z\in\mathcal G_\rho$, the decomposition
\[
z=\nabla\phi_z+(z-\nabla\phi_z)
\]
is the orthogonal decomposition of $z$ into its gradient
and weighted-solenoidal components. Therefore,
$P_\rho z=\nabla\phi_z$.

Finally, for every $\psi\in\dot H^1(\rho)$,
\[
\begin{aligned}
\langle L_\rho\phi_z,\psi\rangle
&=\int\nabla\phi_z\cdot\nabla\psi\,\rho\,dx\\
&=\int z\cdot\nabla\psi\,\rho\,dx\\
&=\langle Q_\rho z,\psi\rangle.
\end{aligned}
\]
The two continuous functionals agree on all of
$\dot H^1(\rho)$, proving $L_\rho\phi_z=Q_\rho z$.
\end{proof}

\paragraph{Change of reference density.}
We next evaluate the same vector field under two different
densities. The following lemma shows how a field with zero
weighted divergence under its reference density can acquire
nonzero weighted divergence under a new density.

\begin{lemma}[Weighted divergence under a change of density]
\label{lem:app_change_measure}
Let $p$ and $\rho$ be strictly positive probability densities,
and let $r:=\rho/p\in C^1_{\mathrm{loc}}(\mathbb R^d)$.
Suppose that $w\in L^1_{\mathrm{loc}}(\rho;\mathbb R^d)$,
$w\cdot\nabla\log r\in L^1_{\mathrm{loc}}(\rho)$, and
$\nabla\cdot(pw)=0$ in $\mathcal D'(\mathbb R^d)$.
Then
\begin{equation}
-\nabla\cdot(\rho w)
=-\rho w\cdot\nabla\log\frac{\rho}{p}
\quad\text{in }\mathcal D'(\mathbb R^d).
\label{eq:app_change_measure}
\end{equation}
If additionally $w\in L^2(\rho;\mathbb R^d)$, this gives
the representation
$Q_\rho w=-w\cdot\nabla\log(\rho/p)$ in the weighted
weak sense: for every $\psi\in C_c^\infty(\mathbb R^d)$,
\[
\langle Q_\rho w,\psi\rangle
=-\int_{\mathbb R^d}\psi\,w\cdot\nabla\log r\,\rho\,dx.
\]
\end{lemma}

\begin{proof}
Put $r=\rho/p$. We first check that the relevant vector
fields are locally integrable. For any compact set $K$,
strict positivity and continuity of $r$ give
$c_K:=\min_K r>0$. Hence
\[
\int_K |pw|\,dx
=\int_K\frac{\rho|w|}{r}\,dx
\leq c_K^{-1}\int_K\rho|w|\,dx<\infty.
\]
Thus both $pw$ and $\rho w$ define locally integrable
vector fields.

Fix $\psi\in C_c^\infty(\mathbb R^d)$. We would like to
test $\nabla\cdot(pw)=0$ against $r\psi$. This function is
compactly supported and $C^1$, but need not be smooth.
To justify its use, let $u=r\psi$ and choose smooth
approximations $u_n\in C_c^\infty(\mathbb R^d)$ by
mollification. Their supports lie in a fixed compact set
$K$, and
\[
\|u_n-u\|_\infty+\|\nabla u_n-\nabla u\|_\infty
\longrightarrow0.
\]
Since $\nabla\cdot(pw)=0$, we have
$\int pw\cdot\nabla u_n\,dx=0$ for every $n$. Moreover,
\[
\left|\int pw\cdot(\nabla u_n-\nabla u)\,dx\right|
\leq\|\nabla u_n-\nabla u\|_\infty
\int_K|pw|\,dx
\longrightarrow0.
\]
Passing to the limit therefore permits $r\psi$ as a test
function. Expanding its gradient yields
\begin{equation}
\begin{aligned}
0
&=\int pw\cdot\nabla(r\psi)\,dx\\
&=\int\rho w\cdot\nabla\psi\,dx
  +\int p\psi\,w\cdot\nabla r\,dx\\
&=\int\rho w\cdot\nabla\psi\,dx
  +\int\rho\psi\,w\cdot\nabla\log r\,dx,
\end{aligned}
\label{eq:app_change_measure_proof}
\end{equation}
where the last equality uses $p=\rho/r$.

By the definition of distributional divergence,
\[
\langle-\nabla\cdot(\rho w),\psi\rangle =\int\rho w\cdot\nabla\psi\,dx =-\int\rho\psi\,w\cdot\nabla\log r\,dx.
\]
Since this holds for every compactly supported smooth
$\psi$, it proves
\[
-\nabla\cdot(\rho w)
=-\rho w\cdot\nabla\log(\rho/p)
\quad\text{in }\mathcal D'(\mathbb R^d).
\]

Finally, if $w\in L^2(\rho;\mathbb R^d)$, then $Q_\rho w$
is well defined, and the same identity gives
\[
\langle Q_\rho w,\psi\rangle
=-\int\psi\,w\cdot\nabla\log(\rho/p)\,\rho\,dx.
\]
This is the claimed weighted weak representation.
\end{proof}

\paragraph{Error activation at a fixed guidance strength.}
\begin{proof}[Proof of Theorem~\ref{thm:fixed_guidance}]
Fix $t$ and $\gamma$, and abbreviate
$\rho:=\rho_t^\gamma$, $w:=w_t^\gamma$,
$g:=g_t^\gamma$, and $q:=q_t^\gamma$.

\textit{(i) Change-of-density identity.}
For each $b\in\{c,u\}$, the reference decomposition gives
$w_t^b\in\mathcal S_{p_t^b}$, hence
$\nabla\cdot(p_t^b w_t^b)=0$.
Applying Lemma~\ref{lem:app_change_measure} with
$p=p_t^b$ and $\rho=\rho_t^\gamma$ yields
\[
Q_\rho w_t^b
=-w_t^b\cdot\nabla\log\frac{\rho}{p_t^b},
\]
understood in the weighted weak sense.
Since $Q_\rho$ is linear and
$w=\gamma w_t^c+(1-\gamma)w_t^u$, we obtain
\[
\begin{aligned}
q=Q_\rho w
&=\gamma Q_\rho w_t^c+(1-\gamma)Q_\rho w_t^u\\
&=-\gamma w_t^c\cdot\nabla\log\frac{\rho}{p_t^c}
  -(1-\gamma)w_t^u\cdot\nabla\log\frac{\rho}{p_t^u}.
\end{aligned}
\]
This proves Eq.~\eqref{eq:exact_leakage_source}.

We next show that projecting $w$ onto $\mathcal G_\rho$
preserves its activation residual. By definition,
$g=P_\rho w$, so $w-g$ is orthogonal to
$\mathcal G_\rho$. Consequently, for every
$\psi\in\dot H^1(\rho)$,
\[
\langle Q_\rho(w-g),\psi\rangle
=\int(w-g)\cdot\nabla\psi\,\rho\,dx=0.
\]
Thus $Q_\rho(w-g)=0$, and therefore
\begin{equation}
Q_\rho w=Q_\rho g=q.
\label{eq:app_source_preservation}
\end{equation}

\textit{(ii) Activation energy.}
Applying Lemma~\ref{lem:app_weighted_poisson} to $w$
gives a unique $\phi_t^\gamma\in\dot H^1(\rho)$ such that
\begin{equation}
g_t^\gamma=\nabla\phi_t^\gamma,
\qquad
L_{\rho_t^\gamma}\phi_t^\gamma=q_t^\gamma.
\label{eq:app_leakage_poisson}
\end{equation}
In particular, for every $\psi\in\dot H^1(\rho)$,
\begin{equation}
\langle q,\psi\rangle
=\int g\cdot\nabla\psi\,\rho\,dx.
\label{eq:app_energy_pairing}
\end{equation}
The Cauchy--Schwarz inequality and the definition of the
dual norm give
\[
\begin{aligned}
\|q\|_{\dot H^{-1}(\rho)}
&=\sup_{\|\psi\|_{\dot H^1(\rho)}\leq1}
|\langle q,\psi\rangle|\\
&\leq\|g\|_{L^2(\rho)}.
\end{aligned}
\]

To prove the reverse inequality, suppose first that
$\|g\|_{L^2(\rho)}>0$ and choose
\[
\psi_*:=\frac{\phi_t^\gamma}{\|g\|_{L^2(\rho)}}.
\]
Because $\nabla\phi_t^\gamma=g$, this potential satisfies
$\|\psi_*\|_{\dot H^1(\rho)}=1$. Moreover,
Eq.~\eqref{eq:app_energy_pairing} gives
\[
\langle q,\psi_*\rangle
=\frac{\|g\|_{L^2(\rho)}^2}{\|g\|_{L^2(\rho)}}
=\|g\|_{L^2(\rho)}.
\]
Hence the upper bound is attained. If $g=0$, the same
pairing identity gives $q=0$, so equality also holds in
this case. We conclude that
\begin{equation}
\|q\|_{\dot H^{-1}(\rho)}^2
=\|g\|_{L^2(\rho)}^2
=\mathcal L_t(\gamma),
\label{eq:app_energy_identity}
\end{equation}
which proves Eq.~\eqref{eq:intrinsic_leakage_energy}.

\textit{(iii) Criterion for zero activation.}
The energy identity immediately gives $g=0$ if and only
if $q=0$. To connect this condition to weighted divergence,
observe that, for every $\psi\in C_c^\infty(\mathbb R^d)$,
\[
\langle q,\psi\rangle
=\int\rho w\cdot\nabla\psi\,dx
=\langle-\nabla\cdot(\rho w),\psi\rangle.
\]
Thus $q=0$ implies $\nabla\cdot(\rho w)=0$ in
$\mathcal D'(\mathbb R^d)$. Conversely, if this
distributional divergence vanishes, then $q$ vanishes
on $C_c^\infty(\mathbb R^d)$. Since these test functions
are dense in $\dot H^1(\rho)$ and $q$ is continuous on
that space, $q=0$ on all of $\dot H^1(\rho)$.
Therefore,
\[
g=0
\iff q=0
\iff\nabla\cdot(\rho w)=0,
\]
proving Eq.~\eqref{eq:zero_leakage_criteria}.
By Eq.~\eqref{eq:app_solenoidal_space}, the last condition
is also equivalent to $w\in\mathcal S_\rho$.

Finally, at $\gamma=1$, if $\rho_t^1=p_t^c$, then
$w_t^1=w_t^c\in\mathcal S_{p_t^c}$, and hence
\[
g_t^1=P_{p_t^c}w_t^c=0,
\qquad q_t^1=Q_{p_t^c}w_t^c=0.
\]
The same argument gives $g_t^0=q_t^0=0$ when
$\rho_t^0=p_t^u$.
At these guidance strengths, the guided velocity coincides
with the corresponding conditional or unconditional
reference velocity. Consequently, the matching identities
$\rho_t^1=p_t^c$ and $\rho_t^0=p_t^u$ hold along the
respective marginal paths whenever the corresponding
flows share the same entrance law at $t=T$ and their
continuity equations have unique solutions.
\end{proof}

\paragraph{Equivalent energy representations.}
The following proposition gives two complementary ways to
characterize the size of the activated error: through a
variational problem over potentials and through a minimum
over vector fields with the same activation residual.

\begin{proposition}[Equivalent representations of activation energy] \label{prop:app_energy_representations}

Under Assumption~\ref{ass:fixed_guidance},
\begin{equation}\label{eq:app_dual_energy}
\mathcal L_t(\gamma) =\langle q_t^\gamma,
L_{\rho_t^\gamma}^{-1}q_t^\gamma\rangle =\sup_{\psi\in\dot H^1(\rho_t^\gamma)}
\left\{2\langle q_t^\gamma,\psi\rangle
-\|\nabla\psi\|_{L^2(\rho_t^\gamma)}^2\right\},
\end{equation}
and
\begin{equation}
\mathcal L_t(\gamma)
=\inf_{\substack{z\in L^2(\rho_t^\gamma;\mathbb R^d)\\
Q_{\rho_t^\gamma}z=q_t^\gamma}}
\|z\|_{L^2(\rho_t^\gamma)}^2.
\label{eq:app_minimum_energy}
\end{equation}
The supremum is attained at
$\psi=L_{\rho_t^\gamma}^{-1}q_t^\gamma$, and the unique
minimizer is $z=g_t^\gamma$.
\end{proposition}

\begin{proof}
Abbreviate $\rho:=\rho_t^\gamma$, $q:=q_t^\gamma$, and
$g:=g_t^\gamma$. Let $\phi:=L_\rho^{-1}q$, so that
$g=\nabla\phi$. Testing the Poisson equation with $\phi$
gives $\langle q,\phi\rangle=\|g\|_{L^2(\rho)}^2$.
For any $\psi\in\dot H^1(\rho)$, completing the square yields
\begin{equation}
2\langle q,\psi\rangle-\|\nabla\psi\|_{L^2(\rho)}^2
=\|g\|_{L^2(\rho)}^2
-\|\nabla\psi-g\|_{L^2(\rho)}^2.
\label{eq:app_complete_square}
\end{equation}
The supremum is therefore $\|g\|_{L^2(\rho)}^2$, attained
at $\psi=\phi$. This proves
Eq.~\eqref{eq:app_dual_energy}.

For the minimum-energy characterization, any feasible field
$z$ satisfies $Q_\rho(z-g)=0$, hence
$z-g\in\mathcal S_\rho$. Orthogonality gives
\[
\|z\|_{L^2(\rho)}^2
=\|g\|_{L^2(\rho)}^2+\|z-g\|_{L^2(\rho)}^2.
\]
The minimum is attained uniquely at $z=g$, proving
Eq.~\eqref{eq:app_minimum_energy}.
\end{proof}

In particular, Eq.~\eqref{eq:app_dual_energy} establishes
the variational objective in Eq.~\eqref{eq:leakage_dual}.

\paragraph{Activation in the linearized marginal dynamics.}
We finally connect the activation residual to the leading
change in the sampling density caused by score error.
Throughout this argument, $t$ is the noise-time coordinate:
sampling starts at $T$ and proceeds toward $0$.

\begin{proposition}[Density changing rate under CFG]
\label{prop:app_linearized_response}
Let $\rho_t^{\gamma,\varepsilon}$ solve
\begin{equation}
\partial_t\rho_t^{\gamma,\varepsilon}
+\nabla\cdot\left[
\rho_t^{\gamma,\varepsilon}
(\zeta_t^\gamma-\varepsilon\kappa_t e_t^\gamma)
\right]=0,
\qquad \rho_T^{\gamma,\varepsilon}=\pi,
\label{eq:app_perturbed_continuity}
\end{equation}
where $\zeta_t^\gamma$ is the exact-score guided velocity.
Assume these evolutions are well posed for $\varepsilon$
near zero and differentiable in $\varepsilon$, with
differentiation valid in the weak continuity equation.
Then
$\eta_t^\gamma:=\left.\partial_\varepsilon
\rho_t^{\gamma,\varepsilon}\right|_{\varepsilon=0}$
satisfies
\begin{equation}
\partial_t\eta_t^\gamma
+\nabla\cdot(\zeta_t^\gamma\eta_t^\gamma)
=\kappa_t\nabla\cdot(\rho_t^\gamma e_t^\gamma),
\qquad \eta_T^\gamma=0.
\label{eq:app_linearized_continuity}
\end{equation}
At times when Assumption~\ref{ass:fixed_guidance} holds,
the contribution from $w_t^\gamma$ to this source satisfies
\begin{equation}
\kappa_t\nabla\cdot(\rho_t^\gamma w_t^\gamma)
=\kappa_t\nabla\cdot(\rho_t^\gamma g_t^\gamma).
\label{eq:app_linearized_leakage}
\end{equation}
Its action on a test function $\psi$ is
$-\kappa_t\langle q_t^\gamma,\psi\rangle$.
When $q_t^\gamma$ is represented by a function, this source
is $-\kappa_t\rho_t^\gamma q_t^\gamma$.
\end{proposition}

\begin{proof}
At $\varepsilon=0$, uniqueness gives
$\rho_t^{\gamma,0}=\rho_t^\gamma$.
For $\psi\in C_c^\infty(\mathbb R^d)$, the weak form of
Eq.~\eqref{eq:app_perturbed_continuity} is
\[
\frac{d}{dt}\int\psi\,\rho_t^{\gamma,\varepsilon}\,dx
=\int\nabla\psi\cdot
(\zeta_t^\gamma-\varepsilon\kappa_t e_t^\gamma)
\rho_t^{\gamma,\varepsilon}\,dx.
\]
Differentiating at $\varepsilon=0$ gives
\[
\frac{d}{dt}\langle\eta_t^\gamma,\psi\rangle
=\langle\eta_t^\gamma,\zeta_t^\gamma\cdot\nabla\psi\rangle
-\kappa_t\int\nabla\psi\cdot e_t^\gamma\rho_t^\gamma\,dx,
\]
where the pairings with $\eta_t^\gamma$ are defined by the
assumed weak derivative of the density. This is
Eq.~\eqref{eq:app_linearized_continuity} in the weak sense.
The entrance density is independent of
$\varepsilon$, so $\eta_T^\gamma=0$.

By Eq.~\eqref{eq:app_source_preservation},
$Q_{\rho_t^\gamma}w_t^\gamma
=Q_{\rho_t^\gamma}g_t^\gamma=q_t^\gamma$.
Consequently, for every test function $\psi$,
\[
-\kappa_t\int\rho_t^\gamma w_t^\gamma\cdot\nabla\psi\,dx
=-\kappa_t\int\rho_t^\gamma g_t^\gamma\cdot\nabla\psi\,dx
=-\kappa_t\langle q_t^\gamma,\psi\rangle.
\]
This proves Eq.~\eqref{eq:app_linearized_leakage} and the
stated expression for the source.
\end{proof}

These identities establish the dynamical interpretation in
Eqs.~\eqref{eq:linearized_density_response} and
\eqref{eq:leakage_density_source}.

\subsection{Proof of Corollary~\ref{cor:one_way_transfer}}
\label{app:one_way_transfer}

We establish that a gradient field retains its gradient
structure under a change of reference density, provided
its energy under the new density is finite. The proof
first identifies a weak scalar potential and then
approximates its gradient in the new weighted norm.

\begin{lemma}[Gradient fields under a change of reference density]
\label{lem:app_cross_law_gradient}
Let $p$ and $\rho$ be strictly positive $C^1$ probability
densities on $\mathbb R^d$. If
\begin{equation}
v\in\mathcal G_p,
\qquad
v\in L^2(\rho;\mathbb R^d),
\label{eq:app_cross_law_conditions}
\end{equation}
then $v\in\mathcal G_\rho$.
\end{lemma}

\begin{proof}
Since $v\in\mathcal G_p$, there exist
$\phi_n\in C_c^\infty(\mathbb R^d)$ such that
\begin{equation}
\nabla\phi_n\longrightarrow v
\quad\text{in }L^2(p;\mathbb R^d).
\label{eq:app_native_gradient_approx}
\end{equation}
We construct smooth, compactly supported potentials whose
gradients also converge to $v$ in $L^2(\rho)$.

\textit{Step 1: Identify a weak scalar potential.}
For any compact set $K\subset\mathbb R^d$, strict positivity
and continuity of $p$ give $c_K:=\min_K p>0$. Therefore,
\[
\int_K|\nabla\phi_n-v|^2\,dx
\leq c_K^{-1}\|\nabla\phi_n-v\|_{L^2(p)}^2
\longrightarrow0.
\]
Thus the convergence also holds in the unweighted space
$L^2_{\mathrm{loc}}(\mathbb R^d;\mathbb R^d)$.
Distributional derivatives pass to this limit. Since each
$\nabla\phi_n$ is curl-free, we obtain
\[
\partial_i v_j-\partial_j v_i=0
\quad\text{in }\mathcal D'(\mathbb R^d),
\qquad 1\leq i,j\leq d.
\]
The weak Poincar\'e lemma on the simply connected domain
$\mathbb R^d$ therefore gives a potential
$\Phi\in W^{1,2}_{\mathrm{loc}}(\mathbb R^d)$ such that
\begin{equation}
v=\nabla\Phi
\quad\text{almost everywhere on }\mathbb R^d.
\label{eq:app_global_potential}
\end{equation}
Here $W^{1,2}_{\mathrm{loc}}$ consists of functions whose
values and weak first derivatives are locally
square-integrable.

\textit{Step 2: Truncate the potential.}
We first bound the potential so that a subsequent spatial
cutoff produces a controllable error. For $M>0$, define
\[
T_M(a):=\max\{-M,\min\{a,M\}\},
\qquad
\Phi_M:=T_M(\Phi).
\]
Then $|\Phi_M|\leq M$, and the Sobolev chain rule gives
\[
\nabla\Phi_M
=\mathbf 1_{\{|\Phi|<M\}}v
\quad\text{almost everywhere}.
\]
Consequently,
\begin{equation}
\|\nabla\Phi_M-v\|_{L^2(\rho)}^2
=\int_{\{|\Phi|\geq M\}}|v|^2\rho\,dx
\longrightarrow0.
\label{eq:app_truncation}
\end{equation}
The convergence follows from dominated convergence and
the assumption $v\in L^2(\rho)$.

\textit{Step 3: Introduce a compact cutoff.}
Choose $\chi_R\in C_c^\infty(\mathbb R^d)$ such that
$0\leq\chi_R\leq1$, $\chi_R=1$ on $B_R$,
$\chi_R=0$ outside $B_{2R}$, and
$|\nabla\chi_R|\leq C/R$, where $C$ is independent of $R$.
Set $\Phi_{M,R}:=\chi_R\Phi_M$. Its weak gradient is
\[
\nabla\Phi_{M,R}
=\chi_R\nabla\Phi_M+\Phi_M\nabla\chi_R.
\]
For fixed $M$, dominated convergence gives
\[
\|(\chi_R-1)\nabla\Phi_M\|_{L^2(\rho)}
\longrightarrow0,
\]
because $|\nabla\Phi_M|\leq|v|$.
The additional term created by the cutoff satisfies
\begin{equation}
\begin{aligned}
\|\Phi_M\nabla\chi_R\|_{L^2(\rho)}^2
&\leq\frac{C^2M^2}{R^2}
\int_{B_{2R}\setminus B_R}\rho\,dx\\
&\leq\frac{C^2M^2}{R^2}
\longrightarrow0.
\end{aligned}
\label{eq:app_cutoff}
\end{equation}
The second inequality uses the fact that $\rho$ is a
probability density. Hence, for each fixed $M$,
\[
\nabla\Phi_{M,R}\longrightarrow\nabla\Phi_M
\quad\text{in }L^2(\rho)
\quad\text{as }R\to\infty.
\]

\textit{Step 4: Smooth the potentials and combine the approximations.}
For fixed $M$ and $R$, the function $\Phi_{M,R}$ is
compactly supported and belongs to $W^{1,2}(\mathbb R^d)$.
Mollification therefore produces
$\Phi_{M,R,\delta}\in C_c^\infty(\mathbb R^d)$ such that
\[
\nabla\Phi_{M,R,\delta}
\longrightarrow\nabla\Phi_{M,R}
\quad\text{in }L^2(dx)
\quad\text{as }\delta\to0.
\]
For sufficiently small $\delta$, all these functions have
support in a fixed compact set $K$. Since $\rho$ is
continuous, it is bounded above on $K$, and
\[
\begin{aligned}
&\|\nabla\Phi_{M,R,\delta}
       -\nabla\Phi_{M,R}\|_{L^2(\rho)}^2\\
&\qquad\leq
\left(\sup_K\rho\right)
\|\nabla\Phi_{M,R,\delta}
       -\nabla\Phi_{M,R}\|_{L^2(dx)}^2
\longrightarrow0.
\end{aligned}
\]
Thus smoothing also preserves convergence in the
$\rho$-weighted gradient norm.

Finally, choose $M_n$, then $R_n$, and then $\delta_n$
so that the truncation, cutoff, and smoothing errors are
each less than $1/(3n)$. Setting
$\psi_n:=\Phi_{M_n,R_n,\delta_n}$, the triangle inequality
gives
\[
\|\nabla\psi_n-v\|_{L^2(\rho)}<\frac1n.
\]
Since $\psi_n\in C_c^\infty(\mathbb R^d)$, this proves
$v\in\mathcal G_\rho$ by definition.
\end{proof}

\begin{proof}[Proof of Corollary~\ref{cor:one_way_transfer}]
For each $b\in\{c,u\}$, the reference decomposition gives
\[
v_t^b=P_{p_t^b}e_t^b\in\mathcal G_{p_t^b}.
\]
Assumption~\ref{ass:fixed_guidance} also gives
$v_t^b\in L^2(\rho_t^\gamma;\mathbb R^d)$.
The positivity and $C^1$ regularity of the densities,
together with Lemma~\ref{lem:app_cross_law_gradient},
therefore imply
\[
v_t^c,v_t^u\in\mathcal G_{\rho_t^\gamma}.
\]
Because $\mathcal G_{\rho_t^\gamma}$ is a linear subspace,
their CFG combination satisfies
\[
v_t^\gamma
=\gamma v_t^c+(1-\gamma)v_t^u
\in\mathcal G_{\rho_t^\gamma}.
\]
Orthogonal projection acts as the identity on this
subspace, so
\begin{equation}
P_{\rho_t^\gamma}v_t^\gamma=v_t^\gamma,
\qquad
(I-P_{\rho_t^\gamma})v_t^\gamma=0.
\label{eq:app_no_reverse_transfer}
\end{equation}
In particular, $h_t^\gamma=v_t^\gamma$.
Finally, linearity of the projection gives
\begin{equation*}
    o_t^\gamma =P_{\rho_t^\gamma}(v_t^\gamma+w_t^\gamma) = v_t^\gamma+g_t^\gamma,
\end{equation*}
which proves the corollary.
\end{proof}

The result establishes preservation of gradient structure.
The inherited and activated contributions need not be
orthogonal under the guided density and may reinforce or
cancel one another. Consequently, the result does not
imply monotonicity of the total observable error norm
with respect to the guidance strength.

\subsection{Proof of Proposition \ref{prop:reverse_sde_activation}}
We use similar techniques as in ODE-based sampler to extend the conclusion to SDE-based sampler. 
\begin{proof}
Fix $t,\gamma$ and abbreviate $\rho:=\rho_{t,\mathrm S}^\gamma$.
We first check the local conditions needed for the
change-of-density identity. Positivity and $C^1$ regularity give $\log(\rho/p_t^b)\in C^1_{\mathrm{loc}}$.
For every compact $K\subset\mathbb R^d$,
\begin{equation*}
\begin{aligned}
\int_K\left|w_t^b\cdot\nabla\log\frac\rho{p_t^b}\right|
\rho\,dx
&\leq\sup_K\left|\nabla\log\frac\rho{p_t^b}\right|
\|w_t^b\|_{L^2(\rho)}
\left(\int_K\rho\,dx\right)^{1/2}<\infty.
\end{aligned}
\end{equation*}
Thus the integrability and local regularity requirements in Assumption~\ref{ass:fixed_guidance} hold with $\rho$ in place of $\rho_t^\gamma$.

Since $\nabla\cdot(p_t^b w_t^b)=0$,
Lemma~\ref{lem:app_change_measure} gives
$Q_\rho w_t^b=-w_t^b\cdot\nabla\log(\rho/p_t^b)$.
Combining the two identities with the CFG weights proves Eq.~\eqref{eq:sde_change_density_identity}.
Also, $w_t^\gamma-g_{t,\mathrm S}^\gamma\in\mathcal G_\rho^\perp$, so $Q_\rho g_{t,\mathrm S}^\gamma=Q_\rho w_t^\gamma$.

By Lemma~\ref{lem:app_weighted_poisson}, there is
$\phi\in\dot H^1(\rho)$ such that
$g_{t,\mathrm S}^\gamma=\nabla\phi$ and
$\langle q_{t,\mathrm S}^\gamma,\psi\rangle
=\langle\nabla\phi,\nabla\psi\rangle_\rho$
for every $\psi\in\dot H^1(\rho)$.
Taking the dual norm gives
\begin{equation*}
\|q_{t,\mathrm S}^\gamma\|_{\dot H^{-1}(\rho)}
=\|\phi\|_{\dot H^1(\rho)}
=\|g_{t,\mathrm S}^\gamma\|_{L^2(\rho)}.
\end{equation*}
This proves the energy identity. Its zero case, together with the weak definition of $Q_\rho$, gives
Eq.~\eqref{eq:sde_zero_activation}.

Finally, $v_t^b\in\mathcal G_{p_t^b}\cap L^2(\rho)$.
Lemma~\ref{lem:app_cross_law_gradient} implies
$v_t^b\in\mathcal G_\rho$. Hence their linear combination $v_t^\gamma$ also belongs to $\mathcal G_\rho$, proving Eq.~\eqref{eq:sde_gradient_preservation}.
Linearity of $P_\rho$ then gives Eq.~\eqref{eq:sde_observable_error}.
\end{proof}

\section{Proofs for Section~\ref{sec:leakage_evolution}}
\label{app:variation_with_strength}
This section provides proofs for Lemma~\ref{lem:guidance_projection_continuity}, Theorem~\ref{thm:two_channel}, and Corollary~\ref{cor:sde_guidance_variation}.

\subsection{Proof of Lemma~\ref{lem:guidance_projection_continuity}}
\begin{proof}
Fix $\gamma\in I$ and any sequence $h_n\to0$ with
$\gamma+h_n\in I$. Write
$\rho_n:=\rho_t^{\gamma+h_n}$,
$w_n:=w_t^{\gamma+h_n}$, and
$g_n:=g_t^{\gamma+h_n}$, and let
$\rho:=\rho_t^\gamma$, $w:=w_t^\gamma$, and
$g:=g_t^\gamma$. Define
\[
a_n:=\sqrt{\rho_n}\,g_n,
\qquad b_n:=\sqrt{\rho_n}\,w_n,
\qquad b:=\sqrt\rho\,w.
\]
Choose a compact interval $J\subset I$ containing $\gamma$ in its
interior. For all sufficiently large $n$, the coefficients in $w_n$
are bounded by a constant depending only on $J$. Thus
$|b_n|^2+|b|^2\leq C_J H_{t,J}$ by condition~(iii).
Since $b_n\to b$ pointwise almost everywhere, dominated convergence
gives $b_n\to b$ in $L^2(dx)$.
Projection contractivity implies
$\|a_n\|_{L^2(dx)}\leq\|b_n\|_{L^2(dx)}$, so
$\{a_n\}$ is bounded.

Consider a subsequence along which $a_n$ converges weakly
to a field $a$ in $L^2(dx)$, and set $z:=a/\sqrt\rho$.
On any compact set $K$, condition~(i) and $\min_K\rho>0$ give
$\rho_n^{-1/2}\to\rho^{-1/2}$ uniformly. For any $F\in
L^2(K;\mathbb R^d)$, the fields $F/\sqrt{\rho_n}$ therefore
converge strongly in $L^2(K)$ to $F/\sqrt\rho$. Testing the weak
convergence of $a_n$ against these fields, extended by zero outside
$K$, gives
\[
\int_K g_n\cdot F\,dx
=\int_K a_n\cdot\frac{F}{\sqrt{\rho_n}}\,dx
\longrightarrow\int_K z\cdot F\,dx .
\]
Each $g_n\in\mathcal G_{\rho_n}$ is locally an $L^2$ limit of
smooth gradients because $\rho_n$ is positive and continuous.
Its distributional curl is therefore zero. Testing this identity
against compactly supported smooth functions and passing to the
local weak limit shows that $z$ is also curl-free.
The weak Poincar\'e lemma on $\mathbb R^d$ gives
$z=\nabla\Phi$ with $\Phi\in W^{1,2}_{\mathrm{loc}}(\mathbb R^d)$.
Since $\|z\|_{L^2(\rho)}=\|a\|_{L^2(dx)}<\infty$, the truncation,
cutoff, and mollification argument in the proof of
Lemma~\ref{lem:app_cross_law_gradient} applies to $\Phi$ and yields
$z\in\mathcal G_\rho$.

For every $\psi\in C_c^\infty(\mathbb R^d)$, projection
orthogonality gives
\[
\int(a_n-b_n)\cdot\sqrt{\rho_n}\,\nabla\psi\,dx=0.
\]
The test fields $\sqrt{\rho_n}\,\nabla\psi$ converge
strongly in $L^2(dx)$. Passing to the limit yields
$\langle z-w,\nabla\psi\rangle_\rho=0$.
Since $z\in\mathcal G_\rho$, this identifies
$z=P_\rho w=g$ and $a=\sqrt\rho\,g$.
Every weak subsequential limit has this value, so the
entire sequence $a_n$ converges weakly to $a$.

To recover strong convergence, use the fact that the residual
$w_n-g_n$ is orthogonal to $g_n$ itself in $L^2(\rho_n)$.
Together with $a_n\rightharpoonup a$ and $b_n\to b$ in $L^2(dx)$,
this gives
\[
\|a_n\|_{L^2(dx)}^2
=\langle a_n,b_n\rangle_{L^2(dx)}
\longrightarrow
\langle a,b\rangle_{L^2(dx)}
=\|a\|_{L^2(dx)}^2.
\]
Weak convergence and convergence of the norms imply
strong convergence of $a_n$, proving
Eq.~\eqref{eq:weighted_activation_continuity}.
On each compact set, $\rho_n^{-1/2}$ converges uniformly
to $\rho^{-1/2}$, which gives
Eq.~\eqref{eq:local_activation_continuity}.
Finally,
$\mathcal L_t(\gamma+h_n)=\|a_n\|_{L^2(dx)}^2$
converges to $\mathcal L_t(\gamma)$.
\end{proof}

\subsection{Proof of Theorem~\ref{thm:two_channel}}
\begin{proof}
Fix $t$ and $\gamma$. Write
$\rho_h:=\rho_t^{\gamma+h}$, $w_h:=w_t^{\gamma+h}$,
and $g_h:=g_t^{\gamma+h}$, and omit the subscript $h$
for their values at $h=0$. The projection identity holds
at both guidance strengths. For $\psi\in C_c^\infty(\mathbb R^d)$,
write $\rho_h=\rho+(\rho_h-\rho)$ in
$\int(g_h-w_h)\cdot\nabla\psi\,\rho_h\,dx=0$. Subtracting the
identity at $h=0$ and dividing by $h$ gives
\begin{equation}
\begin{aligned}
\int\frac{g_h-g}{h}\cdot\nabla\psi\,\rho\,dx
&=\int\frac{w_h-w}{h}\cdot\nabla\psi\,\rho\,dx\\
&\quad+\int(w_h-g_h)\cdot\nabla\psi\,
\frac{\rho_h-\rho}{h}\,dx.
\end{aligned}
\label{eq:activation_response_difference_quotient}
\end{equation}
The first term equals
$\langle w_t^c-w_t^u,\nabla\psi\rangle_\rho$ because
$w_t^\gamma$ is affine in $\gamma$.
Let $K$ be a compact set containing the support of $\nabla\psi$.
Since $\min_K\rho>0$ and $w_t^c,w_t^u\in L^2(\rho)$, both fields
belong to $L^2(K)$. Hence $w_h-w=h(w_t^c-w_t^u)\to0$ in
$L^2(K)$. Lemma~\ref{lem:guidance_projection_continuity} gives
$g_h\to g$ in the same space, so $w_h-g_h\to w-g$ in $L^2(K)$.
Condition~(ii) gives $(\rho_h-\rho)/h\to\partial_\gamma\rho_t^\gamma$
in $L^2(K)$. By Cauchy--Schwarz, the product of the two factors
converges in $L^1(K)$. Since $\nabla\psi$ is bounded, the second
term tends to $\mathcal D_{t,\gamma}(\psi)$.
The weak limit therefore exists and equals the right-hand
side of Eq.~\eqref{eq:two_channel_weak} on test functions.

The limiting functional is bounded in the gradient norm:
\[
\left|\langle\partial_\gamma w_t^\gamma,\nabla\psi
\rangle_\rho+\mathcal D_{t,\gamma}(\psi)\right|
\leq\left(
\|\partial_\gamma w_t^\gamma\|_{L^2(\rho)}
+\|\mathcal D_{t,\gamma}\|_{\dot H^{-1}(\rho)}
\right)\|\nabla\psi\|_{L^2(\rho)}.
\]
It therefore extends uniquely to $\dot H^1(\rho)$.
The Riesz representation theorem and the gradient
isomorphism from $\dot H^1(\rho)$ onto $\mathcal G_\rho$
give the unique field $G_t^\gamma$.
Applying the same representation to
$\mathcal D_{t,\gamma}$ gives $D_t^\gamma$, with
$\|D_t^\gamma\|_{L^2(\rho)}
=\|\mathcal D_{t,\gamma}\|_{\dot H^{-1}(\rho)}$.
Projection preserves the pairing of $\partial_\gamma w_t^\gamma$
with every gradient test field. Thus $M_t^\gamma+D_t^\gamma$
represents the same functional as $G_t^\gamma$, and uniqueness
gives $G_t^\gamma=M_t^\gamma+D_t^\gamma$.
Finally, the triangle inequality and contractivity of
$P_\rho$ yield Eq.~\eqref{eq:two_channel_bound}.
\end{proof}

\subsection{Proof of Corollary~\ref{cor:sde_guidance_variation}}
\begin{proof}
The proof of Lemma~\ref{lem:guidance_projection_continuity}
uses only positivity and local continuity of the density,
the error envelope, and weighted projection orthogonality.
It applies with $\rho_t^\eta$ replaced by
$\rho_{t,\mathrm S}^\eta$, proving
Eq.~\eqref{eq:sde_weighted_activation_continuity} and its
consequences for local continuity and activation energy.

For the rate of change, fix $\gamma\in I$ and abbreviate
$\rho_h:=\rho_{t,\mathrm S}^{\gamma+h}$,
$g_h:=g_{t,\mathrm S}^{\gamma+h}$, and
$w_h:=w_t^{\gamma+h}$. Omit $h$ at $h=0$.
Subtracting the projection identities at $\gamma+h$ and
$\gamma$ gives, for $\psi\in C_c^\infty(\mathbb R^d)$,
\begin{equation}
\begin{aligned}
\int\frac{g_h-g}{h}\cdot\nabla\psi\,\rho\,dx
&=\int(w_t^c-w_t^u)\cdot\nabla\psi\,\rho\,dx\\
&\quad+\int(w_h-g_h)\cdot\nabla\psi\,
\frac{\rho_h-\rho}{h}\,dx.
\end{aligned}
\label{eq:app_sde_guidance_difference_quotient}
\end{equation}
On the support of $\nabla\psi$, local continuity gives
$w_h-g_h\to w-g$ in $L^2$, while condition~(ii) gives
$(\rho_h-\rho)/h\to\partial_\gamma\rho_{t,\mathrm S}^\gamma$
in $L^2$. The product therefore converges in $L^1$, and
the last term tends to $\mathcal D_{t,\gamma,\mathrm S}(\psi)$.
This proves Eq.~\eqref{eq:sde_guidance_rate_weak} on test
functions. Condition~(iv) bounds the limiting functional
in the gradient norm, so it extends continuously to
$\dot H^1(\rho_{t,\mathrm S}^\gamma)$. The Riesz
representation theorem gives the unique fields
$G_{t,\mathrm S}^\gamma$ and $D_{t,\mathrm S}^\gamma$,
with
\[
\|D_{t,\mathrm S}^\gamma\|_{L^2(\rho_{t,\mathrm S}^\gamma)}
=\|\mathcal D_{t,\gamma,\mathrm S}\|_{
\dot H^{-1}(\rho_{t,\mathrm S}^\gamma)}.
\]
The definition of $M_{t,\mathrm S}^\gamma$ yields the
decomposition, and the triangle inequality and projection
contractivity give Eq.~\eqref{eq:sde_guidance_rate_bound}.
\end{proof}

\section{Proofs for Section~\ref{sec:analytic_learned_cfg}}
\label{app:learned-CFG-generated-density}
This section provides proofs for the theoretical results in Section~\ref{sec:analytic_learned_cfg}. 

\subsection{Proof of Theorem~\ref{thm:analytic_learned_cfg}}
\begin{proof}
For $b\in\{c,u\}$, divide the reference continuity equation by
$p_t^b$ and insert $\zeta_t^b=f(t)x-\kappa_t s_t^b$. This gives
\[
\partial_t\log p_t^b+f(t)x\cdot s_t^b+d f(t)
=\kappa_t\bigl(\nabla\cdot s_t^b+|s_t^b|^2\bigr).
\]
Take the linear combination with coefficients $\gamma$ and
$1-\gamma$. Since $\nabla\log R_t^\gamma=s_t^\gamma$, the
divergence terms cancel upon forming
\[
\begin{aligned}
\partial_t\log R_t^\gamma
+\zeta_t^\gamma\cdot\nabla\log R_t^\gamma
+\nabla\cdot\zeta_t^\gamma
&=\kappa_t\bigl(\gamma|s_t^c|^2+(1-\gamma)|s_t^u|^2
-|s_t^\gamma|^2\bigr)\\
&=-\kappa_t\gamma(\gamma-1)|\nabla H_t|^2.
\end{aligned}
\]
The last equality follows by expanding
$|\gamma s_t^c+(1-\gamma)s_t^u|^2$.

Along the learned trajectory $x_t=\hat\Phi_{0,t}^\gamma(x)$,
the learned continuity equation gives
$\frac{d}{dt}\log\hat\rho_t^\gamma(x_t)=-(\nabla\cdot\hat\zeta_t^\gamma)(x_t)$.
For $\log R_t^\gamma$, changing the velocity in the preceding
display from $\zeta_t^\gamma$ to $\hat\zeta_t^\gamma$ adds
$-\kappa_t e_t^\gamma\cdot s_t^\gamma$ to the derivative.
Subtracting the two derivatives and using
$\nabla\cdot(\zeta_t^\gamma-\hat\zeta_t^\gamma)
=\kappa_t\nabla\cdot e_t^\gamma$ yields
\[
\frac{d}{dt}
\log\frac{\hat\rho_t^\gamma(x_t)}{R_t^\gamma(x_t)}
=\kappa_t
\left\{\gamma(\gamma-1)|\nabla H_t|^2
+\nabla\cdot e_t^\gamma
+e_t^\gamma\cdot s_t^\gamma\right\}(x_t).
\]
At $t=T$, the density ratio is $\pi(x_T)/R_T^\gamma(x_T)$.
Subtracting the integral of this derivative from the terminal
log-ratio and exponentiating gives
Eq.~\eqref{eq:sec5_analytic_learned_distribution}.
\end{proof}

\subsection{Proof of Theorem~\ref{thm:sec5_native_leakage_distribution}}
\begin{proof}
Along $x_t=\hat\Phi_{0,t}^\gamma(x)$, the exact-score
continuity equation gives
\[
\frac{d}{dt}\log\rho_t^\gamma(x_t)
=\bigl[-\nabla\cdot\zeta_t^\gamma
+(\hat\zeta_t^\gamma-\zeta_t^\gamma)\cdot\nabla\log\rho_t^\gamma\bigr](x_t).
\]
Subtract this from the learned log-density derivative
$-(\nabla\cdot\hat\zeta_t^\gamma)(x_t)$ and use
$\hat\zeta_t^\gamma-\zeta_t^\gamma=-\kappa_t e_t^\gamma$ to obtain
\[
\frac{d}{dt}
\log\frac{\hat\rho_t^\gamma(x_t)}{\rho_t^\gamma(x_t)}
=\kappa_t\left(\nabla\cdot e_t^\gamma
+e_t^\gamma\cdot\nabla\log\rho_t^\gamma\right)(x_t)
=-\kappa_t(Q_{\rho_t^\gamma}e_t^\gamma)(x_t).
\]
Since $w_t^\gamma=e_t^\gamma-v_t^\gamma$ is $C^1$ in space,
$q_t^\gamma=Q_{\rho_t^\gamma}w_t^\gamma$ has the scalar
representative $-(\rho_t^\gamma)^{-1}\nabla\cdot(\rho_t^\gamma w_t^\gamma)$.
Thus at each point of the trajectory,
\[
Q_{\rho_t^\gamma}e_t^\gamma
=Q_{\rho_t^\gamma}v_t^\gamma+q_t^\gamma
=-\nabla\cdot v_t^\gamma
-v_t^\gamma\cdot\nabla\log\rho_t^\gamma+q_t^\gamma.
\]
The log-ratio is zero at $T$. Its value at $0$ is therefore
$\int_0^T\kappa_t(Q_{\rho_t^\gamma}e_t^\gamma)(x_t)\,\mathrm dt$,
which gives Eq.~\eqref{eq:sec5_native_leakage_density}.
\end{proof}

\subsection{Proof of Proposition~\ref{prop:sde_density_representations}}
\label{app:sde_density_representations}

\paragraph{A general density-ratio equation.}
Let $h_t(x)>0$ be a $C^{1,2}$ function satisfying
\begin{equation}
\partial_t h_t+\nabla\cdot(\xi_t^\gamma h_t)
=-\kappa_t\Delta h_t+a_t h_t,
\label{eq:app_sde_reference_residual}
\end{equation}
where $a_t(x)$ is a scalar function. No normalization or
spatial integrability of $h_t$ is required. Define
\[
r_t^h:=\frac{\hat\rho_{t,\mathrm S}^\gamma}{h_t},
\qquad
V_t^h:=a_t-\frac{2\kappa_t}{h_t}
\nabla\cdot(h_t e_t^\gamma).
\]
Insert $h_t r_t^h$ into the learned Fokker--Planck equation,
move all terms to the left, and divide by $h_t$. The cross
term $2\nabla h_t\cdot\nabla r_t^h$ in $\Delta(h_t r_t^h)$
contributes $2\kappa_t\nabla\log h_t$ to the drift. The
coefficient of $r_t^h$ is
\[
\frac{\partial_t h_t+\nabla\cdot(\hat\xi_t^\gamma h_t)
+\kappa_t\Delta h_t}{h_t}
=a_t+\frac{\nabla\cdot[(\hat\xi_t^\gamma-\xi_t^\gamma)h_t]}{h_t}
=V_t^h,
\]
since $\hat\xi_t^\gamma-\xi_t^\gamma=-2\kappa_t e_t^\gamma$.
The ratio thus solves the terminal-value equation
\begin{equation}
\partial_t r_t^h
+\bigl(\hat\xi_t^\gamma+2\kappa_t\nabla\log h_t\bigr)
\cdot\nabla r_t^h
+\kappa_t\Delta r_t^h+V_t^h r_t^h=0,
\qquad r_T^h=\frac{\pi}{h_T}.
\label{eq:app_sde_ratio_backward_equation}
\end{equation}

\paragraph{Feynman--Kac representation and integrability.}
Let $Z^h$ start from $x$ at time zero with drift
$\hat\xi_t^\gamma+2\kappa_t\nabla\log h_t$ and diffusion
coefficient $\sigma(t)$. Since $\sigma^2(t)/2=\kappa_t$,
It\^o's formula applied to
\[
M_u^h:=r_u^h(Z_u^h)
\exp\!\left(\int_0^u V_t^h(Z_t^h)\,\mathrm dt\right)
\]
and Eq.~\eqref{eq:app_sde_ratio_backward_equation}
shows that $M^h$ is a local martingale. Stop on exiting
the ball of radius $n$, at
$\tau_n:=T\wedge\inf\{u\geq0:|Z_u^h|\geq n\}$,
with $n>|x|$. On $[0,\tau_n]$, the ratio and the potential
are bounded by continuity on $[0,T]\times\overline B(0,n)$.
The stopped process is therefore a bounded martingale and
$\mathbb E[M_{\tau_n}^h]=r_0^h(x)$.

Non-explosion gives $M_{\tau_n}^h\to M_T^h$ almost surely.
For $h=R^\gamma$ and $h=\rho_{\cdot,\mathrm S}^\gamma$,
condition~\ref{cond:sec5_sde_fk} gives uniform integrability
of $\{M_{\tau_n}^h:n>|x|\}$, hence convergence in $L^1$.
Consequently,
\begin{equation}
r_0^h(x)
=\mathbb E_x^h\!\left[
\frac{\pi(Z_T^h)}{h_T(Z_T^h)}
\exp\!\left(\int_0^T V_t^h(Z_t^h)\,\mathrm dt\right)
\right].
\label{eq:app_sde_general_feynman_kac}
\end{equation}
For example, boundedness of $r_t^h$ and a uniform upper bound
on $V_t^h$ imply this uniform integrability. A uniform
$L^{1+\delta}$ bound on the stopped variables for some
$\delta>0$ also suffices.

\begin{proof}[Proof of Proposition~\ref{prop:sde_density_representations}]
For part~(i), take $h_t=R_t^\gamma$. Since
$\xi_t^\gamma=\zeta_t^\gamma-\kappa_t s_t^\gamma$ and
$R_t^\gamma s_t^\gamma=\nabla R_t^\gamma$,
\[
\nabla\cdot(\xi_t^\gamma R_t^\gamma)+\kappa_t\Delta R_t^\gamma
=\nabla\cdot(\zeta_t^\gamma R_t^\gamma).
\]
The reference forward equations can be written as the
continuity equations used in the proof of
Theorem~\ref{thm:analytic_learned_cfg}, because
$p_t^b s_t^b=\nabla p_t^b$. Multiplying the identity for $\log R_t^\gamma$
from that proof by $R_t^\gamma$ gives
\begin{equation}
\partial_t R_t^\gamma
+\nabla\cdot(\xi_t^\gamma R_t^\gamma)
=-\kappa_t\Delta R_t^\gamma
-\kappa_t\gamma(\gamma-1)|\nabla H_t|^2R_t^\gamma.
\label{eq:app_sde_powered_product_residual}
\end{equation}
Thus $a_t=-\kappa_t\gamma(\gamma-1)|\nabla H_t|^2$.
Since $\nabla\log R_t^\gamma=s_t^\gamma$,
the auxiliary drift and potential become
\[
\begin{aligned}
\hat\xi_t^\gamma+2\kappa_t\nabla\log R_t^\gamma
&=f(t)x-2\kappa_t e_t^\gamma,\\
V_t^R
&=-\kappa_t\Bigl[
\gamma(\gamma-1)|\nabla H_t|^2
+2\nabla\cdot e_t^\gamma
+2e_t^\gamma\cdot s_t^\gamma\Bigr].
\end{aligned}
\]
Equation~\eqref{eq:app_sde_general_feynman_kac} therefore
gives Eq.~\eqref{eq:sec5_sde_powered_product_density}.

For part~(ii), take $h_t=\rho_{t,\mathrm S}^\gamma$.
The exact-score Fokker--Planck equation gives $a_t=0$, and
\[
V_t^\rho
=2\kappa_tQ_{\rho_{t,\mathrm S}^\gamma}e_t^\gamma
=2\kappa_t\bigl(
Q_{\rho_{t,\mathrm S}^\gamma}v_t^\gamma
+q_{t,\mathrm S}^\gamma\bigr).
\]
Here we used $e_t^\gamma=v_t^\gamma+w_t^\gamma$ and
the definition of $q_{t,\mathrm S}^\gamma$.
The terminal ratio is one because
$\hat\rho_{T,\mathrm S}^\gamma
=\rho_{T,\mathrm S}^\gamma=\pi$.
Equation~\eqref{eq:app_sde_general_feynman_kac} now yields
Eq.~\eqref{eq:sec5_sde_baseline_density}.

For part~(iii), write $p=\hat\rho_{t,\mathrm S}^\gamma$,
$q=\rho_{t,\mathrm S}^\gamma$, $r=p/q$, and
$K(t)=\int p\log r\,\mathrm dx$. In increasing noise time,
Eq.~\eqref{eq:sec5_sde_density_equations} reads
$\partial_t p=-\nabla\cdot J_p$ and
$\partial_t q=-\nabla\cdot J_q$, with fluxes
\[
J_p=p\hat\xi_t^\gamma+\kappa_t\nabla p,
\qquad J_q=q\xi_t^\gamma+\kappa_t\nabla q.
\]
Since $\int\partial_t p\,\mathrm dx=0$, differentiating
$K$ and integrating by parts as assumed in part~(iii) gives
\[
\begin{aligned}
K'(t)
&=\int\bigl[(\partial_t p)\log r-r\,\partial_t q\bigr]\,\mathrm dx
 =\int (J_p-r J_q)\cdot\nabla\log r\,\mathrm dx\\
&=-2\kappa_t\int p\,e_t^\gamma\cdot\nabla\log r\,\mathrm dx
  +\kappa_t\int p\,|\nabla\log r|^2\,\mathrm dx.
\end{aligned}
\]
In the second line we used $\nabla p-r\nabla q=p\nabla\log r$.
A further integration by parts expresses the drift term through $Q_q$, since
\[
\int p\,e_t^\gamma\cdot\nabla\log r\,\mathrm dx
=\int q\,e_t^\gamma\cdot\nabla r\,\mathrm dx
=-\int r\,\nabla\cdot(q e_t^\gamma)\,\mathrm dx
=\int p\,Q_q e_t^\gamma\,\mathrm dx.
\]
Finally, $K(T)=0$ by the common starting density. Substituting
$Q_q e_t^\gamma=Q_q v_t^\gamma+q_{t,\mathrm S}^\gamma$
and integrating $K(0)=-\int_0^T K'(t)\,\mathrm dt$ gives
Eq.~\eqref{eq:sec5_sde_kl_decomposition}, including the negative
sign of the squared-gradient term at the sampling endpoint.
\end{proof}

\subsection{A general sufficient condition for activation dominance}
\label{app:general_activation_dominance}

If $X_0\sim\hat\rho_0^\gamma$, then $\hat\Phi_{0,t}^\gamma(X_0)$
has density $\hat\rho_t^\gamma$. Under the stated space-time
integrability condition, Fubini's theorem thus converts the expectation
of each path integral in Eq.~\eqref{eq:reported_path_contributions}
at time zero into the corresponding integral in
Eq.~\eqref{eq:dominance_cumulative_contributions}.
To obtain Eq.~\eqref{eq:dominance_alignment_identities},
fix $t$ and $\gamma$, and abbreviate
$\rho=\rho_t^\gamma$ and $\hat\rho=\hat\rho_t^\gamma$.
For a vector field $F$ whose weighted divergence is
represented by a function, integration by parts gives
\[
\begin{aligned}
\int\hat\rho\,Q_\rho F\,dx =-\int\frac{\hat\rho}{\rho}
\nabla\cdot(\rho F)\,dx =\int\rho F\cdot\nabla\frac{\hat\rho}{\rho}\,dx
=\int\hat\rho F\cdot
\nabla\log\frac{\hat\rho}{\rho}\,dx.
\end{aligned}
\]
We assume that the displayed pairings are integrable and
that the integrations by parts introduce no boundary
terms. Applying this identity to $v_t^\gamma$ and
$g_t^\gamma$, using $Q_{\rho_t^\gamma}g_t^\gamma=q_t^\gamma$,
and integrating in time proves
Eq.~\eqref{eq:dominance_alignment_identities}.
For nonsmooth projected fields, the same calculation
is understood by approximation in the weak divergence
identity, with convergence of all displayed pairings.

For time-dependent vector fields, write
\begin{equation}
\langle F,G\rangle_\gamma
:=\int_0^T\kappa_t\,
\mathbb E_{\hat\rho_t^\gamma}[F_t\cdot G_t]\,\mathrm dt,
\qquad
\|F\|_\gamma^2:=\langle F,F\rangle_\gamma.
\label{eq:dominance_spacetime_norm}
\end{equation}
The projection $P_{\rho^\gamma}$ below acts at each
time $t$ as $P_{\rho_t^\gamma}$. Define
$\Delta w:=w^c-w^u$ and $\Delta v:=v^c-v^u$.

\begin{proposition}[A sufficient guidance threshold for dominance]
\label{prop:general_activation_dominance}
Assume that Eq.~\eqref{eq:dominance_alignment_identities}
holds and that all fields below have finite
$\|\cdot\|_\gamma$ norms. Suppose there are constants
$\alpha>0$, $\beta,\nu_0,\nu_1\geq0$, and
$\theta\in(0,1]$, independent of $\gamma$, such that
for every $\gamma\geq\gamma_0\geq1$,
\begin{equation}
\begin{aligned}
\|P_{\rho^\gamma}\Delta w\|_\gamma&\geq\alpha,&
\|P_{\rho^\gamma}w^u\|_\gamma&\leq\beta,\\
\|v^u\|_\gamma&\leq\nu_0,&
\|\Delta v\|_\gamma&\leq\nu_1,
\end{aligned}
\label{eq:dominance_general_bounds}
\end{equation}
and the activated field satisfies the cumulative
alignment condition
\begin{equation}
\langle g^\gamma,r^\gamma\rangle_\gamma
\geq\theta\|g^\gamma\|_\gamma\|r^\gamma\|_\gamma.
\label{eq:dominance_positive_alignment}
\end{equation}
Let $R\geq1$ and suppose $\theta\alpha>R\nu_1$.
Whenever $\|r^\gamma\|_\gamma>0$ and
\begin{equation}
\gamma\geq\gamma_0,
\qquad
\gamma>
\frac{R\nu_0+\theta\beta}{\theta\alpha-R\nu_1},
\label{eq:dominance_general_threshold}
\end{equation}
the actual cumulative contributions obey
\[
D_A^\gamma>R|D_I^\gamma|.
\]
\end{proposition}

\begin{proof}
Linearity of the projection gives
\[
g^\gamma
=\gamma P_{\rho^\gamma}\Delta w
+P_{\rho^\gamma}w^u,
\qquad
v^\gamma=\gamma\Delta v+v^u.
\]
The triangle inequality and its reverse therefore imply
\[
\|g^\gamma\|_\gamma\geq\gamma\alpha-\beta,
\qquad
\|v^\gamma\|_\gamma\leq\nu_0+\gamma\nu_1.
\]
By Eq.~\eqref{eq:dominance_alignment_identities},
the alignment assumption and Cauchy--Schwarz yield
\[
\begin{aligned}
D_A^\gamma
&\geq\theta(\gamma\alpha-\beta)
\|r^\gamma\|_\gamma,\\
|D_I^\gamma|
&\leq(\nu_0+\gamma\nu_1)
\|r^\gamma\|_\gamma.
\end{aligned}
\]
The threshold is exactly the strict inequality
$\theta(\gamma\alpha-\beta)
>R(\nu_0+\gamma\nu_1)$, proving the result.
\end{proof}

The assumptions require nondegenerate activation of the
conditional--unconditional solenoidal difference,
comparatively slower growth of the inherited field, and
positive alignment with the density discrepancy. They
are additional conditions, not consequences of
Assumption~\ref{ass:fixed_guidance}. In particular,
if $v^c=v^u$, then $\nu_1=0$ and the threshold reduces
to $(R\nu_0+\theta\beta)/(\theta\alpha)$.
If $D_I^\gamma\geq0$, the choices $R=4$ and $R=9$
give activated shares above $80\%$ and $90\%$ of
endpoint KL. For the mixture in
Proposition~\ref{prop:finite_mixture_activation_dominance},
the next proof derives a threshold directly from the model
parameters.

\subsection{Proof of Proposition~\ref{prop:finite_mixture_activation_dominance}}
\label{app:finite_mixture_activation_dominance}

\paragraph{The score of the mixture.}
For $0\leq k<K$, let
\[
\tau_{k,t}(x)=\frac{\pi_k\varphi_d(x-m_{k,t})}{p_t^u(x)},\qquad
b_t(x)=-\sum_{k=1}^{K-1}\tau_{k,t}(x)m_{k,t},\qquad
\beta_t(x)=u\cdot b_t(x).
\]
The weights $\tau_{k,t}$ are the posterior class probabilities.
Since $\nabla\varphi_d(x-m)=-(x-m)\varphi_d(x-m)$,
the mixture score is the posterior-weighted average of the component scores. Thus
\[
s_t^c(x)=-x,\qquad s_t^u(x)=-x-b_t(x),\qquad
s_t^\gamma(x)=-x+(\gamma-1)b_t(x).
\]
The field $b_t$ takes values in $E_0$ and depends only on the
coordinates in $E_0$. All its posterior weights are positive,
so $\beta_t(x)>0$. Moreover,
\begin{equation}
|b_t(x)|\leq C\beta_t(x),\qquad
C:=\max_{1\leq k<K}\frac{|\mu_k|}{-u\cdot\mu_k}.
\label{eq:mixture_direction_bound}
\end{equation}
This bound follows by summing the individual bounds
$|m_{k,t}|\leq C(-u\cdot m_{k,t})$ with weights $\tau_{k,t}$.
For finite $\gamma$, both velocities are sums of affine fields
and bounded smooth fields with bounded first and second spatial
derivatives. Indeed, the posterior weights have affine log odds.
The flows and their inverses therefore have at most linear growth
and bounded first and second spatial derivatives on $[0,T]$.
Their positive densities transported from $\varphi_d$ have scores
of at most linear growth, so the terms in
Eq.~\eqref{eq:dominance_cumulative_contributions} are integrable.

\paragraph{The coordinate shared by the classes.}
Use $r=T-t$ for elapsed sampling time, and let
$\widehat X(r)$ be the learned trajectory initialized at
$\widehat X(0)=x_T\sim\varphi_d$. In this time coordinate,
the exact-score velocity is $x+s_{T-r}^\gamma(x)=(\gamma-1)b_{T-r}(x)$.
Its component along $\vartheta$ vanishes, and its remaining components depend
only on the coordinates in $E_0\subset \vartheta^\perp$. Hence $\vartheta\cdot x_T$
is conserved and remains independent of the coordinates in $\vartheta^\perp$,
which evolve from their own initial values. The density $\rho_t^\gamma$
therefore factors with a standard normal density in the $\vartheta$ coordinate,
giving $\partial_{\vartheta}\log\rho_t^\gamma(x)=-\vartheta\cdot x$.
Since $Q_{\rho_t^\gamma}(-c^\prime\vartheta)=c^\prime\,\partial_{\vartheta}\log\rho_t^\gamma$,
setting $Z(r)=\vartheta\cdot\widehat X(r)$ gives
\begin{equation}
D_I^\gamma=-c^\prime\int_0^T\mathbb E[Z(r)]\,\mathrm dr.
\label{eq:mixture_inherited_mean}
\end{equation}
Also, for each $\lambda\in\mathbb R$, the density proportional
to $e^{\lambda\vartheta\cdot x}\rho_0^\gamma(x)$ has normalizing
constant $e^{\lambda^2/2}$. Nonnegativity of KL relative to
this tilted density gives
$\mathrm{KL}(\hat\rho_0^\gamma\|\rho_0^\gamma)
\geq\lambda\mathbb E[Z(T)]-\lambda^2/2$. Maximization over
$\lambda$ yields
\begin{equation}
D_I^\gamma+D_A^\gamma
\geq\tfrac12\bigl(\mathbb E[Z(T)]\bigr)^2.
\label{eq:mixture_endpoint_mean_bound}
\end{equation}

\paragraph{Accumulation of the transverse error.}
Put $k_r=\omega_{T-r}=\omega e^{-2r}$.
Since $w_t^\gamma=-\omega_t \boldsymbol{J} s_t^\gamma$, the learned
trajectory satisfies
\[
\frac{\mathrm d\widehat X(r)}{\mathrm dr}
=k_r\boldsymbol{J}\widehat X(r)-c^\prime\vartheta
+(\gamma-1)(\bI_d-k_r\boldsymbol{J})b_{T-r}(\widehat X(r)).
\]
Let $\theta_{s,r}=\int_s^r k_v\,\mathrm dv$ for $0\leq s\leq r$.
The fundamental matrix $e^{\theta_{s,r}\boldsymbol{J}}$ rotates the plane
spanned by $u,\vartheta$ and fixes its orthogonal complement. Write
$b=b_{T-s}(\widehat X(s))$ and $\beta=u\cdot b$. As $b\perp\vartheta$,
we have $\boldsymbol{J}b=\beta\vartheta$. The $\vartheta$ component of
$e^{\theta_{s,r}\boldsymbol{J}}(\bI_d-k_s\boldsymbol{J})b$ is therefore
$\beta(\sin\theta_{s,r}-k_s\cos\theta_{s,r})$.
In the variation-of-constants formula, the initial term
$e^{\theta_{0,r}\boldsymbol{J}}x_T$ has mean zero, so taking the $\vartheta$ component gives
\[
\begin{aligned}
-\mathbb E[Z(r)] =c^\prime\int_0^r\cos\theta_{s,r}\,\mathrm ds +(\gamma-1)\int_0^r (k_s\cos\theta_{s,r}-\sin\theta_{s,r})
\mathbb E[\beta_{T-s}(\widehat X(s))]\,\mathrm ds .
\end{aligned}
\]
Since $\theta_{s,r}=k_s(1-e^{-2(r-s)})/2\leq k_s/2\leq1/2$,
we have $\sin\theta_{s,r}\leq k_s/2$ and
$\cos\theta_{s,r}\geq\cos(1/2)$. Consequently,
\[
0<c_*:=\omega e^{-2T}\bigl(\cos(1/2)-1/2\bigr)
\leq k_s\cos\theta_{s,r}-\sin\theta_{s,r}\leq\omega.
\]
This positive lower bound explains the choice of $\omega_t$.
A constant amplitude $k_r=\omega$ also gives a positive lower
bound whenever $\omega T<\arctan\omega$.
Define the accumulated drift in direction $u$ by
\[
S_\gamma=(\gamma-1)\int_0^T
\beta_{T-s}(\widehat X(s))\,\mathrm ds,\qquad
L_\gamma=\mathbb E[S_\gamma].
\]
For $\gamma\geq1$, the preceding bounds show that
$-\mathbb E[Z(T)]\geq c_*L_\gamma$ and
$0<-\mathbb E[Z(r)]\leq c^\prime r+\omega L_\gamma$ for $0<r\leq T$.
Equations~\eqref{eq:mixture_inherited_mean}
and~\eqref{eq:mixture_endpoint_mean_bound} now imply
\begin{equation}
0<D_I^\gamma\leq (c^\prime T)^2/2+c^\prime\omega T L_\gamma,\qquad
D_I^\gamma+D_A^\gamma\geq c_*^2 L_\gamma^2/2.
\label{eq:mixture_linear_quadratic_bounds}
\end{equation}
At $\gamma=1$, the exact sampler has density $p_t^c$ and
$w_t^1=w_t^c$ is weighted-solenoidal under this density.
Hence $q_t^1=0$ and $D_A^1=0$.

\paragraph{A sufficient threshold.}
We show that $L_\gamma$ eventually exceeds any prescribed
$L>0$ and give a sufficient scale at which this occurs.
Let $E=E_0\oplus\operatorname{span}\{\vartheta\}$, let $P_E$
be Euclidean projection onto $E$, and put $k_*=\dim E$.
For fixed $x_T$, the same variation-of-constants formula,
orthogonality of $e^{\theta \boldsymbol{J}}$, and Eq.~\eqref{eq:mixture_direction_bound}
give
\[
\sup_{0\leq r\leq T}|P_E\widehat X(r)|
\leq |P_E x_T|+c^\prime T+(1+\omega)C S_\gamma .
\]
Set $M=\max_{k\geq1}|\mu_k|$ and
$\delta=\min_{k\geq1}(-u\cdot\mu_k)>0$, with the indices
restricted to $k<K$. For $|P_E x|\leq B$, put
$h=\exp(-MB-M^2/2)\leq1$. Each likelihood ratio
$\varphi_d(x-m_{k,t})/\varphi_d(x)=\exp(m_{k,t}\cdot x-|m_{k,t}|^2/2)$
is at least $h$. The posterior odds $(1-\tau_{0,t})/\tau_{0,t}$
are therefore at least $(1-\pi_0)h/\pi_0$. Converting odds to
probabilities gives
\[
1-\tau_{0,t}\geq
\frac{(1-\pi_0)h}{\pi_0+(1-\pi_0)h}\geq(1-\pi_0)h .
\]
Since $-u\cdot m_{k,t}\geq e^{-T}\delta$ for $k\geq1$, it follows that
\[
\beta_t(x)\geq b(B):=
e^{-T}\delta(1-\pi_0)\exp(-MB-M^2/2)>0.
\]
The event $|P_E x_T|\leq\sqrt{2k_*}$ has probability at least
$1/2$, since $\mathbb E|P_E x_T|^2=k_*$. On this event,
$S_\gamma\leq2L$ would give
$|P_E\widehat X(r)|\leq B_L:=\sqrt{2k_*}+c^\prime T+2(1+\omega)CL$
at every time. It would follow that
$S_\gamma\geq(\gamma-1)T b(B_L)$. Thus if
$\gamma>1+2L/[T b(B_L)]$, the assumption $S_\gamma\leq2L$
leads to a contradiction on this event. Since $S_\gamma\geq0$
everywhere and $S_\gamma>2L$ on an event of probability at least
$1/2$, we obtain $L_\gamma=\mathbb E[S_\gamma]>L$.
This proves $L_\gamma\to\infty$. For $\gamma>1$,
Eq.~\eqref{eq:mixture_linear_quadratic_bounds} gives
\[
0<\frac{D_I^\gamma}{D_I^\gamma+D_A^\gamma}
\leq\frac{(c^\prime T)^2+2c^\prime\omega T L_\gamma}{c_*^2 L_\gamma^2}
\longrightarrow0 .
\]
Subtracting this fraction from one gives the limit of the activated
share. Taking its reciprocal and subtracting one gives
$D_A^\gamma/D_I^\gamma\to\infty$.
For an explicit sufficient threshold for $D_A^\gamma>R D_I^\gamma$,
take
\[
L_R=\max\left\{1,\frac{2(R+1)((c^\prime T)^2/2+c^\prime\omega T)}{c_*^2}\right\},
\qquad
\Gamma_R^{\mathrm{mix}}=1+\frac{2L_R}{T b(B_{L_R})}.
\]
Indeed, $\gamma>\Gamma_R^{\mathrm{mix}}$ implies $L_\gamma>L_R\geq1$,
so $c_*^2L_\gamma^2/2>(R+1)((c^\prime T)^2/2+c^\prime\omega T)L_\gamma
\geq(R+1)D_I^\gamma$. The bound depends on the dimension of $E$ and
the mixture and error parameters. Adding independent standard
Gaussian coordinates leaves the bound unchanged.

Finally, on the line, $(pw)'=0$ implies $pw$ is constant.
Cauchy--Schwarz gives
$\int_{-b}^{b}1/p\,\mathrm dx\geq(2b)^2/\int_{-b}^{b}p\,\mathrm dx$.
Thus $\int_{\mathbb R}1/p\,\mathrm dx=\infty$, and
$\int |w|^2p\,\mathrm dx<\infty$ forces that constant to be zero.

\section{Experimental details}
\subsection{Simulation settings}
\label{sec:experiment_settings}
\label{sec:simu_settings}

\paragraph{Noise schedule and sampling steps.}
For both distribution families, we use the variance-preserving
Gaussian perturbation
\begin{equation*}
    X_t=\alpha_t X_{\mathrm{data}}+\bar\sigma_t\epsilon,
    \qquad \epsilon\sim\mathcal N(0,I_d),
    \qquad \alpha_t^2+\bar\sigma_t^2=1.
\end{equation*}
Here $\bar\sigma_t$ is the cumulative perturbation standard
deviation, as in Section~2; it is distinct from the instantaneous
diffusion coefficient $\sigma(t)$.
We use a cosine-inspired schedule~\citep{ho2022classifier},
truncated to the finite log-SNR range $[-20,20]$.
We normalize this truncated noise-time interval to $[0,T]$ with
$T=1$, and set
\begin{align*}
    &\theta_{\min} =\arctan(e^{-10}),\quad \theta_{\max} =\arctan(e^{10}),\quad
    \theta_t=(1-t/T)\theta_{\min}+(t/T)\theta_{\max},\\
    &\alpha_t=\cos\theta_t, \ \bar\sigma_t=\sin\theta_t,
    \ \lambda_t=\log\frac{\alpha_t^2}{\bar\sigma_t^2}.
\end{align*}
Thus $\lambda_t$ decreases from $20$ to $-20$ during forward noising and increases from $-20$ to $20$ during sampling.
The finite endpoints approximate clean data and complete noising. In particular, the numerical endpoint at $t=0$ has a small, nonzero perturbation variance.
Sampling uses the 101 evaluation nodes
\begin{equation*}
    t_n=T(1-n/100),\qquad n=0,\ldots,100.
\end{equation*}
The step index $n$ increases from initialization at $t_0=T$ to the low-noise endpoint at $t_{100}=0$. Results are reported every 10 steps.

\paragraph{Component distributions.}
Both experiments have dimension $d=20$ and five equally weighted components. With $p_{k,\mathrm{data}}$ denoting component $k$, define
\begin{equation*}
    p_{\mathrm{data}}^c=p_{0,\mathrm{data}}, \qquad p_{\mathrm{data}}^u=\frac15\sum_{k=0}^{4}p_{k,\mathrm{data}}.
\end{equation*}
The two distribution families share the same component means and covariances. We set $\mu_0=0$ and sample
\begin{equation*}
    \mu_k=4\frac{a_k}{\|a_k\|_2},
    \qquad a_k\sim\mathcal N(0,I_d),
    \qquad k=1,\ldots,4.
\end{equation*}
For each component, let
\begin{equation*}
    \Sigma_k=Q_k\operatorname{diag}
    (\nu_{k,1},\ldots,\nu_{k,d})Q_k^\top,
    \qquad \nu_{k,j}\sim\mathcal U(0.5,1.5),
\end{equation*}
where $Q_k$ is sampled from the Haar distribution on the orthogonal group $O(d)$. The realized parameters $(\mu_k,Q_k,\nu_k)$ are shared between the two experiments and fixed throughout the
noise schedule.

\textbf{Gaussian mixture.}
The component densities are
\begin{equation*}
    p_{k,\mathrm{data}}(x)=\mathcal N(x;\mu_k,\Sigma_k).
\end{equation*}

\textbf{Gamma mixture.}
Write the Gamma density with shape $a$ and scale $\vartheta$ as
\begin{equation*}
    g(z;a,\vartheta) =\frac{z^{a-1}e^{-z/\vartheta}}
    {\Gamma(a)\vartheta^a}\mathbf1_{\{z>0\}}.
\end{equation*}
We sample independent coordinates
$Z_{k,j}\sim\mathrm{Gamma}(4,\vartheta_{k,j})$,
where $\vartheta_{k,j}=\sqrt{\nu_{k,j}/4}$, and set
\begin{equation*}
    X_{\mathrm{data}}\mid k=\mu_k+Q_k(Z_k-4\vartheta_k).
\end{equation*}
Consequently,
\begin{equation*}
    p_{k,\mathrm{data}}(x) =\prod_{j=1}^{d}
      g\!\left([Q_k^\top(x-\mu_k)]_j + 4\vartheta_{k,j};4,\vartheta_{k,j}\right),\quad
    \mathbb E[X_{\mathrm{data}}\mid k]=\mu_k,
    \quad
    \operatorname{Cov}(X_{\mathrm{data}}\mid k)=\Sigma_k.
\end{equation*}
The shape parameter is $4$ in every coordinate, and scales and rotations vary across components. Coordinates are independent in the rotated component coordinates, but generally dependent in the original coordinates.

\paragraph{Noised densities and exact scores.}
For either family, the noised component has mean and covariance
\begin{equation*}
    m_{k,t}=\alpha_t\mu_k,
    \qquad C_{k,t}=\alpha_t^2\Sigma_k+\bar\sigma_t^2I_d.
\end{equation*}

The noised component density and score of Gaussian mixture are
\begin{equation*}
\begin{aligned}
    p_{k,t}(x)&=\mathcal N(x;m_{k,t},C_{k,t}),\\
    s_{k,t}(x)&=-C_{k,t}^{-1}(x-m_{k,t}).
\end{aligned}
\end{equation*}

Next, we derive the noised component density and score of Gamma mixture. Define
\begin{equation*}
    y_{k,t}(x)=Q_k^\top(x-m_{k,t})+4\alpha_t\vartheta_k.
\end{equation*}
In these rotated coordinates, the noised density factorizes as
\begin{equation*}
\begin{aligned}
    p_{k,t}(x)&=\prod_{j=1}^{d}h_{k,j,t}(y_{k,t,j}(x)),\\
    h_{k,j,t}(y)
    &=\int_0^\infty g(z;4,\vartheta_{k,j})
        \varphi_{\bar\sigma_t^2}(y-\alpha_t z)\,\mathrm dz,
\end{aligned}
\end{equation*}
where $\varphi_v$ denotes the one-dimensional $\mathcal N(0,v)$
density. The score in the original coordinates is
\begin{equation*}
\begin{aligned}
    s_{k,t}(x)
    &=Q_k\left(
       \frac{h'_{k,j,t}(y_{k,t,j}(x))}
            {h_{k,j,t}(y_{k,t,j}(x))}
       \right)_{j=1}^{d},\\
    h'_{k,j,t}(y)
    &=\int_0^\infty
       \frac{\alpha_t z-y}{\bar\sigma_t^2}
       g(z;4,\vartheta_{k,j})
       \varphi_{\bar\sigma_t^2}(y-\alpha_t z)\,\mathrm dz.
\end{aligned}
\end{equation*}
Thus the component scores are determined by one-dimensional
convolution densities and their derivatives.

\textbf{Conditional and unconditional densities and scores.}
For both families,
\begin{equation*}
\begin{aligned}
    p_t^c(x)&=p_{0,t}(x),
    &p_t^u(x)&=\frac15\sum_{k=0}^{4}p_{k,t}(x),\\
    s_t^c(x)&=s_{0,t}(x),
    &s_t^u(x)&=\sum_{k=0}^{4}\pi_{k,t}(x)s_{k,t}(x),
\end{aligned}
\end{equation*}
where
\begin{equation*}
    \pi_{k,t}(x) =\frac{p_{k,t}(x)}{\sum_{\ell=0}^{4}p_{\ell,t}(x)}
\end{equation*}
is the posterior probability of component $k$.

\paragraph{Construction of the score errors.}
For each of five independent error realizations, draw matrices
for $b\in\{c,u\}$ according to
\begin{equation*}
    (G_b)_{ij},(Z_b)_{ij}
    \overset{\mathrm{i.i.d.}}{\sim}\mathcal N(0,1/d),
    \qquad
    H_b=\frac{G_b+G_b^\top}{2},
    \qquad A_b=\frac{Z_b-Z_b^\top}{2}.
\end{equation*}
The gradient construction uses
\begin{equation*}
    H_{c,t}=H_c,
    \qquad
    H_{u,t}=H_c+\kappa_H\alpha_t^2(H_u-H_c),
\end{equation*}
and the weighted-solenoidal construction uses
\begin{equation*}
    A_c^{\mathrm{eff}}=A_c,
    \qquad A_u^{\mathrm{eff}}=A_c+\kappa_A(A_u-A_c),
    \qquad \kappa_H=\kappa_A=0.5.
\end{equation*}
To express the errors in the notation of the main text, set
\begin{equation*}
    M_{b,t}=  H_{b,t},
    \qquad w_{b,t}= \widetilde w_{b,t},
\end{equation*}
where the auxiliary fields $\widetilde w_{b,t}$ are defined below.
Then the prescribed error and its reference-density decomposition are
\begin{equation*}
\begin{aligned}
    e_t^b(x)&=\frac12M_{b,t}x+\frac12w_{b,t}(x)
             =v_t^b(x)+w_t^b(x).
\end{aligned}
\end{equation*}
The amplitudes are absorbed into $M_{b,t}$ and $w_{b,t}$.
Here $w_{b,t}$ denotes the field before the factor $1/2$ in the
prescribed error, whereas $w_t^b$ denotes the weighted-solenoidal
error itself. Since $H_{b,t}$ is symmetric,
\begin{equation*}
    v_t^b=\nabla_x\!\left(\frac\eta2x^\top H_{b,t}x\right).
\end{equation*}

In the Gaussian mixture: for each component and conditional or unconditional construction, 
define
\begin{equation*}
K_{k,b,t}=C_{k,t}^{1/2}A_b^{\mathrm{eff}}C_{k,t}^{-1/2}, \quad \widetilde w_{k,b,t}(x) =K_{k,b,t}(x-m_{k,t}),
\end{equation*}
using the symmetric positive-definite square roots.
Skew symmetry gives $\operatorname{tr}(K_{k,b,t})=0$ and
$s_{k,t}^\top\widetilde w_{k,b,t}=0$. Hence
\begin{equation*}
    \nabla\cdot(p_{k,t}\widetilde w_{k,b,t})=0.
\end{equation*}

In Gamma mixture: 
Let $F_{k,j,t}$ be the cumulative distribution function associated with $h_{k,j,t}$, and let $\Phi$ and $\varphi$ be the standard normal cumulative distribution function and density. Set
$B_{k,b}=Q_k^\top A_b^{\mathrm{eff}}Q_k$ and define
\begin{equation*}
\begin{aligned}
    \chi_{k,t,j}(x)
    &=\Phi^{-1}\!\left(F_{k,j,t}(y_{k,t,j}(x))\right),\\
    [S_{k,t}(x)]_{jj}
    &=\frac{\varphi(\chi_{k,t,j}(x))}
            {h_{k,j,t}(y_{k,t,j}(x))},
    \qquad j=1,\ldots,d,\\
    \widetilde w_{k,b,t}(x)
    &=Q_kS_{k,t}(x)B_{k,b}\chi_{k,t}(x).
\end{aligned}
\end{equation*}
Under component $k$, $\chi_{k,t}$ has the standard normal
distribution. Since $B_{k,b}$ is skew-symmetric, the field
$B_{k,b}\chi$ is weighted-solenoidal under that density.
The derivative of the inverse quantile map is $Q_kS_{k,t}$; transporting the field through this map therefore yields
\begin{equation*}
    \nabla\cdot(p_{k,t}\widetilde w_{k,b,t})=0.
\end{equation*}
For a Gaussian component, this construction reduces to the
Gaussian-mixture formula above.

For either family, combine the component fields as
\begin{equation*}
    \widetilde w_{c,t}(x)=\widetilde w_{0,c,t}(x),
    \qquad
    \widetilde w_{u,t}(x)
    =\sum_{k=0}^{4}\pi_{k,t}(x)\widetilde w_{k,u,t}(x).
\end{equation*}
Because
\begin{equation*}
    p_t^u\widetilde w_{u,t}
    =\frac15\sum_{k=0}^{4}p_{k,t}\widetilde w_{k,u,t},
\end{equation*}
both $\widetilde w_{c,t}$ and $\widetilde w_{u,t}$ have zero weighted divergence under their respective reference densities.
Scaling preserves this property, so
\begin{equation*}
    \nabla\cdot(p_t^b w_t^b)=0,
    \qquad b\in\{c,u\}.
\end{equation*}
The fields have finite reference-density energies. The gradient and weighted-solenoidal subspaces are therefore orthogonal:
\begin{equation*}
    \mathbb E_{p_t^b}[v_t^b\cdot w_t^b]=0.
\end{equation*}
Thus $v_t^b$ and $w_t^b$ are precisely the gradient and
weighted-solenoidal components of $e_t^b$ under $p_t^b$.

\paragraph{Error magnitude under reference distribution.}
Using the notation of the main text, define
\begin{equation*}
\begin{aligned}
    E_{\mathrm{vis}}^b(t)&=\mathbb E_{p_t^b}[\|v_t^b\|^2],\quad E_{\mathrm{inv}}^b(t)=\mathbb E_{p_t^b}[\|w_t^b\|^2],\\
E_{\mathrm{tot}}^b(t)&=E_{\mathrm{vis}}^b(t)+E_{\mathrm{inv}}^b(t),
    \qquad R_{\mathrm{inv}}^b(t)
    =\frac{E_{\mathrm{inv}}^b(t)}{E_{\mathrm{tot}}^b(t)}.
\end{aligned}
\end{equation*}
The subscripts $\mathrm{vis}$ and $\mathrm{inv}$ refer to the gradient and weighted-solenoidal components under the specified reference density. These reference error energies are distinct from the guided activation energy $\mathcal L_t(\gamma)$.
We average the energies over the five error realizations and report the ratio of the averaged weighted-solenoidal and total energies.

In Gaussian mixture, the error $L_2$ norms under conditional density are computed analytically:
\begin{equation*}
\begin{aligned}
    E_{\mathrm{vis}}^c(t)
    &=(1/4)\!\left\{
      \operatorname{tr}(H_cC_{0,t}H_c^\top)
      +\|H_cm_{0,t}\|^2\right\},\\
    E_{\mathrm{inv}}^c(t)
    &=(1/4)\operatorname{tr}(C_{0,t}A_cA_c^\top).
\end{aligned}
\end{equation*}
Here $m_{0,t}=0$ because $\mu_0=0$. Both unconditional energies are estimated using 65,536 independent samples from each reported marginal.

In Gamma mixture, the error $L_2$ norms under conditional and unconditional densities are estimated by
Monte Carlo sampling from their respective noised reference densities.

\paragraph{Guided sampling and density-ratio contributions.}
For both families, the exact and perturbed scores are
\begin{equation*}
\begin{aligned}
    \hat s_t^b&=s_t^b+e_t^b,\\
    s_t^\gamma&=\gamma s_t^c+(1-\gamma)s_t^u,\\
    \hat s_t^\gamma&=s_t^\gamma+e_t^\gamma,
    \qquad e_t^\gamma=\gamma e_t^c+(1-\gamma)e_t^u.
\end{aligned}
\end{equation*}
The corresponding velocities are $\zeta_t^\gamma$ and
$\hat\zeta_t^\gamma=\zeta_t^\gamma-\kappa_t e_t^\gamma$,
as in the main text. Although the scores are available directly in these experiments, we retain the notation $\hat s_t^b$ for the prescribed approximation to the exact score.
For numerical integration, we reparameterize the ODE by
$\lambda=\lambda_t$. With $X_\lambda=X_{t(\lambda)}$ and the same convention for the coefficients and scores, the exact-score flow satisfies
\begin{equation*}
    \frac{\mathrm dX_\lambda}{\mathrm d\lambda}
    =\frac{\bar\sigma_\lambda^2}{2}
       \bigl(X_\lambda+s_\lambda^\gamma(X_\lambda)\bigr).
\end{equation*}
The perturbed flow replaces $s_\lambda^\gamma$ by
$\hat s_\lambda^\gamma$. Here $\lambda$ increases during sampling, while the noise time $t$ decreases.

Both flows start from the same finite-noise conditional density:
\begin{equation*}
    \rho_T^\gamma=\hat\rho_T^\gamma=\pi, \qquad \pi=p_T^c.
\end{equation*}
This equality between the two guided starting laws is imposed exactly; it does not require $p_T^u=p_T^c$ at the finite noising endpoint. In particular, it removes the starting-density-ratio term from the comparison of $\hat\rho_t^\gamma$ with $\rho_t^\gamma$.

We evaluate $\gamma\in\{1,3,5,7.5,10,12\}$ using 512 initial samples per error realization. Within each distribution family, the initial sample set is shared across guidance strengths and error realizations. At each reported time $t$, the contributions along the learned trajectory are
\begin{equation*}
\begin{aligned}
    C_I^{\gamma,t}(x)
    &=-\int_t^T\kappa_u
      \bigl(\nabla\cdot v_u^\gamma
       +v_u^\gamma\cdot\nabla\log\rho_u^\gamma\bigr)
      \bigl(\hat\Phi_{t,u}^\gamma(x)\bigr)\,\mathrm du,\\
    C_A^{\gamma,t}(x)
    &=\int_t^T\kappa_u q_u^\gamma
      \bigl(\hat\Phi_{t,u}^\gamma(x)\bigr)\,\mathrm du.
\end{aligned}
\end{equation*}
Here $v_t^\gamma=\gamma v_t^c+(1-\gamma)v_t^u$,
$w_t^\gamma=\gamma w_t^c+(1-\gamma)w_t^u$, and
$q_t^\gamma=Q_{\rho_t^\gamma}w_t^\gamma$ is the activation
residual. The map $\hat\Phi_{t,u}^\gamma(x)$ follows the learned flow from position $x$ at time $t$ to time $u$.
For each error realization, the characteristic identity gives
\begin{equation*}
    C_I^{\gamma,t}(x)+C_A^{\gamma,t}(x)
    =\log\frac{\hat\rho_t^\gamma(x)}{\rho_t^\gamma(x)}.
\end{equation*}
As in the main text, write
\begin{equation*}
    D_I^\gamma=\mathbb E_{\hat\rho_t^\gamma}[C_I^{\gamma,t}],
    \qquad
    D_A^\gamma=\mathbb E_{\hat\rho_t^\gamma}[C_A^{\gamma,t}],
\end{equation*}
with the evaluation time $t$ implicit. For expectations over the full perturbed distribution, and under the stated integrability conditions,
\begin{equation*}
    D_I^\gamma+D_A^\gamma
    =\mathrm{KL}(\hat\rho_t^\gamma\|\rho_t^\gamma).
\end{equation*}
These are signed inherited and activated contributions to the log-density ratio, rather than separate divergences or energies. Reported means also average over the five error realizations. The identity applies to each realization before this averaging as numerical integration and finite-sample averaging introduce approximation error.

Under strong guidance, some perturbed trajectories enter regions of extremely low density in exact-score guided flow, producing extreme log-density ratios. We exclude trajectories whose inverse-flow norm exceeds $1{,}000$. For $\gamma=1,3,5,7.5,10$, the excluded fractions are 0.00\%, 0.00\%, 0.31\%, 1.33\%, and 4.41\%,
respectively.
Contributions are averaged over the retained trajectories within each error realization and then over the five realizations.
The common initial sample set includes one block resampled
after inspection of an extreme contribution. The resulting set is shared across all guidance strengths. Rejection and this data-dependent resampling affect the averaging distribution, so the reported sums need not equal the full-distribution KL divergence.

\subsection{Additional results for the synthetic experiments}
\subsubsection{Gaussian mixture}\label{sec:gmm decomposition}
Table~\ref{tab:gmm_native_energy} reports gradient and weighted-solenoidal error magnitudes under corresponding reference distribution along the noise schedule.
The weighted-solenoidal component accounts for approximately 47\% of the conditional error magnitude and 32--38\% of the unconditional error magnitude. Thus a substantial fraction of the prescribed error has zero marginal source under its respective reference density.
\begin{table}[H]
\centering
\caption{
Native score-error energies along the sampling schedule
with $\eta=\xi=0.05$,
averaged over five error realizations.
The conditional branch corresponds to Class 0, while
the unconditional branch uses the full five-component
mixture. Step 0 is the high-noise endpoint and Step 100
is the generation endpoint.
}
\label{tab:gmm_native_energy}
\setlength{\tabcolsep}{3pt}
\resizebox{\textwidth}{!}{%
\begin{tabular}{rcccccccc}
\toprule
& \multicolumn{4}{c}{Conditional (Class 0)}
& \multicolumn{4}{c}{Unconditional} \\
\cmidrule(lr){2-5}
\cmidrule(lr){6-9}
Step
& $E_{\mathrm{vis}}$
& $E_{\mathrm{inv}}$
& $E_{\mathrm{tot}}$
& $R_{\mathrm{inv}}$ (\%)
& $E_{\mathrm{vis}}$
& $E_{\mathrm{inv}}$
& $E_{\mathrm{tot}}$
& $R_{\mathrm{inv}}$ (\%) \\
\midrule
0
& 0.026205 & 0.023607 & 0.049812 & 47.39
& 0.026205 & 0.012386 & 0.038591 & 32.10 \\
10
& 0.026239 & 0.023631 & 0.049870 & 47.39
& 0.025909 & 0.012263 & 0.038172 & 32.13 \\
20
& 0.026336 & 0.023702 & 0.050038 & 47.37
& 0.025026 & 0.011980 & 0.037006 & 32.37 \\
30
& 0.026488 & 0.023811 & 0.050299 & 47.34
& 0.023665 & 0.011712 & 0.035377 & 33.11 \\
40
& 0.026679 & 0.023950 & 0.050629 & 47.30
& 0.022062 & 0.011558 & 0.033620 & 34.38 \\
50
& 0.026891 & 0.024103 & 0.050994 & 47.27
& 0.020569 & 0.011506 & 0.032075 & 35.87 \\
60
& 0.027103 & 0.024256 & 0.051359 & 47.23
& 0.019535 & 0.011498 & 0.031032 & 37.05 \\
70
& 0.027294 & 0.024394 & 0.051688 & 47.19
& 0.019129 & 0.011494 & 0.030623 & 37.53 \\
80
& 0.027446 & 0.024504 & 0.051950 & 47.17
& 0.019246 & 0.011484 & 0.030730 & 37.37 \\
90
& 0.027543 & 0.024574 & 0.052117 & 47.15
& 0.019556 & 0.011473 & 0.031029 & 36.97 \\
100
& 0.027577 & 0.024599 & 0.052175 & 47.15
& 0.019709 & 0.011468 & 0.031177 & 36.78 \\
\bottomrule
\end{tabular}%
}
\end{table}

\subsubsection{Gamma mixture}
\label{sec:gamma decomposition}
Table~\ref{tab:gamma_native_energy} reports comparable error magnitudes for the Gamma mixture. The weighted-solenoidal component accounts for approximately 47--48\% of the conditional error magnitude and 32--38\% of the unconditional error.
\begin{table}[H]
\centering
\caption{
Native score-error energies for the Gamma mixture
along the sampling schedule, averaged over five error realizations.
}
\label{tab:gamma_native_energy}
\setlength{\tabcolsep}{3pt}
\resizebox{\textwidth}{!}{%
\begin{tabular}{rcccccccc}
\toprule
& \multicolumn{4}{c}{Conditional (Class 0)}
& \multicolumn{4}{c}{Unconditional} \\
\cmidrule(lr){2-5}
\cmidrule(lr){6-9}
Step
& $E_{\mathrm{vis}}$
& $E_{\mathrm{inv}}$
& $E_{\mathrm{tot}}$
& $R_{\mathrm{inv}}$ (\%)
& $E_{\mathrm{vis}}$
& $E_{\mathrm{inv}}$
& $E_{\mathrm{tot}}$
& $R_{\mathrm{inv}}$ (\%) \\
\midrule
0
& 0.026166 & 0.023598 & 0.049764 & 47.42
& 0.026166 & 0.012395 & 0.038561 & 32.14 \\
10
& 0.026215 & 0.023639 & 0.049854 & 47.42
& 0.025868 & 0.012285 & 0.038153 & 32.20 \\
20
& 0.026338 & 0.023728 & 0.050066 & 47.39
& 0.024990 & 0.012001 & 0.036990 & 32.44 \\
30
& 0.026522 & 0.023863 & 0.050385 & 47.36
& 0.023636 & 0.011721 & 0.035357 & 33.15 \\
40
& 0.026748 & 0.024055 & 0.050804 & 47.35
& 0.022042 & 0.011569 & 0.033611 & 34.42 \\
50
& 0.026996 & 0.024320 & 0.051317 & 47.39
& 0.020559 & 0.011561 & 0.032121 & 35.99 \\
60
& 0.027241 & 0.024663 & 0.051903 & 47.52
& 0.019531 & 0.011660 & 0.031191 & 37.38 \\
70
& 0.027458 & 0.025062 & 0.052520 & 47.72
& 0.019127 & 0.011824 & 0.030951 & 38.20 \\
80
& 0.027627 & 0.025462 & 0.053088 & 47.96
& 0.019239 & 0.011992 & 0.031231 & 38.40 \\
90
& 0.027730 & 0.025769 & 0.053500 & 48.17
& 0.019540 & 0.012139 & 0.031679 & 38.32 \\
100
& 0.027759 & 0.025875 & 0.053634 & 48.24
& 0.019684 & 0.012203 & 0.031887 & 38.27 \\
\bottomrule
\end{tabular}%
}
\end{table}

\subsection{Details of the image experiments}
\label{sec:edm_performance}

\subsubsection{Strong guidance degradation with pretrained EDM}
\label{sec:edm_strong_guidance}

\begin{figure}[t]
    \centering
    \begin{minipage}[t]{0.49\textwidth}
        \centering
        \includegraphics[width=\linewidth]
        {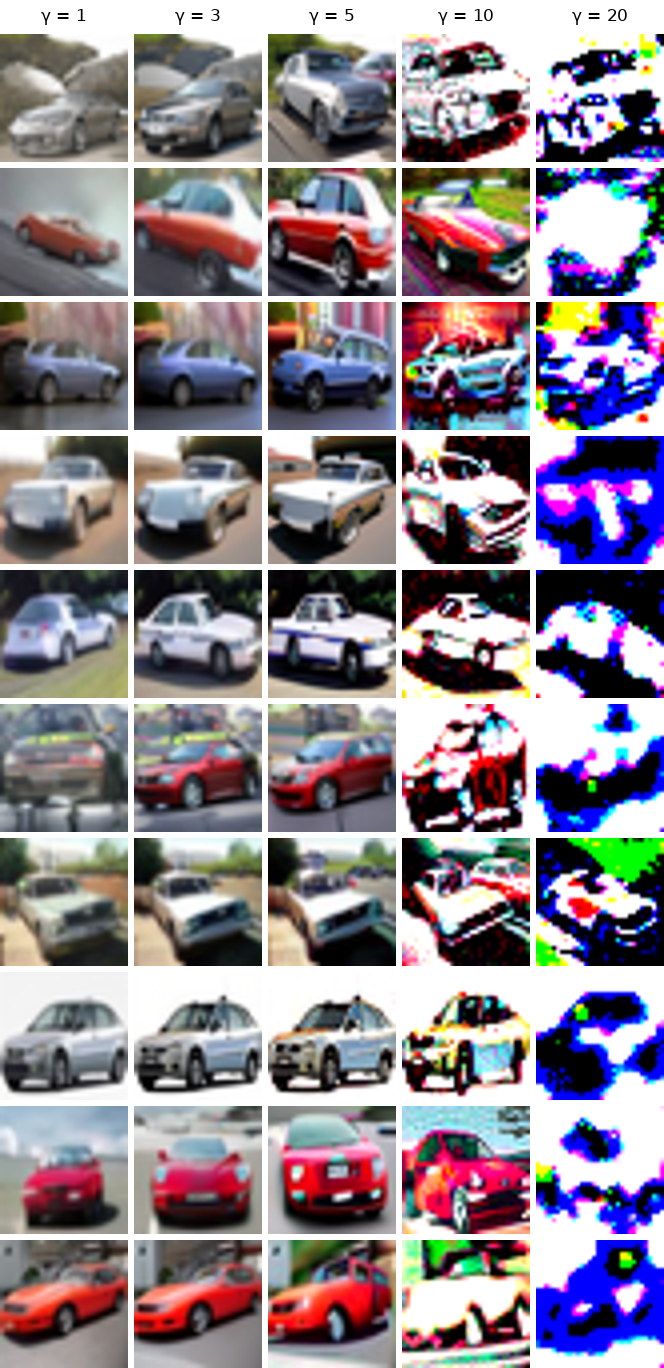}
        \par\smallskip
        {\small (a) DDIM}
    \end{minipage}
    \hfill
    \begin{minipage}[t]{0.49\textwidth}
        \centering
        \includegraphics[width=\linewidth]
        {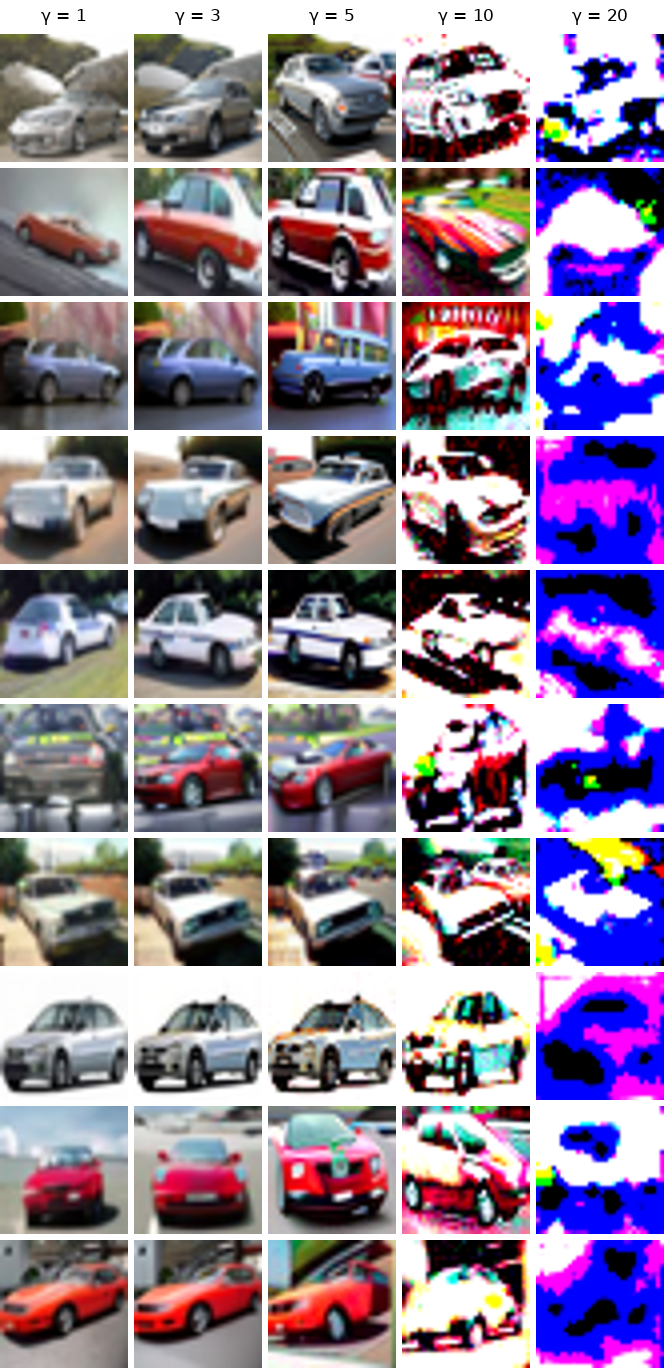}
        \par\smallskip
        {\small (b) Heun}
    \end{minipage}
    \caption{
    Automobile samples from pretrained CIFAR-10 EDM models, generated using (a) DDIM and (b) Heun, each with 40 sampling steps. Columns correspond to
    $\gamma=1,3,5,10,20$. All samples use the ``automobile'' condition, with initial noise matched across guidance strengths and samplers. At large guidance strengths, the displayed samples exhibit severe saturation and loss of fine detail with both samplers. }
    \label{fig:edm_strong_guidance}
\end{figure}

We first examine strong guidance degradation using pretrained EDM models on CIFAR-10. Figure~\ref{fig:edm_strong_guidance} compares automobile samples generated by DDIM and Heun at
$\gamma\in\{1,3,5,10,20\}$. Both samplers use 40 steps.
For each seed, the class label and initial noise are matched across guidance strengths and samplers.

Moderate guidance sharpens vehicle features in several examples relative to the conditional baseline at $\gamma=1$.
At $\gamma=10$, samples exhibit pronounced saturation and loss of fine detail. At $\gamma=20$, saturated color regions obscure much of the vehicle structure. Similar degradation appears with both samplers in these examples.

\begin{figure}[tp]
    \centering
    \begin{minipage}{0.49\linewidth}
        \centering
        \includegraphics[width=\linewidth]
        {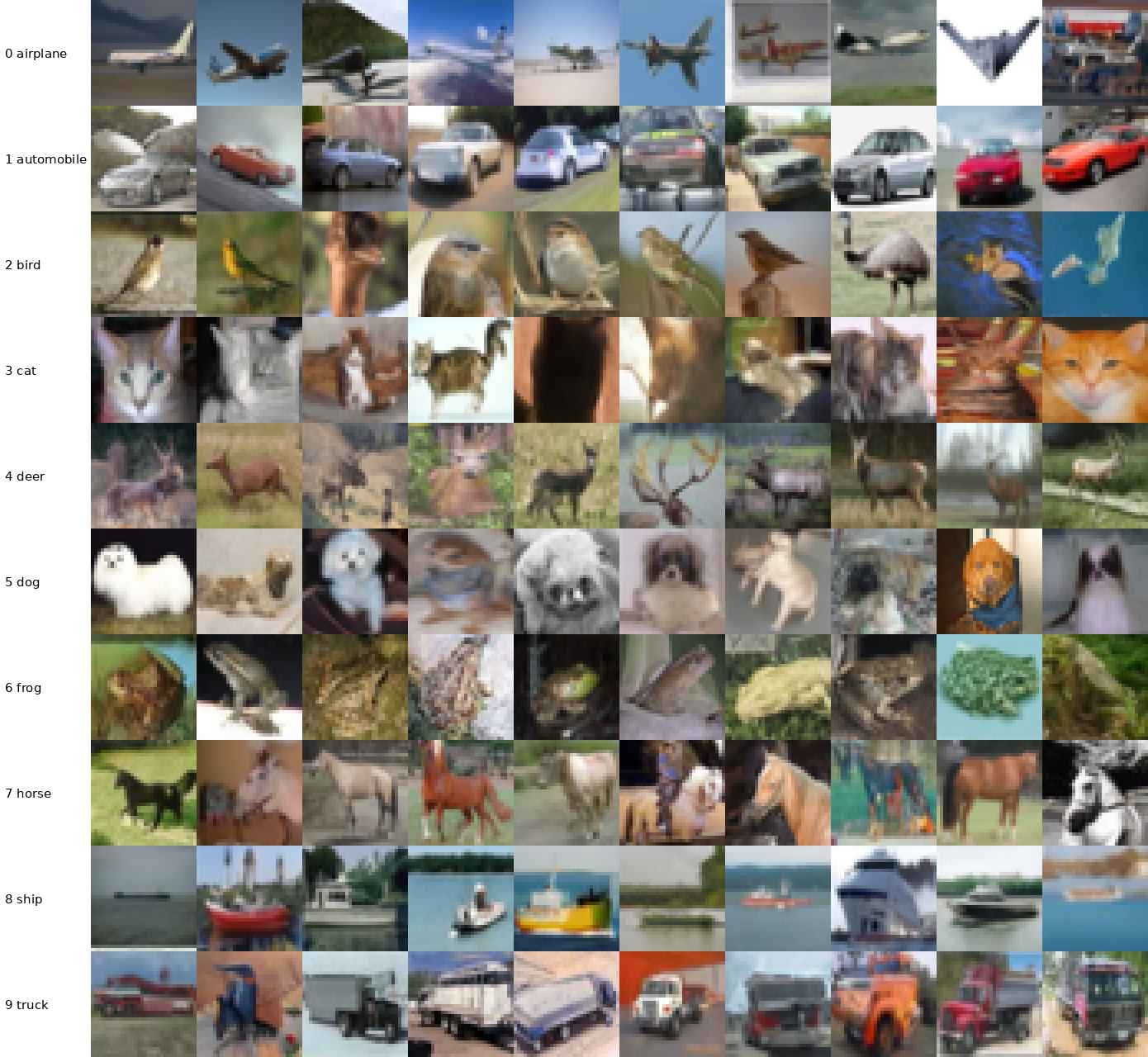}
        \par\smallskip
        {\small (a) $\gamma=1$}
    \end{minipage}
    \hfill
    \begin{minipage}{0.49\linewidth}
        \centering
        \includegraphics[width=\linewidth]
        {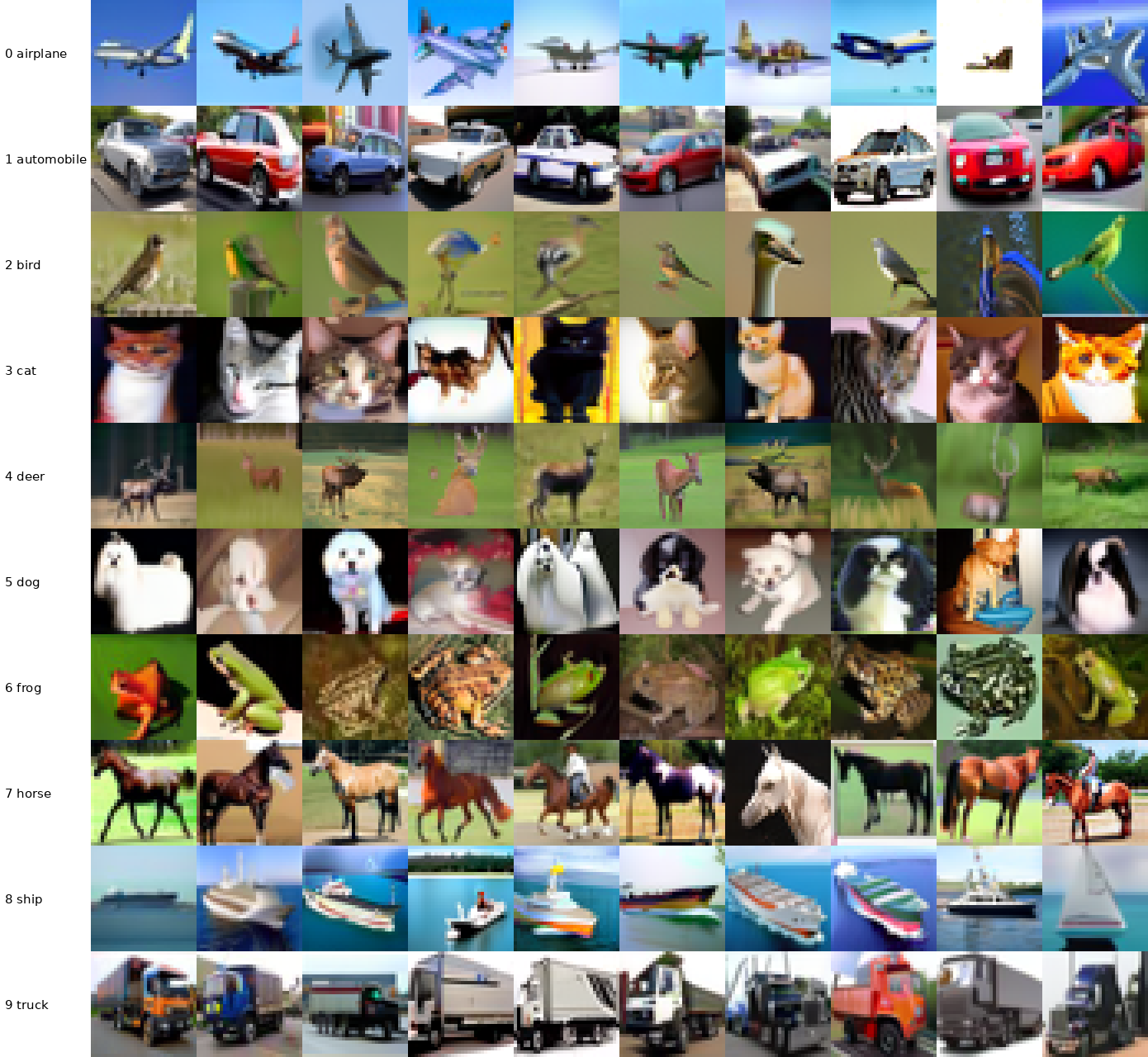}
        \par\smallskip
        {\small (b) $\gamma=5$}
    \end{minipage}

    \par\medskip

    \begin{minipage}{0.49\linewidth}
        \centering
        \includegraphics[width=\linewidth]
        {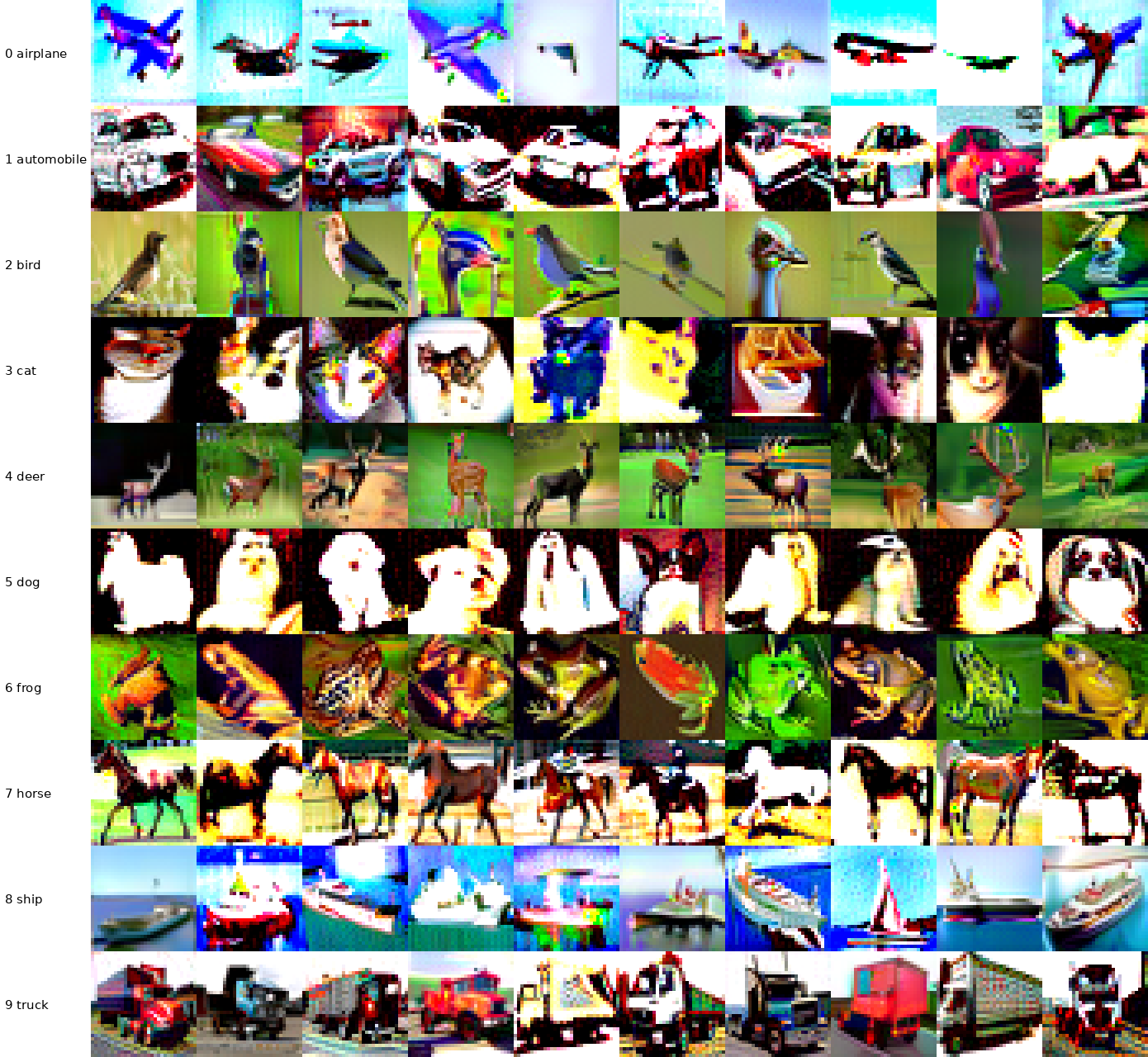}
        \par\smallskip
        {\small (c) $\gamma=10$}
    \end{minipage}
    \hfill
    \begin{minipage}{0.49\linewidth}
        \centering
        \includegraphics[width=\linewidth]
        {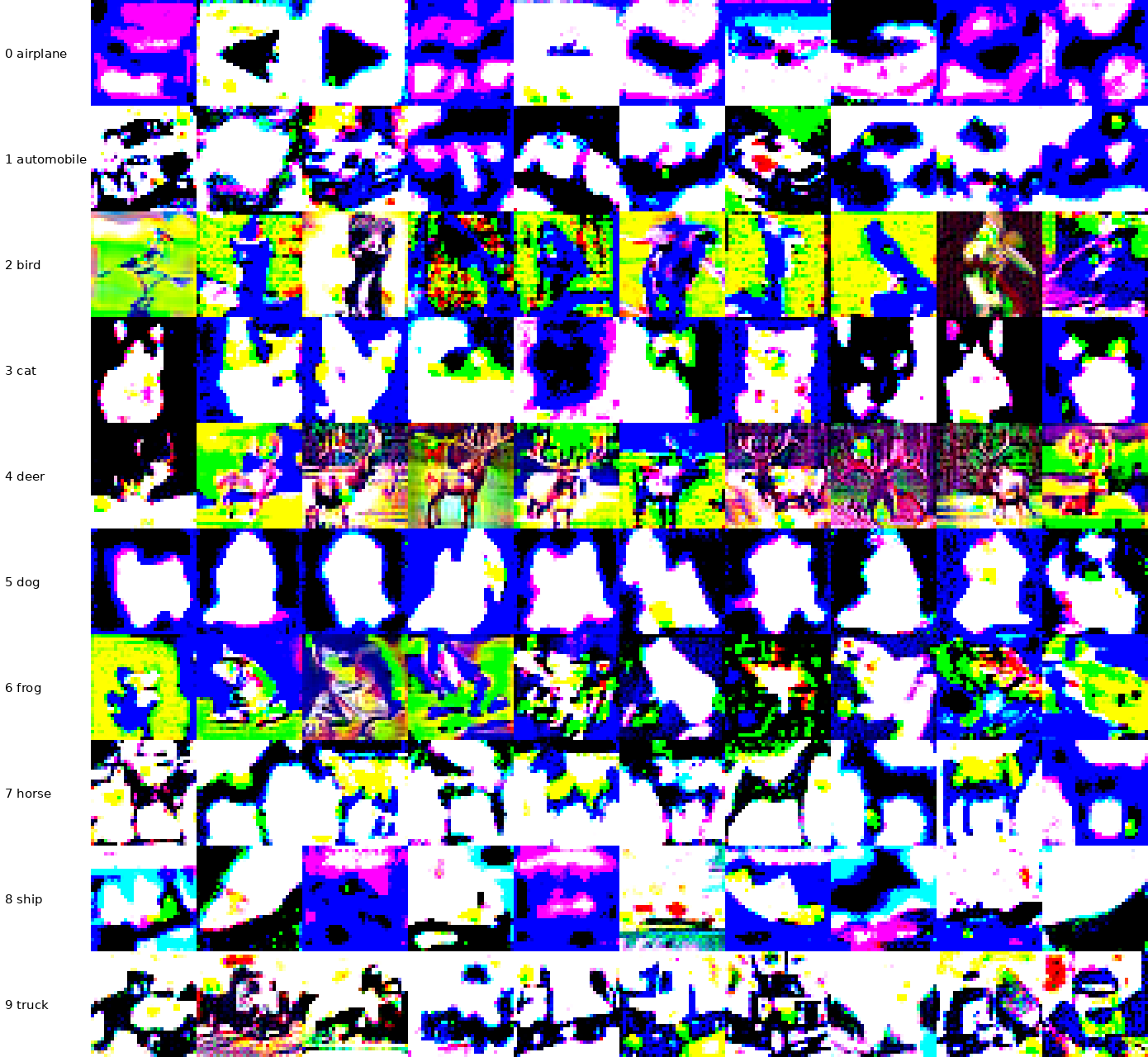}
        \par\smallskip
        {\small (d) $\gamma=20$}
    \end{minipage}
   \caption{
    Samples from pretrained EDM across all ten CIFAR-10 classes using 40-step DDIM sampling. Panels (a)--(d) correspond to $\gamma=1,5,10,20$, respectively. Each panel contains 100 samples, with one class per row and ten samples per class. Corresponding sample positions use the same initial noise and class labels across panels. At $\gamma=10$ and $20$, samples exhibit pronounced color saturation and loss
    of fine detail across multiple classes. }
    \label{fig:edm_ddim_all_classes}
\end{figure}

Figure~\ref{fig:edm_ddim_all_classes} extends this comparison to all ten CIFAR-10 classes using DDIM. Initial noise and class labels are matched across guidance strengths. Strong guidance produces saturation and loss of recognizable structure across multiple classes, consistent with the automobile examples.

\subsubsection{Negative controls for score filtering}
\label{sec:edm_distilled_branches}

\begin{figure}[H]
    \centering
    \begin{minipage}[t]{0.49\linewidth}
        \centering
        \includegraphics[width=\linewidth]
        {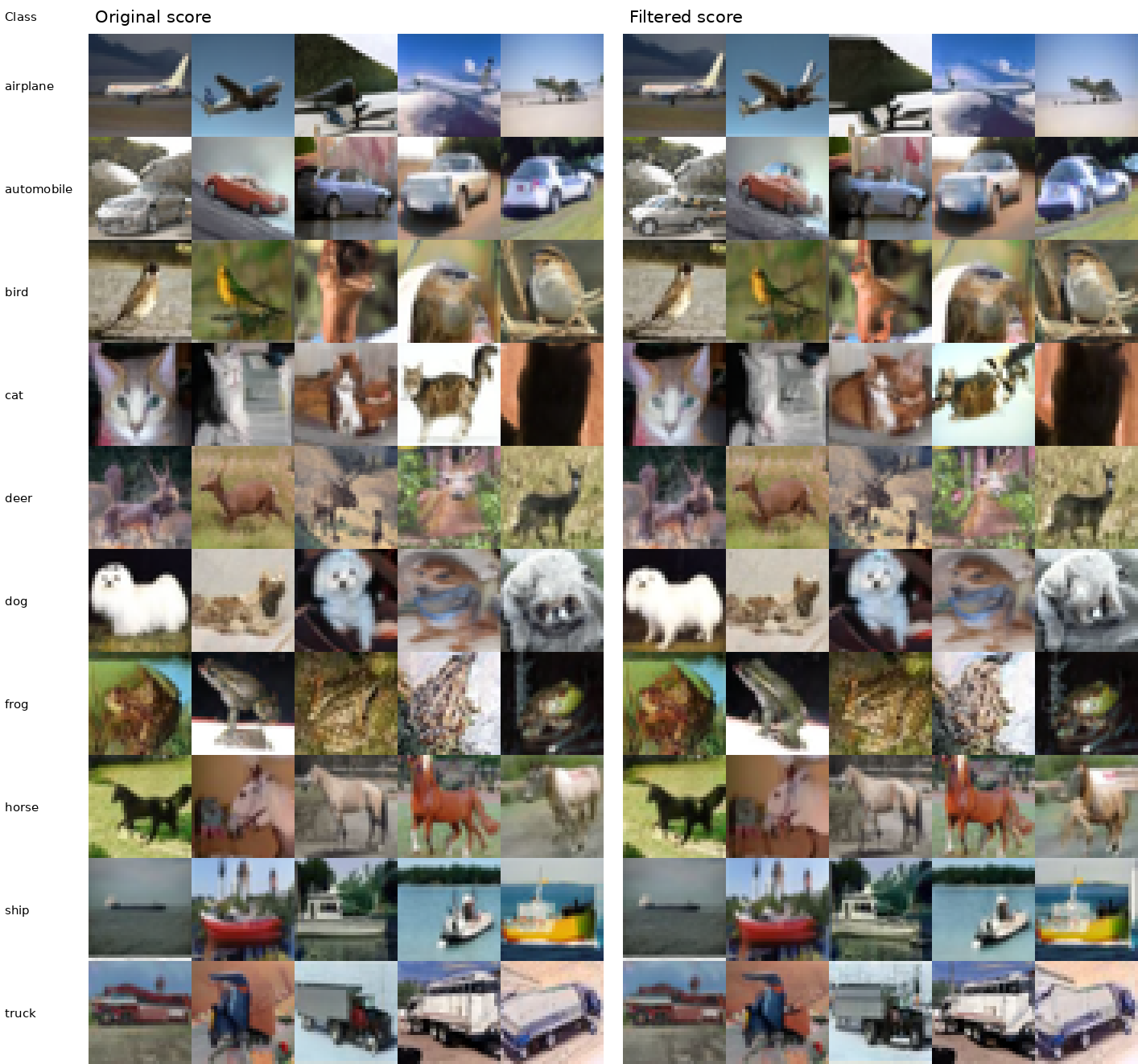}
        \par\smallskip
        {\small (a) Conditional-only ODE sampling}
    \end{minipage}
    \hfill
    \begin{minipage}[t]{0.49\linewidth}
        \centering
        \includegraphics[width=\linewidth]
        {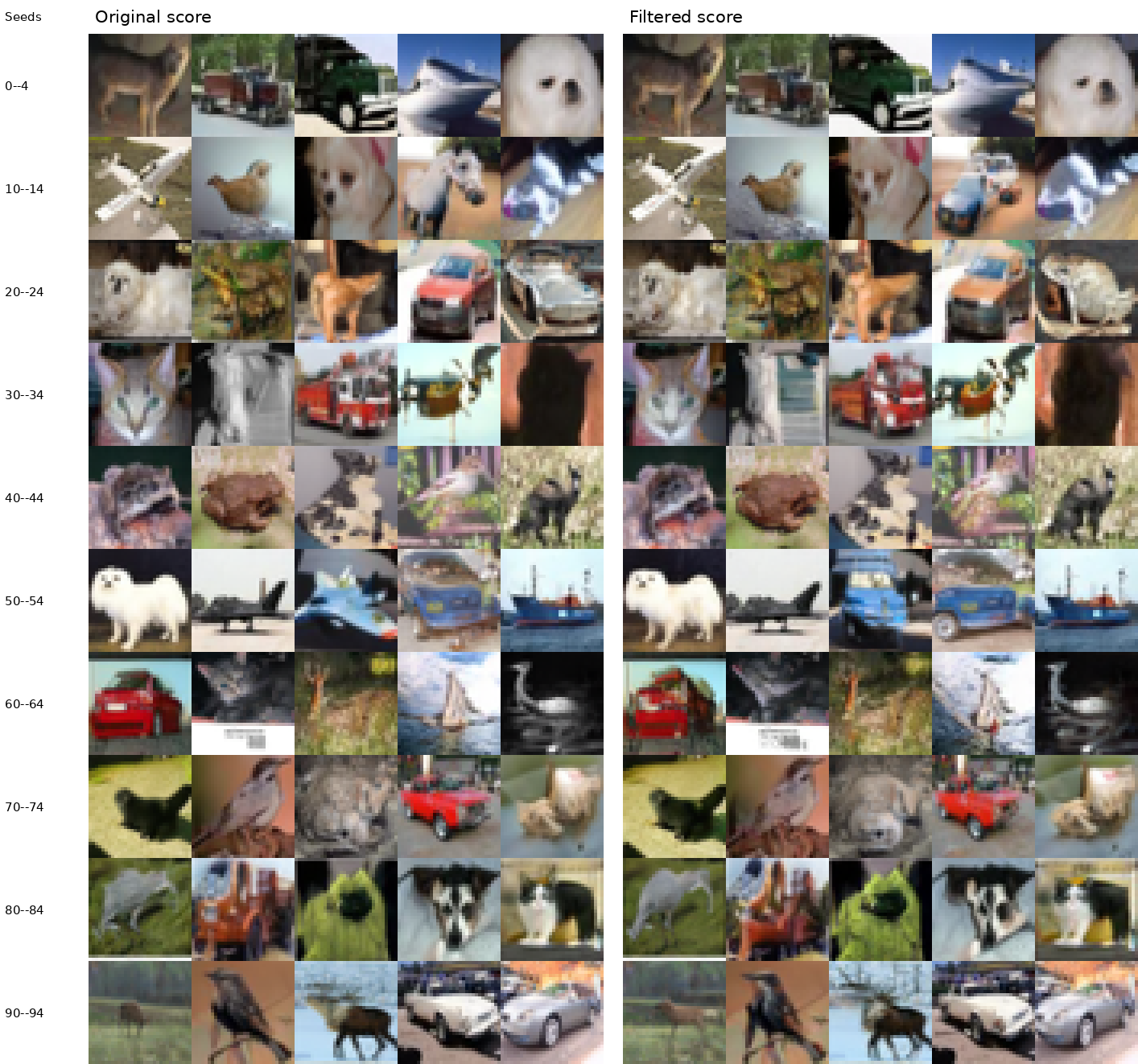}
        \par\smallskip
        {\small (b) Unconditional-only ODE sampling}
    \end{minipage}
    \caption{
        Negative controls for score filtering using 40-step Heun sampling. Each panel compares 50 original samples (left) with 50 filtered samples (right). Here ``original sample'' means samples generated by ODE steered by original score estimates, and ``filtered sample'' means samples generated by ODE steered by filtered scores.
        The conditional panel contains five matched sample pairs per CIFAR-10 class, with one class per row. Rows in the unconditional panel group matched seeds without prescribed class assignments. Initial Gaussian noise and applicable class labels are matched between the original and filtered models.
    }
    \label{fig:edm_distilled_branches}
\end{figure}

To assess the effect of filtering on the individual score
estimators, Figure~\ref{fig:edm_distilled_branches} compares
matched samples from the original and filtered models in
conditional-only and unconditional-only ODE sampling.
All samples use deterministic Heun sampling with 40 steps.
The original and filtered models share the same initial noise
and, for conditional sampling, the same class labels.
Conditional-only sampling corresponds to $\gamma=1$, and
unconditional-only sampling uses the unconditional score
without class labels.

Each fitted score is the gradient of a scalar potential,
$s_\sigma^{b,\mathrm{filter}}=\nabla_x\Phi_\theta^b$,
using the noise-level indexing of the main-text image
experiments. Since the exact score is also a gradient field,
the filtered score error has no weighted-solenoidal component
under its reference density, subject to the stated positivity
and integrability assumptions. Ideal filtering also preserves
the original gradient error, but fitting the potential may
modify it.

The displayed samples largely preserve the visual quality of their original counterparts, with some local differences in shape and texture. Together with the stronger differences observed under CFG, these controls are consistent with the proposed activation mechanism. They do not establish that the gradient error is unchanged or isolate its contribution to the guided comparisons.

\end{document}